\documentclass[review]{elsarticle}

\usepackage{lineno,hyperref}
\modulolinenumbers[5]

\usepackage[utf8]{inputenc} \usepackage[T1]{fontenc}    \usepackage{hyperref}       \usepackage{url}            \usepackage{booktabs}       \usepackage{amsfonts}       \usepackage{nicefrac}       \usepackage{microtype}      \usepackage[utf8]{inputenc} \usepackage[T1]{fontenc}    \usepackage{hyperref}       \usepackage{url}            \usepackage{color,xcolor}   \usepackage{times}          \usepackage{xspace}
\usepackage{amsmath, amssymb, amsthm, enumerate, bm, capt-of,mathtools}
\usepackage{natbib}
\usepackage{graphicx}
\usepackage{booktabs} 
\usepackage{natbib}
\usepackage{cite}	
\usepackage{amssymb,amsthm}
\usepackage{amsmath}
\usepackage{amsfonts}
\usepackage{breqn} 

\usepackage{algorithm}
\usepackage{algpseudocode}
   \usepackage[small, compact]{titlesec}
\usepackage{enumerate}

\usepackage{epstopdf}

\usepackage{graphicx}

\usepackage{caption}
\usepackage{subcaption}
\usepackage{grffile}

\usepackage[toc,page]{appendix}

\newcommand{\ours}{{ResGP}\xspace}
\newtheorem{theorem}{Theorem}[section]

\newtheorem{lem}[theorem]{Lemma}
\newtheorem{definition}[theorem]{Definition}
\newtheorem{assumption}[theorem]{Assumption}
\newtheorem{rem}[theorem]{Remark}

\newcommand{\Rset}{\mathbb{R}}

\begin{document}

\begin{frontmatter}
	\title{Residual Gaussian Process: A Tractable Nonparametric Bayesian Emulator for Multi-fidelity Simulations}
	\author[rvt3,rvt]{W.W.~Xing}
		\author[rvt2]{A.A. Shah\corref{cor1}}
				\author[rvt3,beihang2]{P. Wang}
	\author[rvt4]{S.~Zhe}
			\author[rvt2]{Q.~Fu}
	\author[rvt]{R.~M.~Kirby}
	\cortext[cor1]{Corresponding Author. E-mail: akeelshah@cqu.edu.cn;  Tel: +86 178 8023 0871}
		\address[rvt2]{School of Energy and Power Engineering, Chongqing University, 174 Shazhengjie, Shapingba, Chongqing 400044, China.}
\address[rvt3]{School of Integrated Circuit Science and Engineering, Beijing University of Aeronautics and Astronautics Address: No. 37 Xueyuan Road, Haidian District, 100191, China.}
	\address[rvt]{Scientific Computing and Imaging Institute, University of Utah, 72 S Central Campus Drive, 
Salt Lake City, UT 84112}
\address[rvt4]{School of Computing, University of Utah,University of Utah, 72 S Central Campus Drive, Room 3750 Salt Lake City, UT 84112}
\address[beihang2]{LMIB \& School of Mathematical Sciences, Beihang University, Beijing, China}

\begin{abstract}  
    Challenges in multi-fidelity modelling relate to accuracy, uncertainty estimation and high-dimensionality. A novel additive structure is introduced  in which the  highest fidelity solution is written as a sum of the lowest fidelity solution and residuals between the solutions at successive  fidelity levels, with Gaussian process priors placed over the low fidelity solution and each of the residuals. The resulting model is equipped with a closed-form solution for the predictive posterior, making it applicable to advanced, high-dimensional tasks that require uncertainty estimation. Its advantages are demonstrated on univariate benchmarks and on three challenging multivariate problems. It is shown how active learning can be used to enhance the model, especially with a limited computational budget. Furthermore, error bounds are derived for the mean prediction in the univariate case.
\end{abstract} 

\end{frontmatter}

\newcommand{\var}{{\rm Var}}
\newcommand{\Tr}{^{\rm T}}
\newcommand{\vtrans}[2]{{#1}^{(#2)}}
\newcommand{\kron}{\otimes}
\newcommand{\schur}[2]{({#1} | {#2})}
\newcommand{\schurdet}[2]{\left| ({#1} | {#2}) \right|}
\newcommand{\had}{\circ}
\newcommand{\diag}{{\rm diag}}
\newcommand{\invdiag}{\diag^{-1}}
\newcommand{\rank}{{\rm rank}}
\newcommand{\nullsp}{{\rm null}}
\newcommand{\tr}{{\rm tr}}
\renewcommand{\vec}{{\rm vec}}
\newcommand{\vech}{{\rm vech}}
\renewcommand{\det}[1]{\left| #1 \right|}
\newcommand{\pdet}[1]{\left| #1 \right|_{+}}
\newcommand{\pinv}[1]{#1^{+}}
\newcommand{\erf}{{\rm erf}}
\newcommand{\hypergeom}[2]{{}_{#1}F_{#2}}

\renewcommand{\a}{{\bf a}}
\renewcommand{\b}{{\bf b}}
\renewcommand{\c}{{\bf c}}
\renewcommand{\d}{{\rm d}}  \newcommand{\e}{{\bf e}}
\newcommand{\f}{{\bf f}}
\newcommand{\g}{{\bf g}}
\newcommand{\h}{{\bf h}}
\renewcommand{\k}{{\bf k}}
\newcommand{\m}{{\bf m}}
\newcommand{\mb}{{\bf m}}
\newcommand{\n}{{\bf n}}
\renewcommand{\o}{{\bf o}}
\newcommand{\p}{{\bf p}}
\newcommand{\q}{{\bf q}}
\renewcommand{\r}{{\bf r}}
\newcommand{\s}{{\bf s}}
\renewcommand{\t}{{\bf t}}
\renewcommand{\u}{{\bf u}}
\renewcommand{\v}{{\bf v}}
\newcommand{\w}{{\bf w}}
\newcommand{\x}{{\bf x}}
\newcommand{\y}{{\bf y}}
\newcommand{\z}{{\bf z}}
\newcommand{\A}{{\bf A}}
\newcommand{\B}{{\bf B}}
\newcommand{\C}{{\bf C}}
\newcommand{\D}{{\bf D}}
\newcommand{\E}{{\bf E}}
\newcommand{\F}{{\bf F}}
\newcommand{\G}{{\bf G}}
\renewcommand{\H}{{\bf H}}
\newcommand{\I}{{\bf I}}
\newcommand{\J}{{\bf J}}
\newcommand{\K}{{\bf K}}
\renewcommand{\L}{{\bf L}}
\newcommand{\M}{{\bf M}}
\newcommand{\N}{\mathcal{N}}  \newcommand{\MN}{\mathcal{MN}} 
\newcommand{\Acal}{\mathcal{A}}
\newcommand{\Ocal}{\mathcal{O}}
\newcommand{\Dcal}{\mathcal{D}}
\newcommand{\Ycal}{\mathcal{Y}}
\newcommand{\Zcal}{\mathcal{Z}}
\newcommand{\Fcal}{\mathcal{F}}
\newcommand{\Vcal}{\mathcal{V}}
\newcommand{\Lcal}{\mathcal{L}}
\newcommand{\Tcal}{\mathcal{T}}
\newcommand{\Gcal}{\mathcal{G}}
\newcommand{\Hcal}{\mathcal{H}}
\newcommand{\Scal}{\mathcal{S}}
\newcommand{\Xcal}{\mathcal{X}}

\renewcommand{\O}{{\bf O}}
\renewcommand{\P}{{\bf P}}
\newcommand{\Q}{{\bf Q}}
\newcommand{\R}{{\bf R}}
\renewcommand{\S}{{\bf S}}
\newcommand{\T}{{\bf T}}
\newcommand{\U}{{\bf U}}
\newcommand{\V}{{\bf V}}
\newcommand{\W}{{\bf W}}
\newcommand{\X}{{\bf X}}
\newcommand{\Y}{{\bf Y}}
\newcommand{\Z}{{\bf Z}}
\newcommand{\Mcal}{{\mathcal{M}}}
\newcommand{\Wcal}{{\mathcal{W}}}
\newcommand{\Ucal}{{\mathcal{U}}}

\newcommand{\bfLambda}{\boldsymbol{\Lambda}}

\newcommand{\bsigma}{\boldsymbol{\sigma}}
\newcommand{\balpha}{\boldsymbol{\alpha}}
\newcommand{\bpsi}{\boldsymbol{\psi}}
\newcommand{\bphi}{\boldsymbol{\phi}}
\newcommand{\boldeta}{\boldsymbol{\eta}}
\newcommand{\Beta}{\boldsymbol{\eta}}
\newcommand{\btau}{\boldsymbol{\tau}}
\newcommand{\bvarphi}{\boldsymbol{\varphi}}
\newcommand{\bzeta}{\boldsymbol{\zeta}}
\newcommand{\bepsilon}{\boldsymbol{\epsilon}}

\newcommand{\blambda}{\boldsymbol{\lambda}}
\newcommand{\bLambda}{\mathbf{\Lambda}}
\newcommand{\bOmega}{\mathbf{\Omega}}
\newcommand{\bomega}{\mathbf{\omega}}
\newcommand{\bPi}{\mathbf{\Pi}}

\newcommand{\btheta}{\boldsymbol{\theta}}
\newcommand{\bpi}{\boldsymbol{\pi}}
\newcommand{\bxi}{\boldsymbol{\xi}}
\newcommand{\bSigma}{\boldsymbol{\Sigma}}

\newcommand{\bgamma}{\boldsymbol{\bGamma}}
\newcommand{\bGamma}{\boldsymbol{\Gamma}}

\newcommand{\bmu}{\boldsymbol{\mu}}
\newcommand{\bnu}{\boldsymbol{\nu}}
\newcommand{\1}{{\bf 1}}
\newcommand{\0}{{\bf 0}}

\newcommand{\bs}{\backslash}
\newcommand{\ben}{\begin{enumerate}}
\newcommand{\een}{\end{enumerate}}

 \newcommand{\notS}{{\backslash S}}
 \newcommand{\nots}{{\backslash s}}
 \newcommand{\noti}{{\backslash i}}
 \newcommand{\notj}{{\backslash j}}
 \newcommand{\nott}{\backslash t}
 \newcommand{\notone}{{\backslash 1}}
 \newcommand{\nottp}{\backslash t+1}

\newcommand{\notk}{{^{\backslash k}}}
\newcommand{\notij}{{^{\backslash i,j}}}
\newcommand{\notg}{{^{\backslash g}}}
\newcommand{\wnoti}{{_{\w}^{\backslash i}}}
\newcommand{\wnotg}{{_{\w}^{\backslash g}}}
\newcommand{\vnotij}{{_{\v}^{\backslash i,j}}}
\newcommand{\vnotg}{{_{\v}^{\backslash g}}}
\newcommand{\half}{\frac{1}{2}}
\newcommand{\msgb}{m_{t \leftarrow t+1}}
\newcommand{\msgf}{m_{t \rightarrow t+1}}
\newcommand{\msgfp}{m_{t-1 \rightarrow t}}

\newcommand{\proj}[1]{{\rm proj}\negmedspace\left[#1\right]}
\newcommand{\argmin}{\operatornamewithlimits{argmin}}
\newcommand{\argmax}{\operatornamewithlimits{argmax}}

\newcommand{\dif}{\mathrm{d}}
\newcommand{\abs}[1]{\lvert#1\rvert}
\newcommand{\norm}[1]{\lVert#1\rVert}

\newcommand{\ie}{{i.e., }}
\newcommand{\eg}{{e.g., }}
\newcommand{\etc}{{etc. }}

\newcommand{\EE}{\mathbb{E}}
\newcommand{\dr}[1]{\nabla #1}
\newcommand{\VV}{\mathbb{V}}
\newcommand{\sbr}[1]{\left[#1\right]}
\newcommand{\rbr}[1]{\left(#1\right)}
\newcommand{\cmt}[1]{}

\newcommand{\bi}{{\bf i}}
\newcommand{\bj}{{\bf j}}
\newcommand{\bK}{{\bf K}}
\newcommand{\Vtr}{\mathrm{Vec}}

\newcommand{\kl}{{\rm KL}}
\newcommand{\cov}{{\rm Cov}}	

\newtheorem{theore}{Theorem}
\newtheorem{propos}{Proposition}
\newtheorem{lemma}{Lemma}
\newtheorem{coroll}{Corollary}
\newtheorem{remark}{Remark}

\newcommand{\RR}{\mathbb{R}}

\newcommand{\bPsi}{\boldsymbol{\Psi}}
\newcommand{\bXi}{\boldsymbol{\Xi}}
\newcommand{\btx}{\textbf{\textit{x}}}
\newcommand{\bty}{\textbf{\textit{y}}}
\newcommand{\btz}{\textbf{\textit{z}}}
\newcommand{\btk}{\textbf{\textit{k}}}

\newcommand{\itk}{\textit{k}}
\newcommand{\ity}{\textit{y}}

\newcommand{\bupsilon}{\boldsymbol{\upsilon}} 

\section{Introduction}
{The design, optimization, and control of many systems in science and engineering  can rely heavily on computational modelling. Different approaches can be adopted, depending upon the problem at hand or the computational budget available. One way to categorize a computational model is via its {\/\it fidelity\/}. Roughly speaking, low-fidelity computational models are those of low complexity or resolution in terms of the physics or any adjustable setting of the computer-based approximation \citep{16M1082469}. High-fidelity models are defined in a similar manner. The former are usually associated with a lower computational burden, the penalty for which is a loss in accuracy.} In  all computational models, the settings can be adjusted  to obtain outputs of different fidelities (the grid point spacing,  time step, order of an approximating basis, error tolerances, and so on). It is also usually the case that there is a range of mathematical models to describe a given physical problem, with different levels of physical detail incorporated into the equations, initial-boundary conditions and geometry that make up the model.

In some applications, high-fidelity models are  impractical, especially when a high number of runs of the model is required across a parameter (input) space. {In such applications,  it is common to replace the model  with a computationally-inexpensive approximation, termed a surrogate model  \citep{kennedy2000predicting}. Example applications include design optimization \citep{J052375}, real-time control \citep{Galelli}, sensitivity analysis  \citep{Santner2003} and uncertainty quantification\citep{owenN}. Surrogate models mainly rely on machine learning \citep{kennedy2000predicting,conti2010bayesian,raissi2019physics,doi:10.2514/2.333} or model order reduction (MOR) \citep{Gunzburger}. MOR projects numerical formulations onto a low-dimensional subspace of the original space in which solutions are sought, and does not extend naturally to nonlinear or parameter-dependent problem.  Both MOR and machine-learning methods  require a large set of data generated from the original  high-fidelity model (either  for training or for constructing a basis), which may not be desirable or even feasible.

{Another route for reducing the computational burden is multi-fidelity modelling, in which models of different fidelity  are combined  \citep{kennedy2000predicting}. In most cases, multi-fidelity models involve the construction of one or more surrogate models that use information from the underlying models of different fidelity \citep{fernndezgodino2016review}. Other approaches include corrections to the  low-fidelity results by leveraging information from a limited number of high-fidelity simulations, using, e.g., a Taylor series expansion or a GP model  \citep{LEIFSSON201545}. Many of the multi-fidelity methods that have been developed are specific to certain tasks, especially optimization and uncertainty quantification. Importantly, the vast majority are concerned with scalar outputs or outputs in a low-dimensional space (we refer to \citep{16M1082469} for a recent review). }
 
{In the seminal autoregressive (AR) model of \citet{kennedy2000predicting} for univariate outputs, a linear relationship between the different fidelity levels was assumed to hold. 
\citet{legratiet2013multi}  enhanced this method by employing  a deterministic parametric form of the mapping from low- to high-fidelity, together with an efficient numerical scheme  to reduce the computational cost. Despite its advantages, this parametric approach requires expert knowledge for model selection, as well as a large training data set. \citet{perdikaris2017nonlinear} introduced the nonlinear autoregressive model  (NARGP)  to overcome some of these limitations. The authors placed a GP prior over the unknown cross-fidelity mapping, thereby increasing model flexibility and alleviating any overfitting issues. NARGP has been  applied to a number of low-dimensional problems \citep{perdikaris2017nonlinear}, and  has been generalized to high-dimensional outputs by \citet{parussini2017multi}. These methods, on the other hand,  lack a systematic approach to model training and rely on ad-hoc methods to select  a basis for  outputs in high-dimensional spaces. NARGP uses the low-fidelity solution as an input for the high-fidelity GP model, which leads to a concatenating GP structure known as the \textit{deep GP}~\citep{damianou2013deep}. As a consequence, tractability is lost and expensive sampling or variational  methods are required  for training and inference.}

{Another prominent multi-fidelity approach uses stochastic collocation (SC).}
\citet{narayan2014stochastic}  developed a greedy procedure to select low-fidelity samples and  identify inputs at which to conduct a low number of  high-fidelity simulations. The authors then used the low-fidelity results to approximate the coefficients for a high-fidelity  SC approximation. This and related SC approaches are based on the naive assumption that  adjacent fidelities share the same correlation structure, which is not always justified for complex models.
The most serious drawback of SC approaches is that they require out-of-sample executions of the low-fidelity model for making  predictions. When, as is frequently the case,  the low-fidelity simulations are also expensive, SC approaches are not suitable for applications requiring many runs, especially sensitivity/uncertainty analyses and optimization. 

{NARGP and deep GP approaches  becomes impractical or non-viable for multi-fidelity simulations involving outputs in high-dimensional spaces; the number of  parameters in these approaches scales linearly and quadratically with the output space dimension, respectively.
The recently developed Greedy NAR~\citep{XING2020100012}  attempts to bridge the gap between NARGP and SC,  leveraging  the advantages of both methods. This is achieved by a generalised AR model, in which the high-fidelity solution is given as linear map of the low-fidelity solution in a feature space. The feature map is implicitly defined by integrating out a weight matrix and kernelizing. Greedy NAR can take  advantage of a sequential active learning framework to select low- and high-fidelity samples efficiently. It was shown to be more efficient and accurate than NARGP, and more flexible than SC in terms of making high-fidelity predictions since it does not rely on out-of-sample low-fidelity experiments.} 

{In this work, an additive GP structure for multi-fidelity modelling is proposed, in which the highest fidelity solution is treated as the sum of the lowest fidelity solution and residuals between successive fidelities, over each of which a GP prior is placed. This structure leads to a flexible, tractable and highly-scalable multi-fidelity GP model, termed \ours. Both the  likelihood (for model training) and the posterior (for predictions) are given explicitly, so that  expensive   approximate inference methods are avoided, and no additional data is required to make predictions. Importantly, 
\ours can be scaled to  high-dimensional problems, which are common candidates  for surrogate modeling, without any compromise in the prediction accuracy.}

{Equipped with the tractable posterior, an active learning method is implemented so that \ours can automatically select inputs that maximize the information gain (or any other desired gain) without the need for a special experimental design or \textit{a priori} assumptions about the physical model. For univariate \ours we develop error bounds by assuming that the underlying high-fidelity solution is a sample from its GP prior \citep{lederer2019uniform}, together with additional mild assumptions related to  the regularities of the solution and  kernel. The computational complexity of \ours is compared with other state-of the-art methods, highlighting its advantages in terms of  scalability and parameter count. }

{\ours  is applied  to  five synthetic univariate examples and three challenging multivariate examples that involve several quantities of interest. In the univariate examples \ours is demonstrated to  outperform AR, NARGP and the multi-fidelity deep GP (MF-DGP) model of \citet{cutajar2019deep} in terms of both prediction accuracy and uncertainty estimation in almost all cases.  {Compared to the other methods, the root mean square error on 1000 test points is at least 21\%, 28\%, 58\%, 58\% and 7\% lower on the five examples, while the mean negative log likelihood is at least 13\%, 79\%, 38\% and 6\% lower on four of the examples.} The results from the multivariate examples are compared to those from NARGP, SC and Greedy NAR,  demonstrating considerable improvements in prediction accuracy, as well as  stable performance for high-dimensional data sets. {The normalised root mean square errors on the test sets  are shown to be  lower than those for the other methods (by up to 97\%) for 7 out of 9 quantities of interest across the three examples. Significantly, \ours, especially with active learning, performs particularly well for low numbers of high-fidelity training points.}

\section{Statement of the problem}\label{sec:Statement}

{We are interested in numerical solutions to systems of ordinary or partial  differential equations obtained from a computational model, and where repeated runs of the computational model for different input parameter values (associated with the system of equations and/or the accompanying initial-boundary conditions) are required. Such systems of equations are usually derived from conservation laws and, together with initial-boundary conditions, govern quantities of interest such as a species concentration or the temperature of a medium. Depending upon the types of equations, we may denote one such quantity of interest by $u(\x,t;\bxi)$, $u(\x;\bxi)$ or $u(t;\bxi)$,  where   $\x \in \Omega \subset \mathbb{R}^p$, $p=1,2,3$, is the spatial coordinate and $t$ is time.  $\bxi \in \mathcal{X}\subset \mathbb{R}^l$  is a vector of parameters that appear in the system of equations and/or in the initial-boundary conditions. $\mathcal{X}$  is the admissible input space, which is assumed to be a compact subset of $\mathbb{R}^l$.} 

{Numerical solutions of the system of equations are obtained from a discretization of the equations, initial-boundary conditions and spatio-temporal domain, typically based on the finite element, finite volume, or finite difference method, together with a time-stepping scheme for the transient case. In the case of spatially-uniform systems, only a time discretization is required. The numerical solution takes the form of one or more quantities of interest at a predefined number of points in a discrete spatio-temporal grid, $\x_j$, $j=1,\hdots, N_x$, $t_i$, $i=1,\hdots,N_t$; in  the case of a finite element formulation, the coefficients in a finite-element basis expansion of the quantity of interest are computed, from which values of the quantity at an arbitrary number of spatio-temporal grid points can be extracted. For each input parameter $\bxi$, the obtained values of the quantity of interest can be vectorized as follows 
\begin{equation}\label{vecto}
\y(\bxi) = \left(u(\x_1,t_1,\bxi),\dots,u(\x_{N_x},t_1,\bxi),u(\x_1,t_2,\bxi),\dots,u(\x_{N_x}, t_{N_t}, \bxi) \right)^T,
\end{equation}
or in some other manner that is the same for each $\bxi$. We may then treat $\y(\bxi)$ or any scalar or vector quantity derived from $\y(\bxi)$  as the final quantity of interest; that is, as a function $\y:\mathcal{X}\rightarrow  \RR^d$ of the inputs, for some integer $d\ge 1$ (for example,  $d=N_xN_t$ as in Eq. (\ref{vecto})). Multiple quantities of interest can be modelled using the multi-fidelity method we develop by separately applying the method to each quantity. We therefore limit discussion to a single quantity of interest $\y(\bxi)$  in the presentation of the method below.}

To obtain a high-fidelity/accurate solution for $\y(\bxi)$, we generally need to use a fine discretization in space and time, a high-order stencil, a high-order basis expansion, or tight iteration bounds.
Lower-fidelity solutions can be obtained by relaxing these criteria or by using simpler physical models, e.g., spatial averaging, considering a 2-d slice or  linearizing. In this way, we can obtain numerical solutions $\y^f(\bxi)$ at different fidelities $f$. Other types of numerical outputs such as those from electronic-structure calculations or molecular dynamics simulations  can also be modelled using the framework we develop. The requirement is simply a computational model with variable input parameters and options for generating different fidelity solutions by adjusting  settings as described above. 

{In multi-fidelity modelling, we first conduct simulations at different fidelity levels $f=1,\hdots,F$ using inputs $\mathcal{X}^f=\{\bxi_n^f\}_{n=1}^{N_f}\subset \mathcal{X}$ to obtain outputs $\{\y_n^f\}_{n=1}^{N_f}$, where $\y_n^f=\y^f(\bxi_n^f)$. The outputs can be represented compactly as $\{\Y^{f}\}_{f=1}^F$, where the rows of $\Y^{f}\in \mathbb{R}^{N_f\times d}$ are the $N_f$ solutions $\y_n^f$ at fidelity level $f$. 
In line with the common setting for multi-fidelity emulation ~\citep{perdikaris2017nonlinear,cutajar2019deep}, we assume that the training inputs for fidelity $f$ are a subset of those for the preceding  fidelity $f-1$, i.e., $\mathcal{X}^{f}\subset\mathcal{X}^{f-1}$. We introduce an index notation for the extraction of subsets. Let  $\e_f\subset\{1,\hdots,N_{f-1}\}$, $f=2,\hdots,F$  be the  indices that extract the inputs from $\mathcal{X}^{f-1}$ to obtain $\mathcal{X}^f$. 
Extraction of the rows $\e_{f}$ of $\Y^{f-1}$ leads to a matrix   denoted 
$\Y^{f-1}_{\e_{f}}\in \mathbb{R}^{N_f\times d}$. Each row of $\Y^{f-1}_{\e_{f}}$ is  an $f-1$ fidelity solution that shares the same input as the corresponding row (output) in  $\Y^{f}$. This allows us to later compactly write residuals by subtracting $\Y^{f-1}_{\e_{f}}$ from the matrix $\Y^{f}$.}

The goal of this paper is to accurately approximate one or more high-fidelity quantities of interest $\y^{F}(\bxi)$ by efficiently combining lower-fidelity information. We are especially interested in quantities of interest that lie in high-dimensional spaces (large $d$). Moreover, the method we develop will be able to evaluate the predictive uncertainty efficiently and effectively. We also develop an efficient design-of-experiment that leads to optimal surrogates of the quantities of interest given limited computational resources. Specifically, the method selects locations at which to conduct the high-fidelity experiments according to the maximum information gain. {We also provide an analysis of the time and space complexity of our method and develop error bounds for the univariate case.}

\section{Residual Gaussian process model}
{In this section we  introduce a novel tractable and scalable structure to model multi-fidelity simulation  data. 
Rather than imposing a concatenating structure as in NARGP, which forgoes the tractable nature of a GP  \citep{rasmussen2006gaussian}, we decompose the GP defined over the high-fidelity output into a sum of GPs relating to the differences between successive fidelities. 
Specifically, we impose the following residual structure
\begin{equation}
    \y^{f}(\bxi) = \sum_{k=1}^f \r^{k}(\bxi), \quad f=1,\hdots,F,
\end{equation}
where $\r^{f}(\bxi)=\y^{f}(\bxi)-\y^{f-1}(\bxi)$, $f=2,\hdots,F$, are the residual functions between fidelities. The function $\r^{1}(\bxi)$, on the other hand,  is defined as $\r^{1}(\bxi)=\y^{1}(\bxi)$.  Note that we present the model for the multivariate case. The univariate case can be obtained from the formulae presented below in an obvious manner.}

{
The unknown functions $\r^{f}(\bxi)$ are then  treated as random processes that can be approximated using any probabilistic data-driven model. 
We place an independent, zero-mean GP prior over each residual function $\r^{f}(\bxi)$, \ie
\begin{equation}\label{condonlat}
    \r^{f}(\bxi) \sim \mathcal{GP}( \r^{f}(\bxi) \mid\0,k^{f}(\bxi,\bxi'|\btheta^{f}) \otimes \bOmega^{f} +  \delta(\bxi,\bxi')\tau^{f} \I ),  \quad f=1,\hdots,F,
\end{equation}
where $\bOmega^{f} \in \RR^{d \times d}$ is an unknown coregionalization matrix corresponding to the correlations between the components of  $\r^{f}(\bxi)$, $\otimes$ is the Kronecker product and $\delta(\bxi,\bxi')$ is the Kronecker-delta function. Without loss of generality, the mean is assumed to be identically zero by centering the observations. We note that this is the multivariate GP model of \citet{conti2010bayesian}, with a separable covariance structure. That is, the covariance matrices assume the forms $\K^{f}\otimes \bOmega^{f}$, in which $[\K^{f}]_{nm}= k^{f}(\bxi_n, \bxi_m|\btheta^{f})$, $\bxi_n, \bxi_m\in\mathcal{X}^f$, captures the correlations between values of $\r^{f}(\bxi)$ at different inputs $\bxi$ and $\bOmega^{f}$ captures the spatial correlations, i.e., between different components of $\r^{f}(\bxi)$. The covariance functions $k^{f}(\bxi, \bxi|\btheta)$ contain  unknown hyperparameters $\btheta^{f}$. The terms $\delta(\bxi,\bxi')\tau^{f}{\bf I}$ are included to account for zero-mean, i.i.d. measurement error (across inputs and spatial coordinates), or, equivalently, as  regularization terms  to prevent ill-conditioning during training. The $\tau^f$ are treated as hyperparameters or (optionally) fixed to some small values in the regularization interpretation~\citep{kennedy2001bayesian}. The distribution (\ref{condonlat}) is conditioned on the full set of hyperparameters $\{\btheta^{f},\Omega^f,\tau^f\}$  but to avoid notational clutter, conditioning on hyperparameters and inputs or observations is not explicitly indicated. In the measurement error interpretation, the  targets, i.e., given values of $\r^{f}(\bxi)$, are considered to be values of a latent function $\r_l^{f}(\bxi)$ corrupted by the i.i.d. noise $\pmb{\epsilon}^f$}
{
\begin{equation}
\r^{f}(\bxi)=\r_l^{f}(\bxi)+\pmb{\epsilon}^f, \quad f=1,\hdots,F,
\end{equation}
with priors $\r_l^{f}(\bxi) \sim \mathcal{GP}( \r_l^{f}(\bxi) \mid\0,k^{f}(\bxi,\bxi'|\btheta^{f}) \otimes \bOmega^{f})$ and $\pmb{\epsilon}^f\sim\mathcal{GP}( \pmb{\epsilon}^f\mid\0, \delta(\bxi,\bxi')\tau^{f} \I )$.}

Choosing the right kernel function for a specific application is non-trivial. 
When there is no prior knowledge to guide the choice, the automatic relevance determinant (ARD) kernel \citep{rasmussen2006gaussian}
\begin{equation}\label{ARDkernel}
	k^{f}(\bxi, \bxi'|\btheta^{f}) = \theta_0^{f} \exp\left(-(\bxi-\bxi')  \diag(\theta_1^{f},\ldots,\theta_l^{f}) (\bxi-\bxi')^T \right),  \quad f=1,\hdots,F,
\end{equation}
with $\btheta^{f}=(\theta_0^{f},\hdots,\theta_1^{f})^T$ is often used. The ARD kernel can freely capture the  influence of each individual input (coordinate of $\bxi$) on the output.
The hyperparameters $\{\tau^{f},\btheta^{f}\}$ can be estimated by maximizing  the  log-marginal likelihood (see section \ref{secModTrain}).

{The high-fidelity prior in this model can be written as follows
\begin{equation}
    \y^{F}(\bxi) \sim \mathcal{GP}\left(\y^{F}(\bxi)\left\lvert {\bf 0}, \sum_{f=1}^F \left[k^{f}(\bxi,\bxi'|\btheta^{f}) \otimes\bOmega^{f} + \delta(\bxi,\bxi')\tau^{f} \I\right]\right.\right),
\end{equation}
by virtue of the  independence assumption. Again, this is the GP over the noisy observations with cumulative i.i.d. noise variance $\sum_f\tau_f$, and underlying latent function $\y_l^{F}(\bxi)\sim \mathcal{GP}(\y^{F}_l(\bxi)\mid {\bf 0}, \sum_{f=1}^F k^{f}(\bxi,\bxi'|\btheta^{f}) \otimes\bOmega^{f})$. 

Beginning with the lowest fidelity level $f=1$, we assume the prior (\ref{condonlat}) and use inputs $\mathcal{X}^{1}$ together with outputs $\Y^{1}$  to learn $\{\tau^{1},\btheta^{1}\}$ and find the predictive posterior $\r^{1}_l(\bxi)\sim \N\left(\r^{1}_l(\bxi)\mid\bmu^{1}_r(\bxi), \V^{1}_r(\bxi)\right)$. At the next step we use inputs $\mathcal{X}^{2}$ and outputs $\R^{2}:=\Y^{2}-\Y^{1}_{\e_2}$  in the same procedure to learn $\{\tau^{2},\btheta^{2}\}$, and obtain the posterior for $\r_l^{2}(\bxi)\sim \N\left(\r^{2}_l(\bxi)\mid\bmu^{2}_r(\bxi), \V^{2}_r(\bxi)\right)$. Here we define a matrix of residuals $\R^{2}$ using the notation introduced in section \ref{sec:Statement}. This procedure is repeated up to fidelity level $f=F$, i.e., independently learning the hyperparameters associated with each $\r^{f}(\bxi)$, given observations $\{\R^{f}=\Y^{f}-\Y^{f-1}_{\e_f}\}_{f=2}^F$. The predictive posteriors over $\r_l^{f}(\bxi)$ are derived using standard Gaussian conditioning rules  \citep{conti2010bayesian}, and  the posterior for the high-fidelity latent function $\y_l^{F}(\bxi)$ can be written compactly as the following sum of GPs 
\begin{equation}
    \label{eq:rmgp}
    \begin{aligned}
     \y_l^{F}(\bxi) &= \sum_{f=1}^F\r_l^{f}(\bxi)\sim \N\left(\y_l^{F}(\bxi) \;\middle|\; \bmu^{F}(\bxi), \V^{F}(\bxi)\right), \\
     \bmu^{F}(\bxi) & =\sum_{f=1}^F\bmu^{f}_r(\bxi)=\sum_{f=1}^F (\bOmega^{f} \otimes \k^{f}(\bxi))^T (\bOmega^{f} \otimes \K^{f} + \tau^f \I)^{-1} \vec(\R^{f}), \\
     \V^{F}(\bxi) &= \sum_{f=1}^F\V^{f}_r(\bxi)=\sum_{f=1}^F [\bOmega^{f} \otimes k^{f}(\bxi,\bxi|\btheta^f)  - (\bOmega^{f} \otimes \k^{f}(\bxi))^T (\bOmega^{f} \otimes \K^{f} + \tau^f\I)^{-1} (\bOmega^{f} \otimes \k^{f}(\bxi))],
    \end{aligned}
\end{equation}
where $\vec{(\cdot)}$ denotes vectorization and $\k^{f}(\bxi)=(k^{f}(\bxi,\bxi_1|\btheta^{f}) ,\hdots,k^{f}(\bxi,\bxi_{N_f}|\btheta^{f}) )^T$ is the vector of covariances between the latent function values $\r_l^{f}(\cdot)$ at $\bxi$ and points in $\mathcal{X}^f$. We note that since the posteriors over each $\r^{f}(\bxi)$ are learned independently, the training procedure is parallelizable if implemented without active learning (discussed below).}

The formulation \eqref{eq:rmgp} considers a general case in which noise terms $ \tau^{f}\I$ are included.  
A computational model (of any fidelity) can, however, be treated as deterministic function without random noise \citep{xing2020shared}. Thus, we can consider the residual information to be deterministic and model it using GPs without the noise terms, {in which case $ \y_l^{F}(\bxi)= \y^{F}(\bxi)$ and the predictive posterior reduces to}
\begin{equation}
    \begin{aligned}
        \label{eq:rmgp noise free}
        \y^{F}(\bxi) &\sim \N\left(\y^{F}(\bxi) \;\middle|\; \bmu^{F}(\bxi), \V^{F}(\bxi)\right), \\
        \bmu^{F}(\bxi) &=  \sum_{f=1}^f \bmu^{f}_r(\bxi)=\sum_{f=1}^F (\bOmega^{f} \otimes \k^{f}(\bxi))^T (\bOmega^{f} \otimes \K^{f})^{-1} \vec(\R^{f}) \\
        &= \sum_{f=1}^F \I \otimes (\k^{f}(\bxi))^T (\K^{f})^{-1} \vec(\R^{f})\\
        &= \sum_{f=1}^F (\k^{f}(\bxi))^T (\K^{f})^{-1} \R^{f},\\
        \V^{F}(\bxi) &=  \sum_{f=1}^f \V^{f}_r(\bxi)=\sum_{f=1}^F \bOmega^{f} \otimes \left(k^{f}(\bxi, \bxi|\btheta^f) - (\k^{f}(\bxi))^T (\K^{f})^{-1} \k^{f}(\bxi)\right).
\end{aligned}
\end{equation}
Note that the coregionalization matrices $\bOmega^{f}$ cancel out for the expectation predictions (of any fidelity). This is consistent with   autokrigeability \citep{alvarez2012kernels}, which is utilized by \citet{xing2020shared} to deal with high-dimensional, single-fidelity mechanical design simulations. Noiseless data is the usual assumption for simulations (the so-called ground truth approximation), but one could attempt to incorporate systematic errors arising from the model formulation (model inadequacy), from the numerical approximation (numerical errors),  or from parameter uncertainty. 

In the same spirit as \citet{xing2020shared}, we now simplify the model by setting $\bOmega^{f} = \I $ for $f=1,\dots,F$. This assumes that the components of $\y^{F}(\bxi)$ are mutually  independent given $\{\btheta^f\}$. Note that the variance of the prediction then ignores the spatial correlations, which will affect the uncertainty estimation, but the mean prediction is unaffected.
We will discuss this issue further in Section \ref{sec:auto}, in which we see that the active learning process is not affected.} 

{Retaining the noise terms and coregionalisation matrices, i.e., retaining Eq. (\ref{eq:rmgp}), requires an additional set of hyperparameters, namely the $\tau^f$ and the entries of each $\bOmega^{f}$. We may even use a richer covariance structure, such as the linear model of coregionalisation \citep{XING2021228930}  with linear combinations of separable covariances. In low-dimensional output spaces, this could be of benefit in terms of accuracy and variance capture. In very high-dimensional spaces, on the other hand, these approaches lead to problems in terms of maximizing the likelihood (or evidence lower bound) due to the higher number of inputs in the optimization problem  \citep{XING2021228930} (see section \ref{compcomplex}). If modelling the noise is deemed important for a particular problem (perhaps because the solutions are known to be corrupted), a compromise could be achieved by retaining the $\{\tau^f\}$ terms and setting $\bOmega^{f}=\I$. We discuss this formulation later. We point out, however, that in the examples in Section 4, the noise terms are not included.}

{For most data sets, $N_f$ decreases in size with $f$, and is typically  small for large values of $f$. The uncertainty in the predictions for $\r_l^f$ (or $\r^f$) therefore increases with $f$.  We have to keep in mind, however, that that uncertainty is also bounded by the scales of the residuals between different fidelities. For high values of $f$, the simulations for adjacent fidelities would tend to be  similar, and therefore the residuals would tend to be small. Thus,  the added uncertainties for large $f$ will be small. For small $f$, the residuals between adjacent fidelities are expected to be larger but the uncertainties would be  smaller by virtue of the greater number of samples. The additive structure can, therefore,  potentially ensure that the final uncertainty reflects the true model uncertainty. We discuss this further in the first example in section \ref{Examp1}.}

\subsection{Model training and high-fidelity predictions}\label{secModTrain}
Given the residual information $\R^{f}$, we can derive the residual {marginal log-likelihood} at fidelity $f$ 
\begin{equation}
    \begin{aligned}
        \label{eq:likelihood f}
        \mathcal{L}^{f} &= -\frac{1}{2} | \I \otimes \K^{f}| - \frac{1}{2} \vec(\R^{f})^T (\I \otimes \K^{f})^{-1} \vec(\R^{(f}) - \frac{N_f}{2} \ln (2 \pi), \\ 
        &= -\frac{d}{2} |\K^{f}| - \frac{1}{2} \tr ((\R^{f})^T (\K^{f})^{-1} \R^{f} ) - \frac{N_f}{2} \ln (2 \pi),
    \end{aligned}
\end{equation} 
while the joint {marginal log-likelihood} is 
\begin{equation}
    \label{eq:likelihood joint}
    \mathcal{L} = \sum_{f=1}^F \mathcal{L}^{f} = \sum_{f=1}^F -\frac{d}{2} |\K^{f}| - \frac{1}{2} \tr ((\R^{f})^T (\K^{f})^{-1} \R^{f} ) - \frac{N_f}{2} \ln (2 \pi).
\end{equation}
The availability of residual information $\R^{f}$ for each fidelity relies on the fact that $\X^f\subset \X^{f-1}$. 
The marginal log likelihoods $\mathcal{L}^{f}$ are  independent from each other given the inputs, and thus the training process can be performed separately for each fidelity, {i.e., parallelised}. Making predictions using \ours is  straightforward because the posterior for  fidelity $F$ is the sum of the posteriors over  the residuals, each of which is Gaussian. The prediction is given by Eq. (\ref{eq:rmgp noise free}) with 
\begin{equation}\label{varwithout}
    \begin{aligned}
\V^{F}(\bxi) &=\sum_{f=1}^f \V^{f}_r(\bxi)= \sum_{f=1}^F {\bf I} \otimes \left(k^{f}(\bxi, \bxi|\btheta^f) - (\k^{f}(\bxi))^T (\K^{f})^{-1} \k^{f}(\bxi)\right).
    \end{aligned}
\end{equation}
\begin{algorithm}[H]
    \caption{\ours Sequential Construction}
	\label{algo:greedy construct}
	\begin{algorithmic}[h]
        \Require
        A finite-cardinality set $\mathcal{X}_t$ of inputs, and the 
        number of experiments allowed for each fidelity  $N_f$
        \Ensure A trained ResGP model
        \State Randomly select a point from $\mathcal{X}_t$ and form the initial 1-fidelity candidate set $\mathcal{X}^{1}$
        \State Conduct 1-fidelity experiment for $\mathcal{X}^{1}$ and collect the solution in matrix $\Y^{1}$
        \State Update residual matrix $\R^{1} = \Y^{1}$
        \For{$|\mathcal{X}^{1}| < N_1$}
        \State Update $1$-fidelity residual GP model by MLE \eqref{eq:likelihood f} ($f=1$) based on inputs $\mathcal{X}^{1}$ and residuals $\R^{1}$ 
        \State Find $\bxi_*$ based on Eq. \eqref{eq:var max f} for $\mathcal{X}_t \setminus \mathcal{X}^{1}$
        \State Update 1-fidelity candidate set $\mathcal{X}^{1} \leftarrow  \mathcal{X}^{1} \cup  \{\bxi_*\}$
        \State Conduct a 1-fidelity simulation to obtain solution $\y^{1}(\bxi_*)$
        \State Update low-fidelity solution matrix $\R^{1} \leftarrow  [\R^{1},\y^{1}(\bxi_*)]$
        \EndFor
        \For{$f = 2,\dots, F$}
        \State Randomly select a point from $\mathcal{X}^{f-1}$ to form initial candidate set $\mathcal{X}^{f}$
        \State Conduct $f$-fidelity experiment for $\mathcal{X}^{f}$ and collect the solution in matrix $\Y^{f}$  
        \State Update residual matrix  $\R^{f} = \Y^{f} - \Y^{f-1}_{\e_f} $
        \For{$|\mathcal{X}^{f}| < N_f$}
        \State Update $f$-fidelity residual GP model by MLE of \eqref{eq:likelihood f} based on $\mathcal{X}^{f}$, $\R^{f}$ 
        \State Find $\bxi_*$ based on Eq. \eqref{eq:var max f} for $\mathcal{X}^{f-1}  \setminus \mathcal{X}^{f}$
        \State Update $f$-fidelity candidate set $\mathcal{X}^{f} \leftarrow  \mathcal{X}^{f}\cup \{ \bxi_*\}$
        \State Conduct an $f$-fidelity simulation to obtain solution $\y^{f}(\bxi_*)$ 
        \State Update $f$-fidelity residual matrix $\R^{f} \leftarrow  [\R^{f},\y^{f}(\bxi_*)- \y^{f-1}(\bxi_*)]$
        \EndFor
        \EndFor
		\label{code1}
		\end{algorithmic}
\end{algorithm}
{The condition $\mathcal{X}^F \subset \mathcal{X}^{F-1} \subset \hdots \subset \mathcal{X}^{1}$ should in practice not be an issue since the experimental design can be chosen {\/\it a priori\/}. If for any reason this condition is not satisfied,   the  fidelity $f-1$ posterior GP
\begin{equation}
\y_l^{f-1}(\bxi) =\sum_{k=1}^{f-1}\r^k
\end{equation}
can be used to approximate  $\y_l^{f-1}(\bxi)$ at the inputs in $\mathcal{X}^f$, which can then used  to define the residual data $\R^{f}$ within the ResGP framework. In such a case, it may be beneficial to include the noise terms to account for the approximation error in the residual (possibly setting $\tau^1=0$). If we assume $\bOmega^f=\I$, this leads to
\begin{equation}
    \label{eq:rmgpnon}
    \begin{aligned}
    \y_l^{F}(\bxi) &= \sum_{f=1}^F\r_l^{f}(\bxi)\sim \N\left(\y_l^{F}(\bxi) \;\middle|\; \bmu^{F}(\bxi), \V^{F}(\bxi)\right), \\
     \bmu^{F}(\bxi) & =\sum_{f=1}^F (\k^{f}(\bxi))^T(\K^{f} + \tau^f \I)^{-1} \R^{f},\\
     \V^{F}(\bxi) &=\sum_{f=1}^F \I \otimes \left(k^{f}(\bxi,\bxi|\btheta^f)  - (\k^{f}(\bxi))^T(\K^{f} + \tau^f \I)^{-1} \k^{f}(\bxi)\right).
    \end{aligned}
\end{equation}
Such a model is still highly scalable with the output space dimension $d$, requiring only $F$ additional parameters compared to the basic \ours (see section \ref{compcomplex}). We do not consider this model in the examples. The efficacy of such an approach depends upon the disparity between the sets $\mathcal{X}^f$, and requires a deeper investigation that is beyond the scope of the present work.}
\subsection{Active learning via variance reduction}
\label{sec:auto}
In practice, data of high fidelity is expensive to obtain.
It is desirable to allocate computational resources, especially for the high-fidelity simulations, such that the surrogate model can achieve its best performance with the least computational cost. 
We first define the {\/\it information gain\/} for fidelity $f$ at a new parameter $\bxi_*$ as the uncertainty or variance given the current data collection $\overline{\mathcal{X}}$. Maximization of the information gain can then be defined by 
\begin{equation}
    \label{eq:var max f matrix}
    \bxi_*^{f} = \argmax_{\bxi \in \overline{\mathcal{X}}} \; \tr \left(\V^{f}_r(\bxi) \right),
\end{equation}
in which $\mbox{tr}$ denotes the trace operator. {The coregionalization matrices in Eq. \eqref{eq:rmgp noise free} are irrelevant as far as the information gain is concerned for any given $\bxi$  since
\begin{equation}
\displaystyle \begin{array}{ll}
 \tr \left(\V^{f}_r(\bxi) \right)& \displaystyle=\tr\left(\bOmega^{f} \otimes \left(k^{f}(\bxi, \bxi|\btheta^f) - (\k^{f}(\bxi))^T (\K^{f})^{-1} \k^{f}(\bxi)\right)\right)\vspace{2mm}\\
 \displaystyle & \displaystyle=\tr\left(\bOmega^{f}\right)  \left(k^{f}(\bxi, \bxi|\btheta^f) - (\k^{f}(\bxi))^T (\K^{f})^{-1} \k^{f}(\bxi)\right),
 \end{array}
 \end{equation}
and $\tr\left(\bOmega^{f}\right)$ is treated as a constant in the maximization.} Thus, setting $\bOmega^{f}={\bf I}$ does not affect our model either in terms of making predictions or in terms of utilizing the uncertainty for active learning or Bayesian optimization. {For decision making or active learning, the uncertainty related to a new sample (for a new input) rather than to a particular component of a new sample is essential.} Eq.~\eqref{eq:var max f matrix} now reduces to
\begin{equation}
    \label{eq:var max f}
    \bxi_*^{f} = \argmax_{\bxi \in \overline{\mathcal{X}}} \; \left(k^{f}(\bxi, \bxi|\btheta^f) - (\k^{f}(\bxi))^T (\K^{f})^{-1} \k^{f}(\bxi)\right).
\end{equation}

Inspired by the work of \citet{narayan2014stochastic}, we propose to build the multi-fidelity surrogate model in a sequential manner,  starting  from the lowest fidelity.
For each fidelity, based on the available data, we train the GP model and compute the information gain for each candidate input.
Subsequently, the fidelity $f$ experiments corresponding to the candidate inputs that yield  the maximum information gain are conducted and added to the training data. 
This process is repeated until a given condition is met.   
We present the full details of how to construct the model  without requiring the prior execution of low-fidelity simulations for all candidates (which is required by the classic stochastic collocation model) in Algorithm \ref{algo:greedy construct}.
In this algorithm, the stopping criteria is a given number of simulation runs for each fidelity, which should be decided based on the available computational budget.
In the case that we require the system to be fully automatic, we may instead specify a large candidate set $\mathcal{X}_t$ and  the uncertainty bound for determining the number of iterations. 
We can also perform an eigenvalue analysis of each correlation matrix to find the optimal number of samples that fully capture the model behaviour within the parameter space. 
{We note that there are more state-of-the-art active learning methods, such as the work of \citet{Song2019AGF}, which utilizes mutual information across different fidelities. The active learning component is not the main focus of this work, but such recent developments could improve the performance of ResGP. We also point out that parallelization of the training process is not straightforward if active learning is implemented.}

\begin{table}
	\centering
    \caption{{Model complexity comparison.}}
    \label{table:complexity}
	\small
	\begin{tabular}{l|ccl} 
        \hline
        \hline
        {Method} & {Complexity} & {Number of parameters} \\
        \hline
        {\ours with $\bOmega^{f}={\bf I}, \sigma^f=0$ } & {$\Ocal(\sum_{f=1}^F N^3_f)$}    & {$F(l+1)$}\\
                {\ours with $\bOmega^{f}= {\bf I}, \sigma^f\ne 0$ } & {$\Ocal(\sum_{f=1}^F N^3_f)$}      & {$F(l+2)$}\\
                {\ours with $\bOmega^{f}\ne {\bf I}, \sigma^f\ne 0$ } & {$\Ocal(\sum_{f=1}^F N^3_fd^3)$}   & {$F(l+2+d(d+1)/2)$}\\
       {NARGP } & {$\Ocal(\sum_{f=1}^F N^3_f)$}   & {$F(l+1)+(F-1)d$}\\ 
        {MF-DGP  } & {$\Ocal(F N_T M^2 d)$}  & {$(F-1)(Md(2+\frac{1}{2}(Md+1))+l(M+1)+1+d)+Md+l+1$} \\
        {AR1 } &{$\Ocal(N_T^3)$}  & {$F(l+2)-1$}\\
                {Greedy NARGP } &{$\Ocal(\sum_{f=1}^F N_f^3)$}   & {$(l+1)+(F-1)d$}\\
		\hline
    \end{tabular}
\end{table}

\subsection{Computational Complexity}\label{compcomplex}
{For standard GP models with $N$ training samples, the time (computational) complexity for model training is $\Ocal(N^3)$ due to the inversion of an $N\times N$ covariance matrix in the maximum log-likelihood solution (see Eq. (\ref{eq:likelihood f})) \citep{rasmussen2006gaussian}. Given its  structure, as the sum of conditionally independent GPs, \ours can scale well with  the output dimension. 
The computational complexity  is  that for the $F$ GPs with $N_f, f=1,\hdots,F$, training samples, namely $\Ocal(\sum_{f=1}^F N^3_f)$. 
The total number of model parameters  for \ours is $F(l+1)$ with a standard ARD kernel. This follows from the form of the ARD kernel (\ref{ARDkernel}), which requires $l+1$ hyperparameters for each $f$. With added noise, an additional hyperparameter $\sigma^f$ is required for each GP, leading to $F(l+2)$ hyperparameters. If $\bOmega^{f}\ne {\bf I}$, the covariance matrix is of size  $N_f d\times N_f d$ for each $f$, and the entries of each symmetric $d\times d$ matrix $\bOmega^{f}$ need to be estimated, leading to an additional $Fd(d+1)/2$ parameters. }

The computational complexity comparison with other state-of-the-art methods is shown in Table \ref{table:complexity} (for ARD kernels), in which $M$ is the number of inducing points for MF-DGP and $N_T=\sum_{f=1}^F N_f$ is the total number of training samples. {For AR, the covariance matrix is of size $N_T\times N_T$ \citep{kennedy2000predicting} [section 2.3], thus leading to the computational complexity  shown  in Table 1. The number of hyperparameters is $F(l+1)$ for $F$ ARD kernels, with an additional $F-1$ hyperparameters for constants defining the relationships between successive fidelities. NARGP involves $F$ GPs with  covariance matrices of sizes $N_f\times N_f$, leading to a computational complexity equal to that of \ours \citep{perdikaris2017nonlinear}. Since the outputs $\y^{f-1}(\bxi)$ are treated as inputs alongside $\bxi$ for fidelity $f=2,\hdots,F$, there are an additional $(F-1)d$ hyperparameters compared to \ours with $\bOmega^{f}={\bf I}, \sigma^f=0$. }

{Greedy NAR also involves $F$ GP training steps with $N_f\times N_f$ covariance matrices \citep{XING2020100012}. Like NAR, it uses the $f-1$ fidelity outputs as inputs for fidelity $f=2,\hdots,F$, but only involves the model inputs $\bxi$ in the lowest fidelity GP, leading to $(F-1)d+l+1$ hyperparameters for ARD kernels.  Since MF-DGP is equivalent to a deep GP with $F$ layers and outputs in  $d-$dimensional space in each layer, the computational complexity is dominated by the Kullback–Leibler divergence in the evidence lower bound (ELBO) for the variational approximation, namely $\Ocal(FN_TM^2d)$ \citep{salimbeni2017doubly}. The sparse variational approximation  in MF-DGP requires $Md(F-1)+(F-1)Md(Md+1)/2$ variational parameters for the means and symmetric covariance matrices of the distributions over  the inducing points (in $\mathbb{R}^d$) for fidelities $f=2,\hdots,F$, and a further $Md+(F-1)M(d+l)$ variational parameters for the inducing inputs at fidelities $f=1,\hdots,F$ \citep{cutajar2019deep}. The kernel hyperparameters for an ARD kernel are $l+1$ for fidelity 1 and $(F-1)(l+1+d)$ for the other fidelities, and a further $F$ hyperparameters are included for the noise variances. SC does not contain any model parameters and thus no model training is required. It only involves a single computation of the inverse of the Gram matrix at low fidelity \citep{narayan2014stochastic}. It does, however, require low-fidelity experiments in order to make predictions at high fidelity, which is usually far more costly.} 

The computational complexity for MF-DGP is prohibitive for high-dimensional problems, and AR also suffers from high costs along with \ours with $\bOmega^{f}\neq {\bf I}$. The numbers of parameters for NARGP, Greedy NAR and MF-DGP are also excessive, which can mean that training is problematic for large $d$. The same is true for \ours without the assumption $\bOmega^{f}= {\bf I}$; setting $\bOmega^{f}= {\bf I}$ and retaining the noise terms, however, leads again to a highly scalable model.

W-Xing:
I have sent earlier

\subsection{{Error bounds for the univariate ResGP}}

{In this section we will prove an error bound on ResGP in the univariate case. We begin with some definitions. A symmetric kernel function $k:\mathcal{X}\times \mathcal{X}\rightarrow \mathbb{R}$ is positive semi definite (psd)  if the corresponding matrix $[k(\bxi_n,\bxi_m)]_{nm}$ for any finite set $\{\bxi_1,\hdots,\bxi_N\}\subset \mathcal{X}\subset \mathbb{R}^l$ is psd. Henceforth, we consider only  kernels $k$ that are symmetric, psd and bounded on $\mathcal{X}$. A real-valued function $f:\mathcal{X}\rightarrow\mathbb{R}$ is Lipschitz continuous with  Lipschitz constant $L\ge 0$ if $|f(\bxi)-f(\bxi')|\le L\lVert\bxi-\bxi'\rVert$, $\forall \bxi,\bxi'\in \mathcal{X}\subset\mathbb{R}^l$, in which the standard Euclidean norm is used ($\lVert\cdot \rVert$ is used to denote a standard Euclidean norm throughout). We define a kernel $k$ to be Lipschitz continuous with a Lipschitz constant  $L^k$ in the sense that 
\begin{equation}
|k(\bxi,\overline{\bxi})-k(\bxi',\overline{\bxi})|\le A(\overline{\bxi})\lVert\bxi-\bxi'\rVert \le \sup_{\overline{\bxi}}A(\overline{\bxi})\lVert\bxi-\bxi'\rVert:=L^k\lVert\bxi-\bxi'\rVert,\quad \forall \bxi,\bxi',\overline{\bxi}\in \mathcal{X}.
\end{equation}
Most commonly used kernels, including the squared-exponential and the Matern class of kernels, are Lipschitz continuous in the sense defined above. A function $f:\mathcal{X}\rightarrow\mathbb{R}$ admits a monotonic function $\omega(\cdot)$ as a modulus of continuity iff $|f(\bxi)-f(\bxi')|\le \omega(\lVert\bxi-\bxi'\rVert)$, $\forall \bxi,\bxi'\in \mathcal{X}$. The $\tau$ covering number $M(\tau, \mathcal{X})$ of a set $\mathcal{X}$ 
is defined as the minimum number of open balls with radius $\tau$ (with respect to the standard Euclidean metric) that
is required to completely cover $\mathcal{X}$. We use $\lVert{\bf A}\rVert_2=\sup_{\lVert{\bf x}\rVert=1}\lVert{\bf A}{\bf x}\rVert$ to denote the matrix norm of ${\bf A}\in \mathbb{R}^{n\times m}$ induced by the standard Euclidean norm  in $\mathbb{R}^m$. }

{A number of results have been obtained in relation to error bounds for the simplest univariate GP models. Most of the bounds are derived based on the theory of reproducing kernel Hilbert spaces (RKHS) (a recent review can be found in  \citep{Motonobu}). For every psd kernel $k(\cdot,\cdot)$ there exists a unique RKHS $\mathcal{H}_k$, the functions $f$ in which inherit the smoothness properties of the kernel. This is easily seen by the following characterization of a RKHS: define an inner product space $\mathcal{V}$  
\begin{equation}
\mathcal{V}=\left\{f: f(\cdot)=\sum_{n=1}^\infty a_nk(\cdot,\bxi_n),\quad  a_n\in\mathbb{R},\;\bxi_n\in\mathcal{X}\right\} 
\end{equation}
with inner product $\langle f,g\rangle_{\mathcal{H}_k}=\sum_n\sum_ma_nb_mk(\bxi_n,\bxi_m')$, where $g(\cdot)=\sum_m b_mk(\cdot,\bxi_m')$, 
and  induced norm $\lVert f\rVert_{\mathcal{H}_k}=\langle f,f\rangle_{\mathcal{H}_k}^{1/2}$. $\mathcal{V}$ is a pre-Hilbert space from which we obtain the unique RKHS as ${\mathcal{H}_k}=\{f\in\overline{\mathcal{V}}:\lVert f\rVert_{\mathcal{H}_k}<\infty\}$, where the closure is defined with respect to the metric induced by $\lVert \cdot\rVert_{\mathcal{H}_k}$.
For the squared exponential and other common kernels, ${\mathcal{H}_k}\subset C^{\infty}(\mathcal{X})$, while for Matern kernels,  $\lVert \cdot\rVert_{\mathcal{H}_k}$ is norm-equivalent to a Sobolev space $W^{s,2}(\mathcal{X})$, $s>l/2$. In general, therefore, RKHSs are quite restrictive, and are small compared to the support of a prior GP distribution with the covariance function (kernel) that defines the RKHS.}

{Bounds can be derived for the GP regression error using the equivalence between GP regression and stationary kernel interpolation \citep{10.1090/S0025-5718-99-01009-1}, the latter of which can be posed as an optimization problem in the RKHS $\mathcal{H}_k$ corresponding to the interpolation kernel $k$ \citep{Motonobu}. If the true function in GP modelling is hypothesised to lie in this space, the results can be carried over. The bounds are given in terms of a power function, which is identified with the posterior variance of an equivalent GP model. For noisy data, analogous results for kernel ridge regression can be used, in which case the  error bounds depend on
 the norm $\lVert f\rVert_{\mathcal{H}_k}$ of the unknown function $f$, as well as an empirical covering number with respect to the $L^2(\mu)$ norm, where $\mu$ is the unknown distribution over the data   \citep{10.1109/TIT.2002.1013137}. Again for noisy observations, information theoretic and RKHS approaches were  used by \citet{6138914},  later improved upon by \citet{pmlr-v70-chowdhury17a}, to find error bounds for GP regression. These bounds involve constants that in practice are difficult to obtain. 
An alternative hypothesis is to take the support of the prior distribution of the GP as the belief space from which to seek the true function. This hypothesis  has been employed in stochastic bandit problems based on GPs \citep{10.5555/3042573.3042697,doi:10.1080/01621459.2019.1598868} and more recently has been used to establish general interpretable  bounds for basic GP models \citep{lederer2019uniform,lederer2021uniform}. The sample space is the largest possible space of candidate functions, and leads to bounds that can be approximated for common settings with relative ease, in comparison to the RKHS approaches. In the analysis below, we use the interpolation (noise-free) results of \citet{doi:10.1080/01621459.2019.1598868}, and, in particular, follow closely the analysis  of \citet{lederer2019uniform} to extend their univariate, single-fidelity probabilistic uniform error bound to \ours. Such a bound is defined as follows
\begin{definition}\label{defGP}
		A GP estimate $\mu(\bxi)$ of an unknown  function $y(\bxi)$ has a uniformly bounded error 
		on a compact set~$\mathcal{X}\subset\mathbb{R}^l$  if there exists a function $g(\bxi)$ such that 
		\begin{align}
	|\mu(\bxi)-y(\bxi)|\leq g(\bxi),\quad \forall \bxi \in\mathcal{X}.
		\end{align}
		If this bound holds with probability of at least $1-\delta$ for some 
		$\delta\in (0,1)$, it is called a probabilistic uniform error bound.
	\end{definition}
\noindent As stated above, we require the following main assumption over the  unknown function, in this case $y^F(\bxi)$, namely that it belongs to the sample space of the prior GP. 
\begin{assumption}\label{assumpGP}
$y^F(\bxi)$ is a realisation (is in the sample space) of the following GP 
\begin{equation}\label{priorass}
\mathcal{GP}\left(y^{F}(\bxi)\mid 0, \sum_{f=1}^F k^{f}(\bxi,\bxi'|\btheta^{f}) \right),
\end{equation}
and observations $y_n^F$ are evaluations of this function at the design points, that is $y_n^F=y^{F}(\bxi_n)$,  $\forall \bxi_n\in\mathcal{X}^F$.
\end{assumption}
The properties of sample paths  of this GP (the belief space) are again closely related to the smoothness of the kernel, but are more difficult to establish and quantify. For stationary kernels $k(\bxi-\bxi')$,  almost surely (a.s.) or sample path continuity is ensured by sufficient smoothness of $k$ at the origin \citep{adler2009random} (section 2.5). Similar  continuity results can be established for derivatives of the sample paths, essentially requiring 2$k$ times differentiability of the kernel to establish $k$ times differentiability of sample paths \citep{adler2009random} (section 2.5.2). Although Lipschitz continuity is a rather strong form a continuity,  implying amongst other things uniform continuity,  this assumption is still rather mild in comparison to the RKHS hypothesis. The following Lemma (see Appendix \ref{proofs} for the proof) enables us to prove the main result on the uniform error bound for the ResGP approximation of $y^F$.}
{
	\begin{lem}
		\label{th:errbound_with}
		Consider the posterior GP process defined by \ref{priorass}, in which all of the  
		kernels $k^{f}(\bxi, \bxi'|\btheta^{f})$ are assumed to be   Lipschitz continuous with Lipschitz constants $L_k^f$ 
		on the compact sets $\mathcal{X}^f\subset \mathcal{X}$, $f=1,\hdots,F$. Furthermore, consider a continuous unknown 
		function $y^F:\mathcal{X}\rightarrow \Rset$ with Lipschitz constant $L_y$ 
		and $N_F$ observations $y_n^F$ satisfying Assumption \ref{assumpGP}. Then the 
		posterior mean ~$\mu^F(\cdot)$ defined in (\ref{eq:rmgp noise free}) and the standard deviation 
		$\sigma^F(\cdot)=\sqrt{v^F(\bxi)}$ (Eq. (\ref{varwithout}))
		of the GP conditioned on $\{(\bxi_n,y_n^F)\}_{n=1}^{N_F}$
		are both continuous, with Lipschitz constant $L_{\mu^F}$ and modulus of continuity
		$\omega_{\sigma}^F(\cdot) $ on $\mathcal{X}$, respectively, satisfying 
		\begin{align}
		\label{eq:L_nu}
		L_{\mu^F}&\leq \sum_{f=1}^FL^f_k \sqrt{N_f}\lVert (\K^{f})^{-1}\R^{f}\rVert \\
		\omega_{\sigma}^F(\lVert\bxi-\bxi'\rVert)&\leq\left[\sum_{f=1}^F2 L^f_k\left(1
 +N_f\max\limits_{\bxi,\bxi'\in\mathcal{X}}k^{f}(\bxi, \bxi'|\btheta^{f})\lVert(\K^{f} )^{-1}\rVert_2\right)\lVert\bxi-\bxi'\rVert\right]^{1/2}.
		\label{eq:omega_sigma}
		\end{align}
	\end{lem}
		\begin{rem}
	Any kernel {that is} everywhere differentiable and has bounded partial derivatives is  Lipschitz continuous and any Lipschitz constant satisfies $L_k^f\le \sup_{\bxi,\bxi'\in\mathcal{X}}\|\nabla_\bxi k(\bxi,\bxi')\|_\infty$, where $\|\cdot\|_\infty$ is the $l^{\infty}$ norm. The Lipschitz constant $L_y$ is clearly not available in practice. However, a probabilistic bound can be obtained as in Theorem 3.2 in \citep{lederer2019uniform}. The remaining terms in these expressions depend only on the training data and 
	kernels, which are explicitly known. Using Cauchy-Schwartz and the fact that $\lVert\cdot\rVert_2$ is subordinate to $\lVert\cdot\rVert$, the bound for $L_{\mu^F}$ can be elaborated further as follows
\begin{equation}
\sum_{f=1}^FL^f_k \sqrt{N_f}\lVert (\K^{f})^{-1}\R^{f}\rVert \le \sum_{f=1}^FL^f_k \sqrt{N_f}\frac{\lVert\R^{f}\rVert}{\rho_m(\K^{f})},
\end{equation}
in which $\rho_m(\K^{f})=\lVert(\K^{f} )^{-1}\rVert_2^{-1}$ is the smallest singular value of $\K^{f}$. $\rho_m(\K^{f})>0$ $\forall f$ since $\K^{f}$ is psd. 
	\end{rem}}
{The error on any design $\mathcal{X}_{\tau}^F$ is bounded as  in \citet{6138914} [Lemma 5.1], introducing a constant $\beta(\tau)$, which depends on an upper bound $\tau$ for the fill distance of $\mathcal{X}_{\tau}^F$, defined as $h=\sup\limits_{\bxi\in\mathcal{X}} \min\limits_{\bxi'\in\mathcal{X}_{\tau}^F}
	\|\bxi-\bxi'\|$. $h$ represents the radius of the largest ball in $\mathcal{X}$ that does not
contain any point in the design $\mathcal{X}_{\tau}^F$. In  greedy designs that minimise the GP posterior variance (as in the active learning component), for any
kernel $k$ that induces a RKHS that is norm equivalent to $W^{s,2}(\mathbb{R}^l)$ with $s>l/2$, \citet{santin2017convergence}   showed that $\forall \epsilon>0$, $\exists C_{\epsilon}>0$ such that $h\le C_{\epsilon}|\mathcal{X}_{\tau}^F|^{-l/2+\epsilon}$, in which $|\mathcal{X}_{\tau}^F|$ is the cardinality of $\mathcal{X}_{\tau}^F$.  \\
\indent The minimum number of grid points satisfying the upper bound on the fill distance is $M(\tau,\mathcal{X})$, an upper bound for which is $M(\tau,\mathcal{X})\leq (1+e/\tau)^d$,  assuming a hypercubic set 
	$\mathcal{X}$  
with edge length $e$. By utilizing Lemma \ref{th:errbound_with} and the continuity of $y^F$, the error bound on $\mathcal{X}_{\tau}^F$ can then be extended to the whole of  $\mathcal{X}$, which leads to Theorem \ref{th:errbound}  below
	\begin{theorem}[\citet{lederer2019uniform},Theorem 3.1]
		\label{th:errbound}
		Let the conditions of Lemma \ref{th:errbound_with} be satisfied. Pick $\delta\in (0,1)$, $\tau\in\Rset$, $\tau>0$, and set 
		\begin{align}
		\label{eq:beta}
		\beta(\tau)&=2\log\left(\frac{M(\tau,\mathcal{X})}{\delta}\right)\\
		\gamma(\tau)&=\left( L_{\mu^F}+L_y\right)\tau+\sqrt{\beta(\tau)}\omega_{\sigma}^N(\tau).
		\label{eq:gamma}
		\end{align}
		Then, it holds that
		\begin{align}
		\label{eq:errorbound}
		\mathbb{P}\left(|y^F(\bxi)-\mu^F(\bxi)|\leq 
		\sqrt{\beta(\tau)}\sigma^F(\bxi)+\gamma(\tau), \forall \bxi\in \mathcal{X}\right)\geq 1-\delta.
		\end{align}
	\end{theorem}
		\begin{rem}	
		\noindent Eq.  
	\eqref{eq:errorbound} can be computed for fixed $\tau$ and $\delta$ given the probabilistic upper 
	bound for  $L_y$. We note that $\beta(\tau)$ grows only  
	logarithmically as $\tau$ decreases, which limits the growth of $\gamma(\tau)$ as $\tau\rightarrow 0$.
\end{rem}
}

\section{Results and discussion}
\label{sec:exp}
\newcommand{\angstrom}{\textup{\AA}}

\subsection{Test problem 0: double pendulum}
We first examine an ODE test case  to illustrate the capacity of  \ours to handle complex nonstationary problems. 
We consider a rigid pendulum problem with two masses. 
The full system describing the evolutions of angles $\theta_1$ and $\theta_2$ is given by
\begin{equation}
	\begin{aligned}
		(m_1 + m_2)l_1 \ddot{\theta}_1 &+ m_2 l_2 \ddot{\theta}_2 \cos(\theta_1-\theta_2) \\ 
		 & + m_2 l_2 (\dot{\theta}_2)^2 \sin(\theta_1-\theta_2)
		+g(m_1+m_2) \sin \theta_1 = 0\\
		m_2 l_1 \ddot{\theta}_1 \cos(\theta_1-\theta_2) m_2 l_2 \ddot{\theta}_2
		&-m_2 l_1 (\dot{\theta}_1)^2 \sin(\theta_1-\theta_2)
		+m_2 g \sin(\theta_2) =0,
	\end{aligned}
\end{equation}
where the dots denote time derivatives, $g$ is the acceleration due to gravity, and $l_1$ and $l_2$ are the lengths of the rods connecting the masses $m_1$ and $m_2$.
The double pendulum system is highly nonlinear and exhibits chaotic motion. It is very sensitive to the initial state, and is thus a very challenging test for any method. We use $\theta_1(0) \in [1.25, 1.57]$ rad as the problem input  and solve the dynamical system to obtain the  solutions $\theta_1(5)$ and $\theta_2(5)$, which are used as the quantities of interest. 
For the remaining  system parameters, we used $\theta_2(0)=2.2$ rad, $l_1=1$ m, $l_2=2$ m, $m_1=2$ kg and $m_2=1$ kg. The system was solved using a five-stage, fourth-order explicit Runge-Kutta scheme with a time step of $\Delta t=0.1s$ for the low-fidelity solutions and $\Delta t=0.01s$ for the high-fidelity solutions.

In order   to predict the F2 ground truth response curve, we trained \ours and NARGP with 41  low-fidelity (F1) observations at equally spaced inputs and 14  high-fidelity (F2) observations, also at equally spaced inputs that formed a  subset of the F1 inputs.  The two scalar quantities of interest were learned independently ($d=1$)  using both methods. 
The results for the predictions of $\theta_1(5)$ and $\theta_2(5)$ are shown in Figs.~\ref{fig:pend1} and \ref{fig:pend2}, respectively. 
It can be seen clearly  that both methods make good overall predictions of the F2 curves, although  \ours performs noticably better  in certain regions (\eg for the case of $\theta_1(5)$, in the parameter range  $1.3<\theta_1(0)<1.5$).
The most important conclusion from  Figs. \ref{fig:pend1} and \ref{fig:pend2} is that the predictive posterior of \ours reflects the model uncertainty well, whereas NARGP is generally over-confident in its predictions. This can make the use of NARGP problematic for applications such as design optimization and risk management, where accurate model uncertainty is crucial.

\begin{figure}
	\centering
	\includegraphics[width=0.45\textwidth]{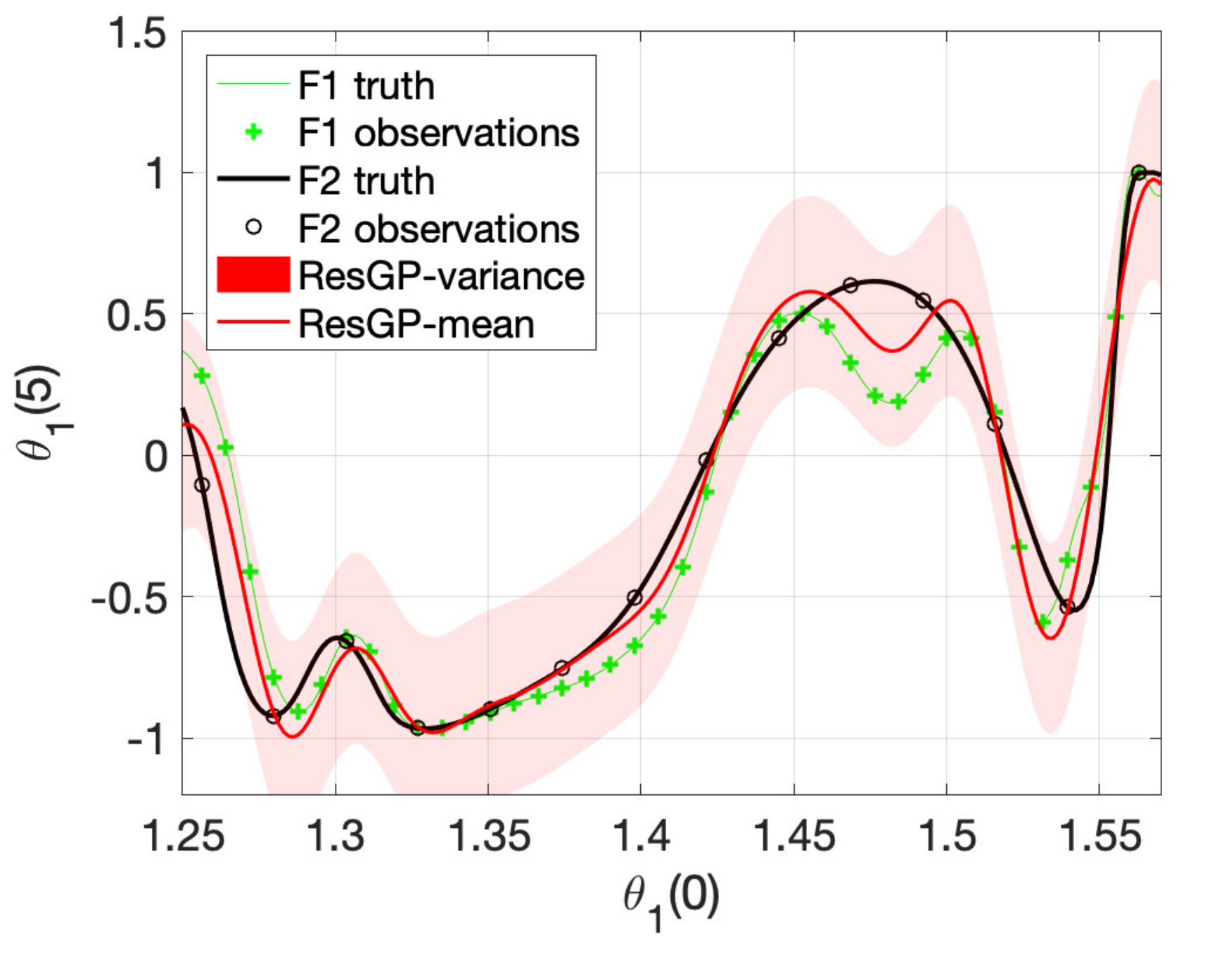}
	\includegraphics[width=0.45\textwidth]{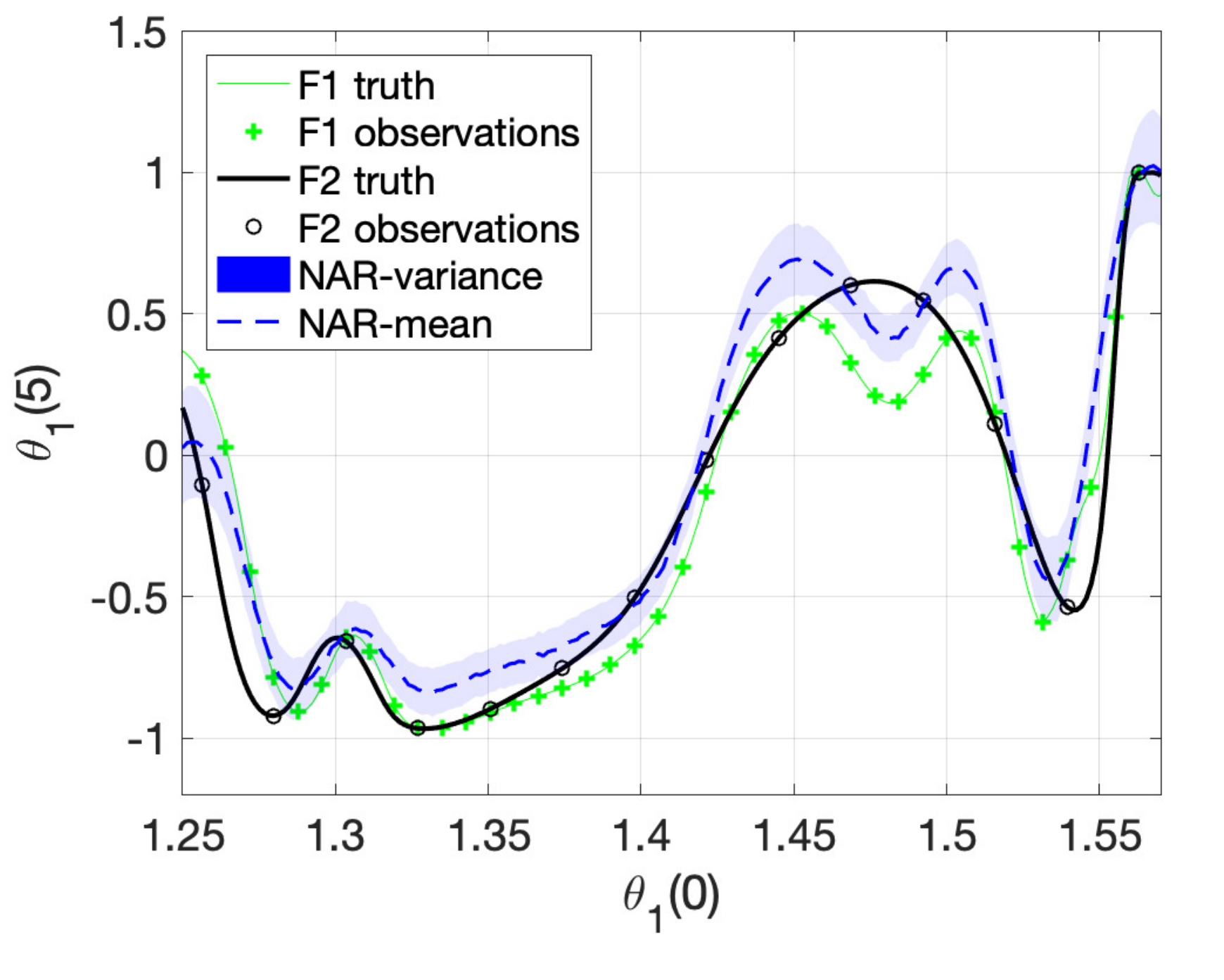}
	\caption{Predictions of $\theta_1$ at $t=5$ s given the initial condition $\theta_1(0)$ for a double pendulum system.}
	\label{fig:pend1} 
\end{figure}

\begin{figure}
	\centering
	\includegraphics[width=0.45\textwidth]{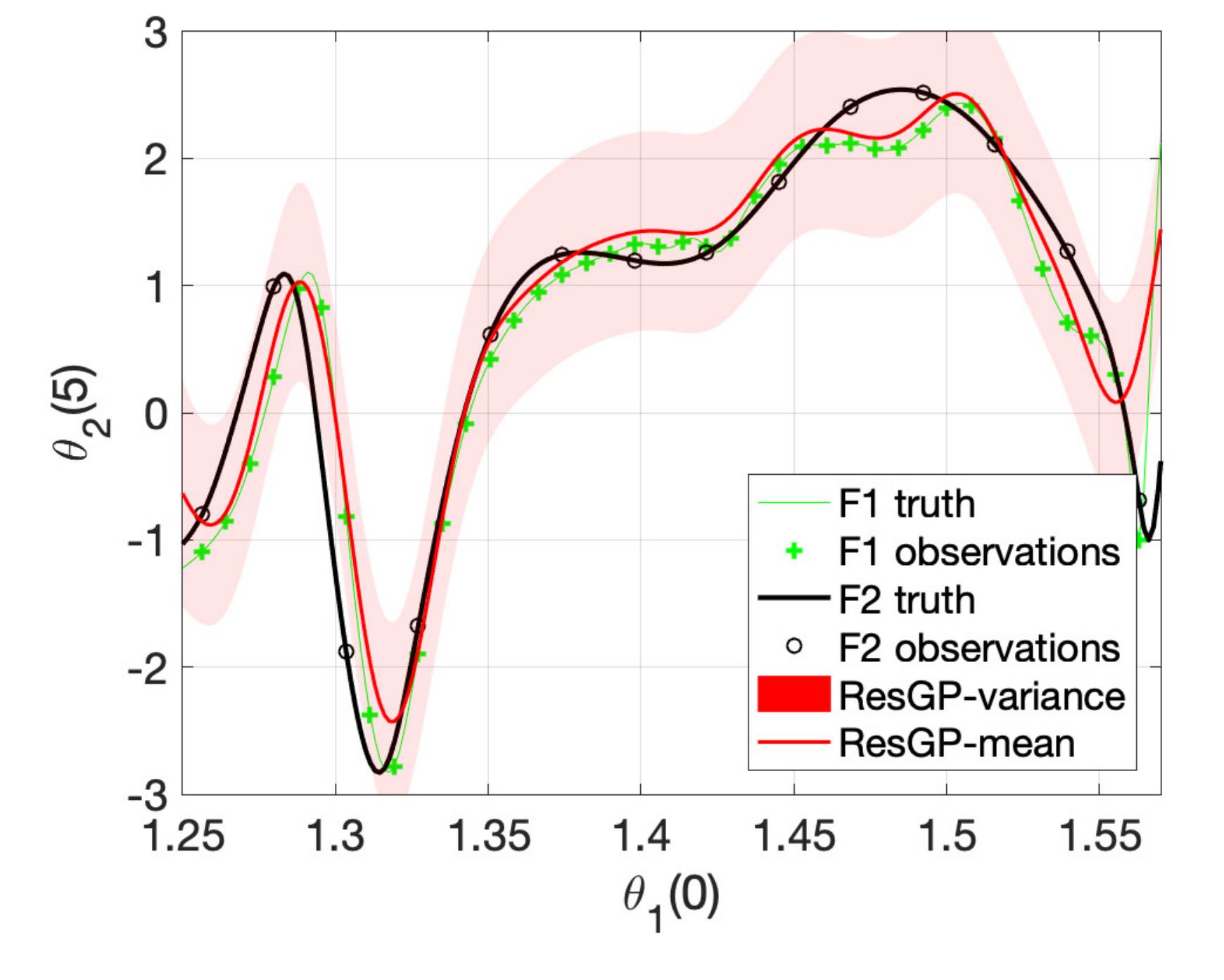}
	\includegraphics[width=0.45\textwidth]{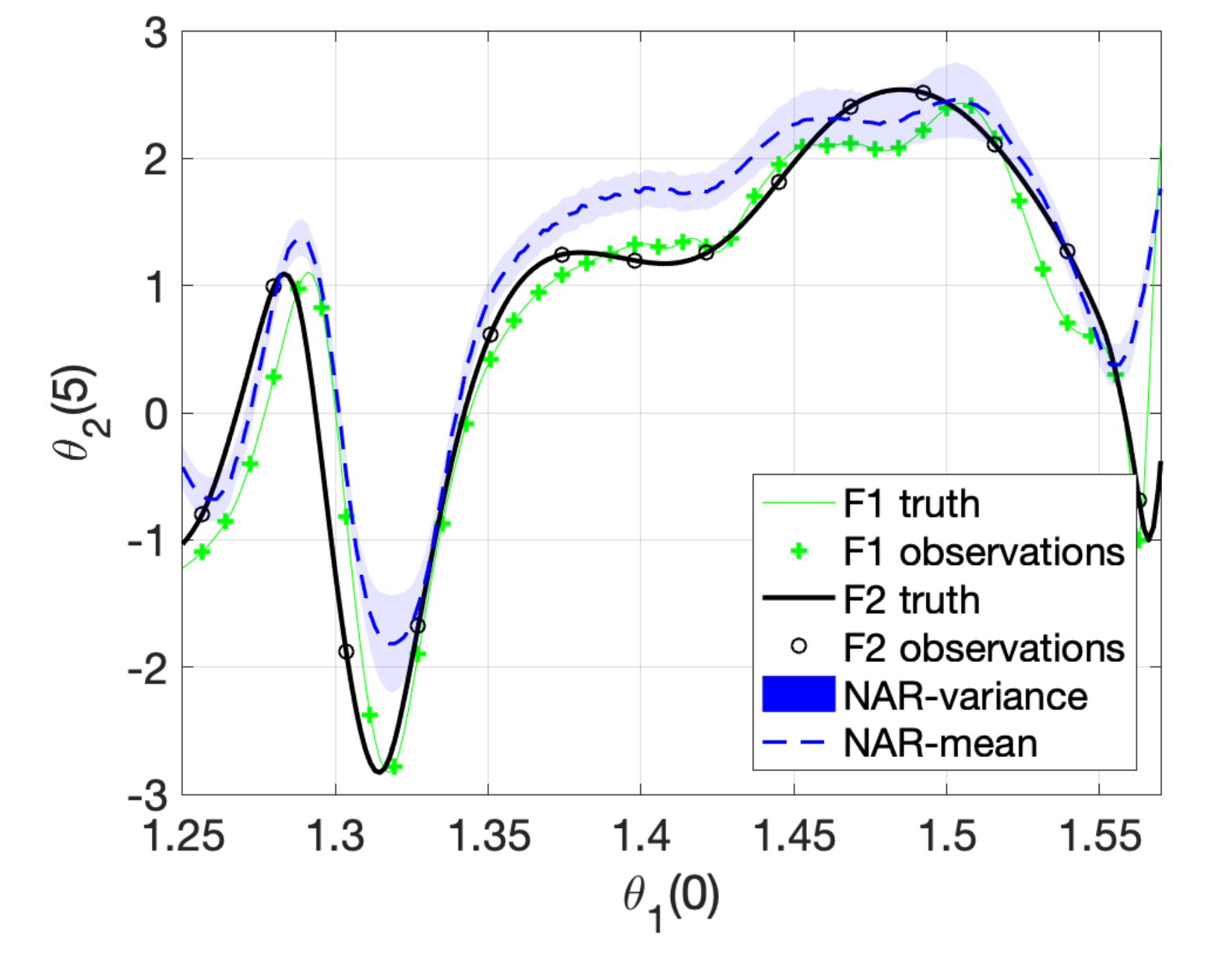}
	\caption{Predictions of $\theta_2$ at $t=5$ s given the initial condition $\theta_1(0)$ for a double pendulum system.}
	\label{fig:pend2} 
\end{figure}

{In the results presented below, AR, NAR, MF-DGP, Greedy NAR and SC are implemented as per their original formulations.  In all of the GP methods, zero-mean functions are assumed by centering the data, i.e., no regression functions are employed. With the exception of  MF-DGP, noiseless data is assumed for all fidelities. The original MF-DGP includes noise at all fidelities. For \ours, AR and Greedy NAR, ARD kernels are used. For NARGP, the fidelity-1 kernel is an ARD kernel, while the fidelity $f=2,\hdots,F$ kernels are of the form introduced by Perdikaris et al.  in the original implementation  \citep{perdikaris2017nonlinear} [Eq. (2.12)]. Each of these kernels  is formed from three ARD kernels by factoring the dependence on $\bxi^f$ and $\y^{f-1}(\bxi)$ and adding a third kernel as a  bias term. In  MF-DGP, enhanced versions of these kernels (adding an additional linear kernel for the dependence on $\y^{f-1}(\bxi)$) are employed as in \citet{cutajar2019deep} [Eq. (11)]. For MF-DGP we used the authors' open source code \citep{emukit2019}, which is available on Github\footnote{\url{https://github.com/EmuKit/emukit/tree/master/emukit/examples/multi_fidelity_dgp}}. For all other methods, we used our own implementations with the settings stated above to generate the results.} 

\subsection{Test problem 1: five univariate benchmark problems}\label{Examp1}
{
\begin{table}[h!]
	\centering
	\caption{	\label{table2}{Comparisons of the root mean square error (RMSE), R$^2$ and mean negative log likelihood (MNLL) for \ours and three state-of-the-art methods on five univariate multi-fidelity  benchmark problems. The number of training samples is  given in the form $N_1$-$N_2$ or $N_1$-$N_2$-$N_3$. $^*$Values were taken from \citep{cutajar2019deep} {[Table  1]}.}}
	\resizebox{\textwidth}{!}{
	\begin{tabular}{|c|c|c|c|c|c|c|c|c|c|c|c|c|c|c|} 
		\hline
		& \multicolumn{1}{c}{}      &                     & \multicolumn{3}{c|}{AR1}                                       & \multicolumn{3}{c|}{NARGP}                                      & \multicolumn{3}{c|}{MF-DGP$^*$}                                       & \multicolumn{3}{c|}{ResGP}          \\ 
		BENCHMARK   & \multicolumn{1}{c}{$l$} &$N_f$ & \multicolumn{1}{c}{R\^{}2} & \multicolumn{1}{c}{RMSE} & MNLL   & \multicolumn{1}{c}{R\^{}2} & \multicolumn{1}{c}{RMSE} & MNLL    & \multicolumn{1}{c}{R\^{}2} & \multicolumn{1}{c}{RMSE} & MNLL     & \multicolumn{1}{c}{R\^{}2} & \multicolumn{1}{c}{RMSE} & MNLL    \\ 
		\hline
		CURRIN      & 2                         & 20-5                & 0.918                      & 0.564                    & 11.136 & 0.947                      & 0.550                    & 14.364  & 0.935                      & 0.601 & 0.763 &  \bf{0.967}                      & \bf{0.436}                    & \bf{0.663}   \\ 
		\hline
		PARK        & 4                         & 30-5                & 0.986                      & 0.552                    & 57.397 & 0.965                      & 0.883                    & 103.891 & 0.985                      & 0.565                    &1.383  & \bf{0.990}                      & \bf{0.397}                    & \bf{0.290} \\ 
		\hline
		BOREHOLE    & 8                         & 60-10               & 0.999                      & 0.012                    & -2.886 & 0.998                      & 0.019                    & -2.668  & 0.999                      & 0.015 & -2.031   & \bf{1.000}                      &\bf{0.005}                    & \bf{-3.996}  \\ 
		\hline
		BRANIN      & 2                         & 80-30-10            & 0.912                      & 0.019                    & \bf{-4.019} & 0.724                      & 0.107                    & 5.487   & 0.965                    & 0.030                    & -2.572 & \bf{0.998}                      & \bf{0.008}                    & -3.675   \\ 
		\hline
		HARTMANN-3D & 3                         & 80-30-10            & 0.996                      & 0.058                    & -1.414 & 0.992                      & 0.083                    & -0.712  & 0.994                      & 0.075                    & -0.731    & \bf{0.997}                      & \bf{0.054}                    & \bf{-1.496}  \\
		\hline
	\end{tabular}}
\end{table}}
{We first examine the univariate case and compare the results to AR, NARGP and MF-DGP. A  selection of well-known univariate multi-fidelity problems is used, following \citep{cutajar2019deep}. This includes three 2-fidelity and two 3-fidelity examples. The functions considered and the definitions of the fildelities  are detailed in Appendix \ref{synth}. A comparison of the accuracy of ResGP (without active learning) with the other methods is shown in Table 2. This table includes values of  the root mean square error (RMSE), the coefficient of determination ($R^2$) and the mean negative log likelihood (MNLL) against 1000 randomly selected test points, following  \citep{cutajar2019deep}. The MNLL provides the most commonly accepted test for the capture of uncertainty. The number of training points for each fidelity $N_f$ and the number of inputs $l$ (selected randomly)  are also given in Table \ref{table2}.}

{As can be seen, ResGP outperforms all other methods in all examples in terms of the accuracy. {For example, the RMSE is at least 21\%, 28\%, 58\%, 58\% and 7\% lower than for the other methods on the Currin, Park, Borehole, Branin and Hartmann-3d examples, respectively. The MNLL is at least 13\%, 79\%, 38\% and 6\% lower than for the other methods on the Currin, Park, Borehole and Hartmann-3d examples, respectively. In the Branin example,  the MNLL is 9\% more negative for AR.} Given their high model flexibilities, it is possible  that MF-DGP and NARGP tend to overfit the data and underestimate the uncertainty in these examples. For \ours, the uncertainties have an additive structure and increase monotonically. If the scale of the residual is large or the sample number small, the uncertainty will likewise be large. In this sense \ours is more `careful' with uncertainty estimations (in some problems it may overestimate the uncertainty). }

We note that we were unable to reproduce the numbers stated  in \citep{cutajar2019deep} using the authors' code \citep{emukit2019}. Thus, the values appearing in Table 1 for MF-DGP are taken from the original paper \citep{cutajar2019deep} {[Table 1]}. For all other methods we used the training and test data  provided by the authors on Github$^1$ (which they used to generate the results in \citep{cutajar2019deep}).

\subsection{Test problem 2: turbulent mixing flow in an elbow-shape pipe}

A number of models have been developed to study turbulent flows, ranging  from simple one-equation models such as the Sparllat-Almaras model to  sophisticated models such as Large Eddy Simulation (LES). The former can be considered  low-fidelity whereas the latter can be considered high-fidelity. For the design and optimization of thermal-fluid systems, high-fidelity models can be impractical. On the other hand, low-fidelity models will lead to sub-optimal designs. Thus, turbulence modelling is prime candidate for combining  low- and high-fidelity models.

{We applied  \ours with and without active learning to a benchmark problem of turbulent mixing flow in a pipe and  compared it with three state-of-the art  methods, namely NARGP, SC (without out-of-sample F1 data, for a fair comparison) and Greedy NAR}. As illustrated in Fig.~\ref{fig2}, water enters from two inlets,  the bottom left end of the pipe and a smaller inlet located on the elbow. The water exits the pipe vertically upwards from the top right. The input parameter space was chosen to be the freestream velocity at the large inlet (with  a diameter of 1 m), taking values between 0.2 to 2 m s$^{-1}$, and the freestream velocity at the smaller inlet (with a diameter 0.5 m), taking values between 1.2 to 3 m s$^{-1}$. The quantities of interest were vectorised profiles at $t=50$ s  of the velocity magnitude  and pressure  in  circular cross sections of the elbow pipe, one located at the elbow  junction (oriented at 45 degrees) and the other near the pipe exit (oriented at 0 degrees). In all cases, the profiles contained $96$ values. All  multi-fidelity methods were applied separately to each quantity of interest, so that $d=96$ in all cases. 
  \begin{figure}[H]
	\centering
    \includegraphics[width=0.49\textwidth]{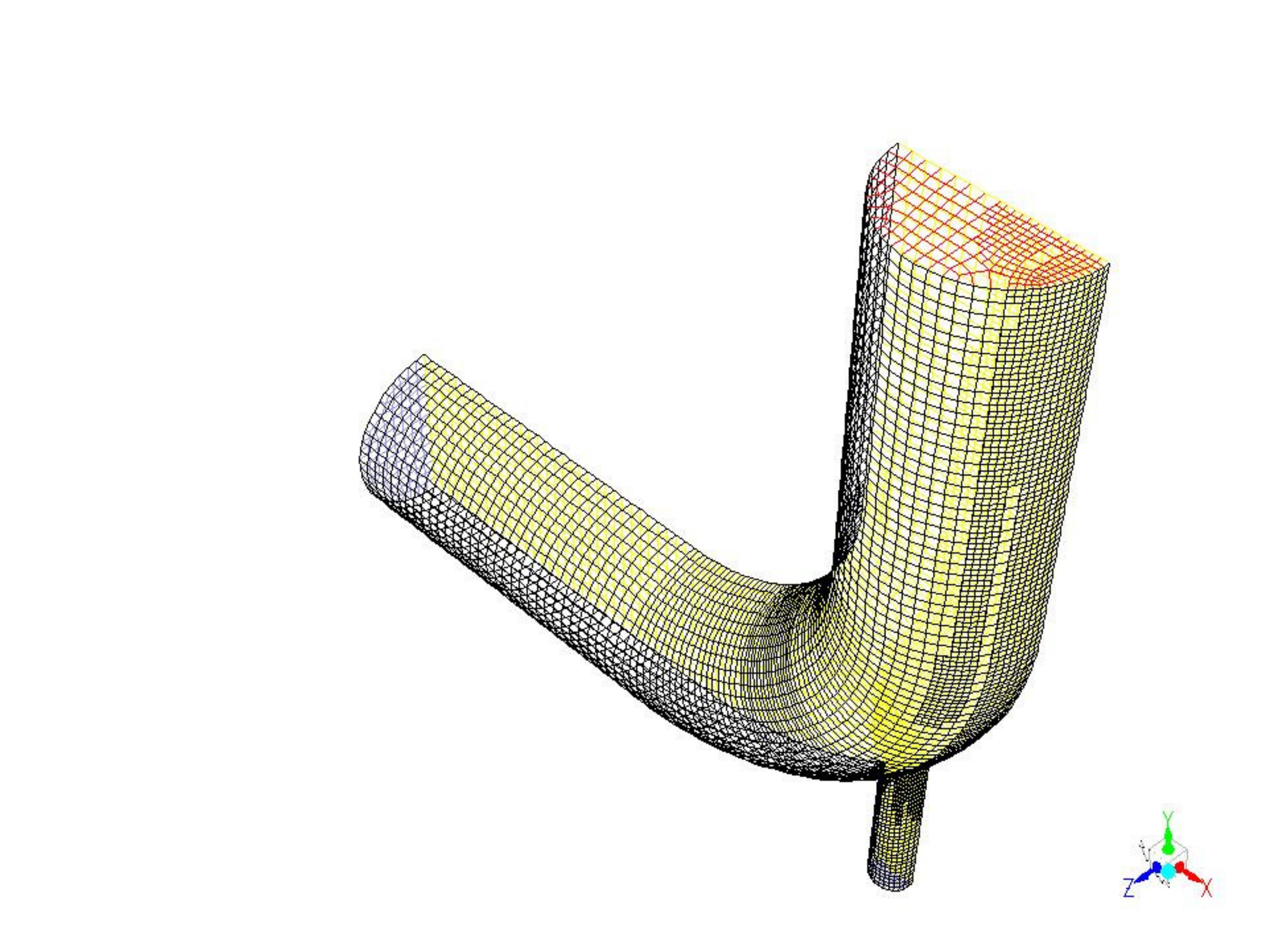}
    \includegraphics[width=0.49\textwidth]{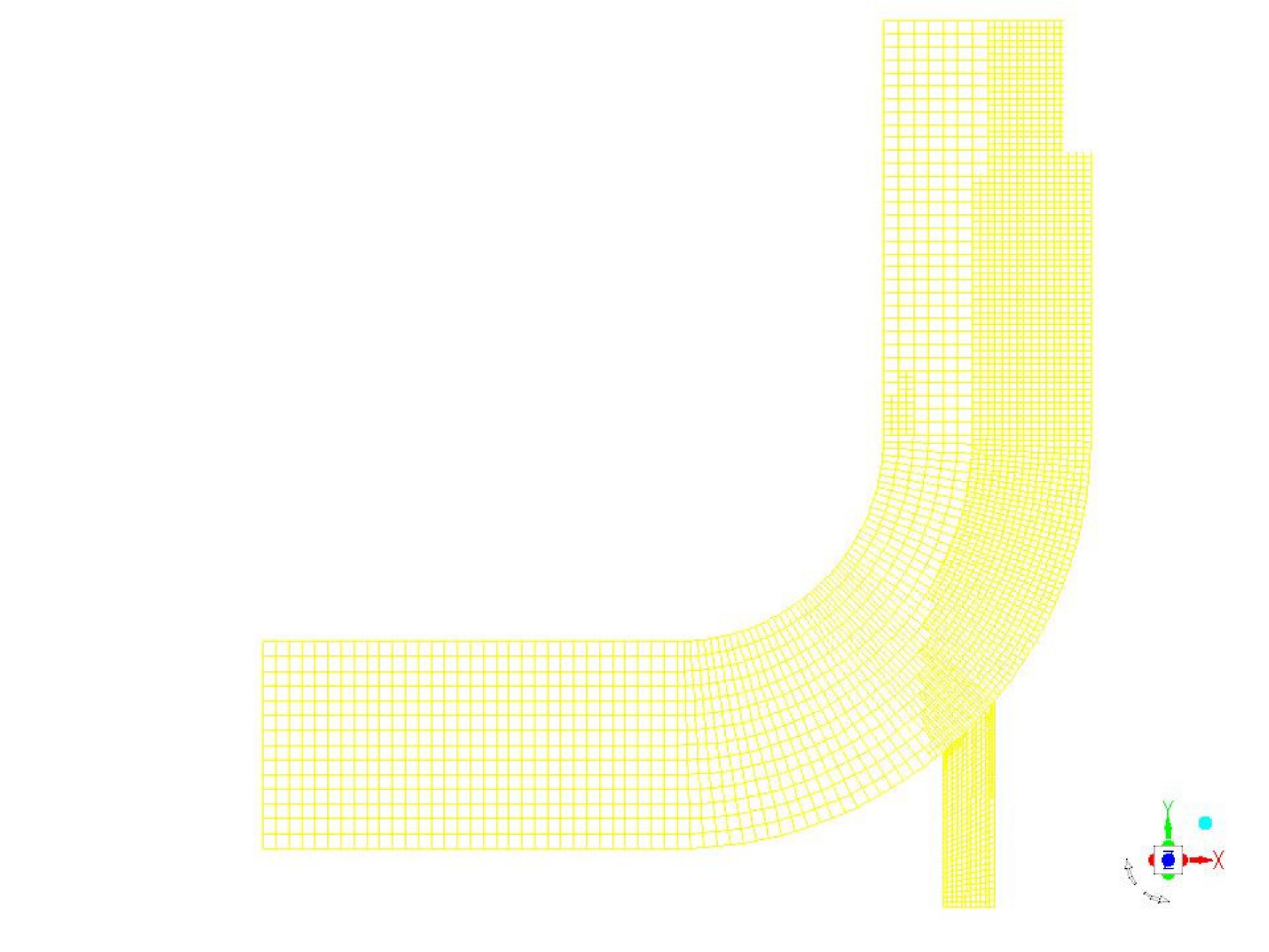}
	\caption{Computational domain and geometrical configuration of an elbow-shaped pipe and its plane of symmetry (Test Problem 2).}
	\label{fig2}
\end{figure}
The high-fidelity model (F2)  was defined as LES while the low-fidelity model (F1) was defined as Sparlart-Allmaras. Both were   implemented in ANSYS Fluent. The LES model employed the  dynamic kinetic Energy subgrid-scale model. For the Sparlart-Allmaras model, vorticity-based production together with low-Reynold's number damping were selected. Default schemes in FLUENT were used for both F1 and F2 models, i.e., second-order implicit time stepping and central differencing for the finite-volume discretization. The meshes contained a total  of 
36134 nodes and 29399 hexahedral cells, and the time step was set to 0.01 s.

We assess the performance of all multi-fidelity models using a normalized root mean square error (NRMSE) for each quantity of interest. The NRMSE for $N$ test points is defined as follows
\[
    \mbox{NRMSE} = \sqrt{ \frac{ \sum_{n=1}^{N} \sum_{j=1}^d (y_{nj} - \hat{y}_{nj})^2 }{\sum_{n=1}^{N} \sum_{j=1}^d \hat{y}_{nj}^2}}
\]
in which $y_{nj}$ and $\hat{y}_{nj}$ denote the $j$-th coefficients of the $n$-th high-fidelity prediction and the $n$-th ground-truth (test) point, respectively.
\begin{figure}[h]
	\centering
	\begin{subfigure}[b]{0.85\linewidth}
		\centering
		\includegraphics[width=1\textwidth]{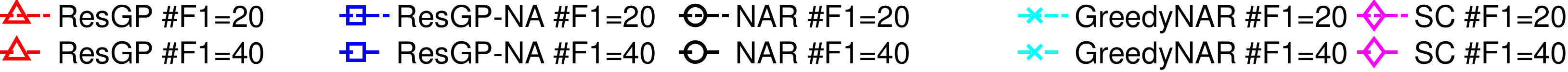}
	\end{subfigure}\\
	\begin{subfigure}[b]{0.49\linewidth}
		\includegraphics[width=1\textwidth]{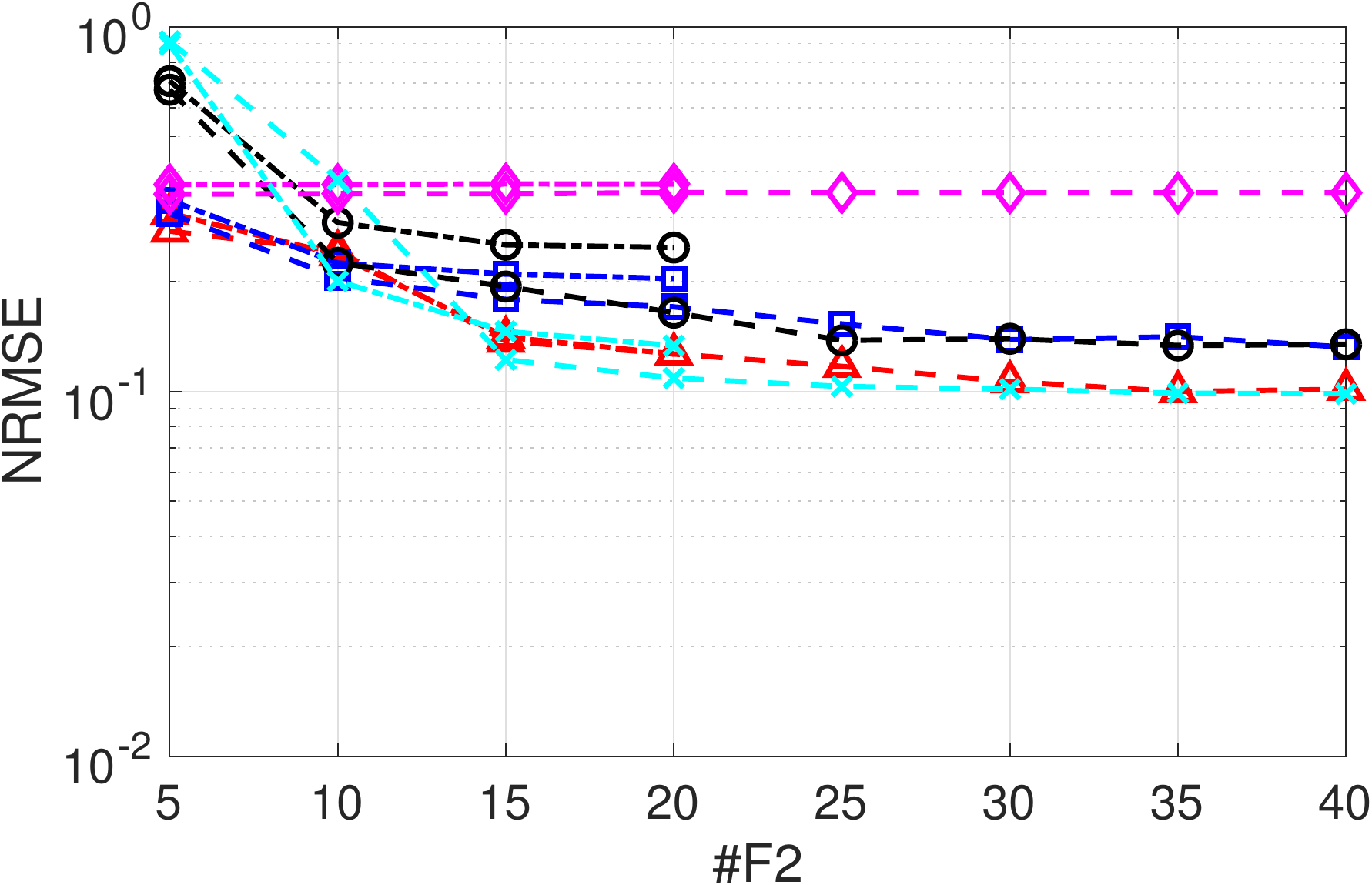}
		\caption{Pressure profile  near the pipe exit.}
	\end{subfigure}
	\begin{subfigure}[b]{0.49\linewidth}
		\includegraphics[width=1\textwidth]{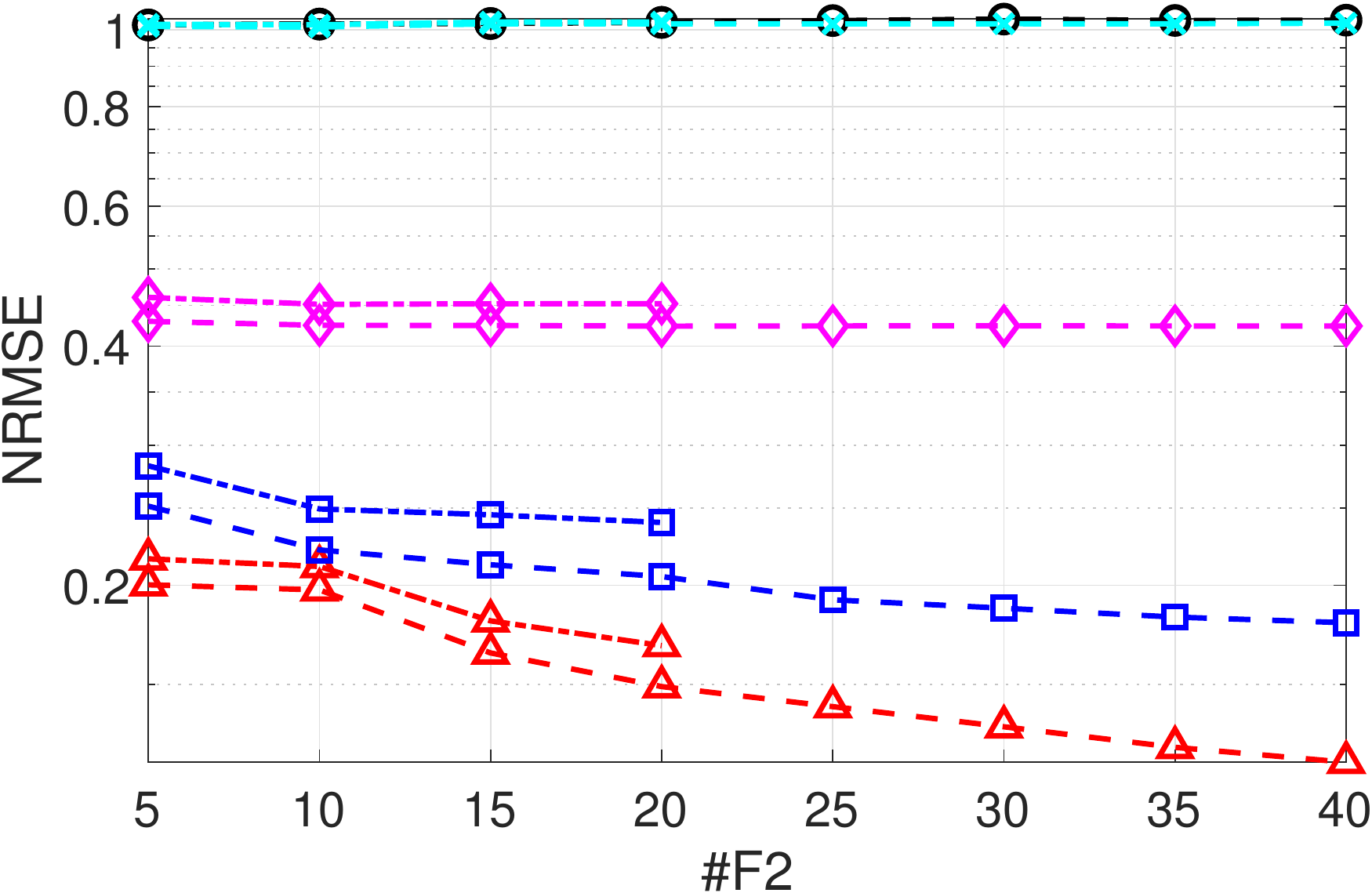}
		\caption{Velocity profile  near the pipe exit.}
	\end{subfigure}
	\begin{subfigure}[b]{0.49\linewidth}
		\includegraphics[width=1\textwidth]{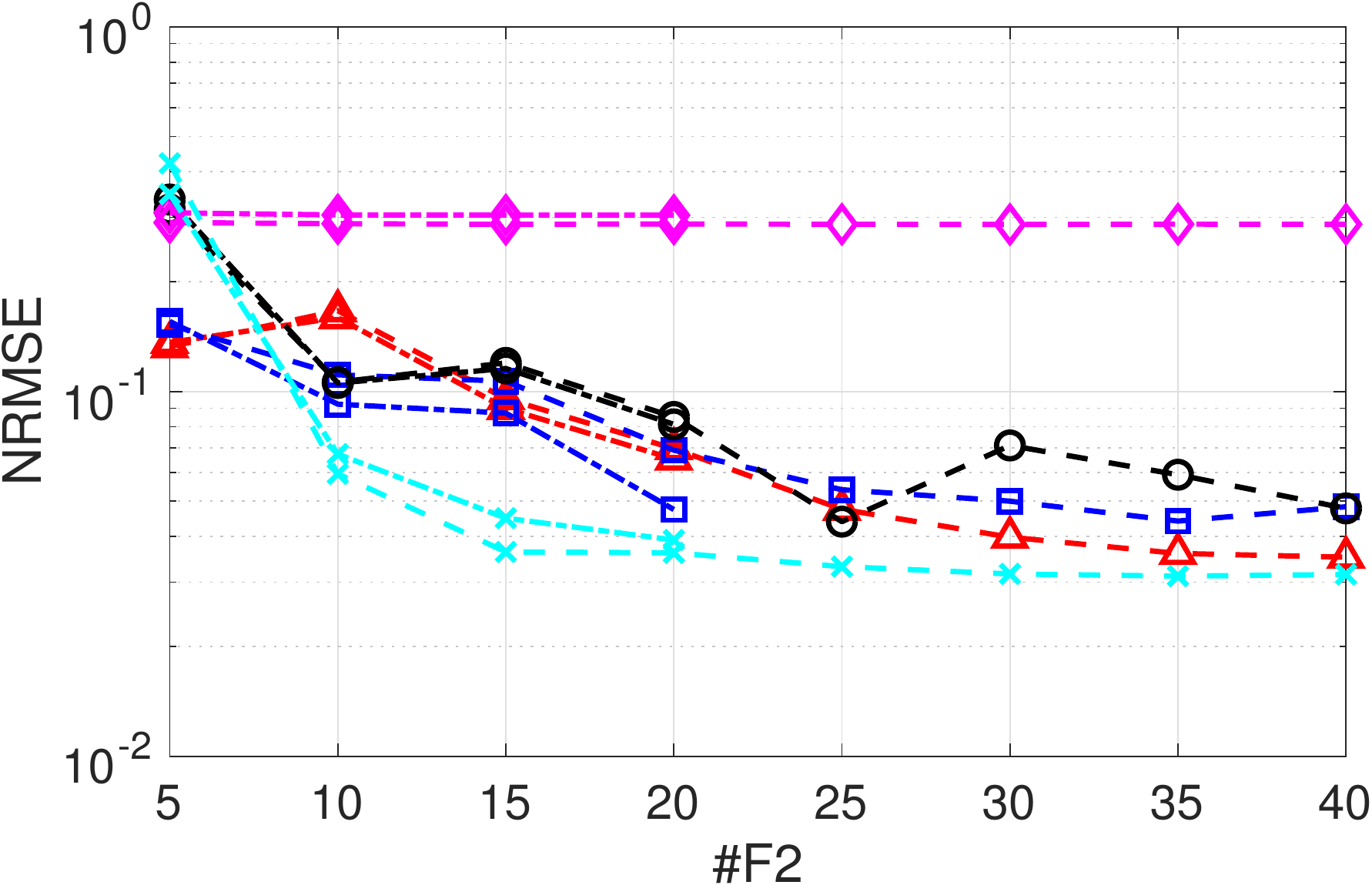}
		\caption{Pressure profile near the pipe junction.}
	\end{subfigure}
	\begin{subfigure}[b]{0.49\linewidth}
		\centering
		\includegraphics[width=1\textwidth]{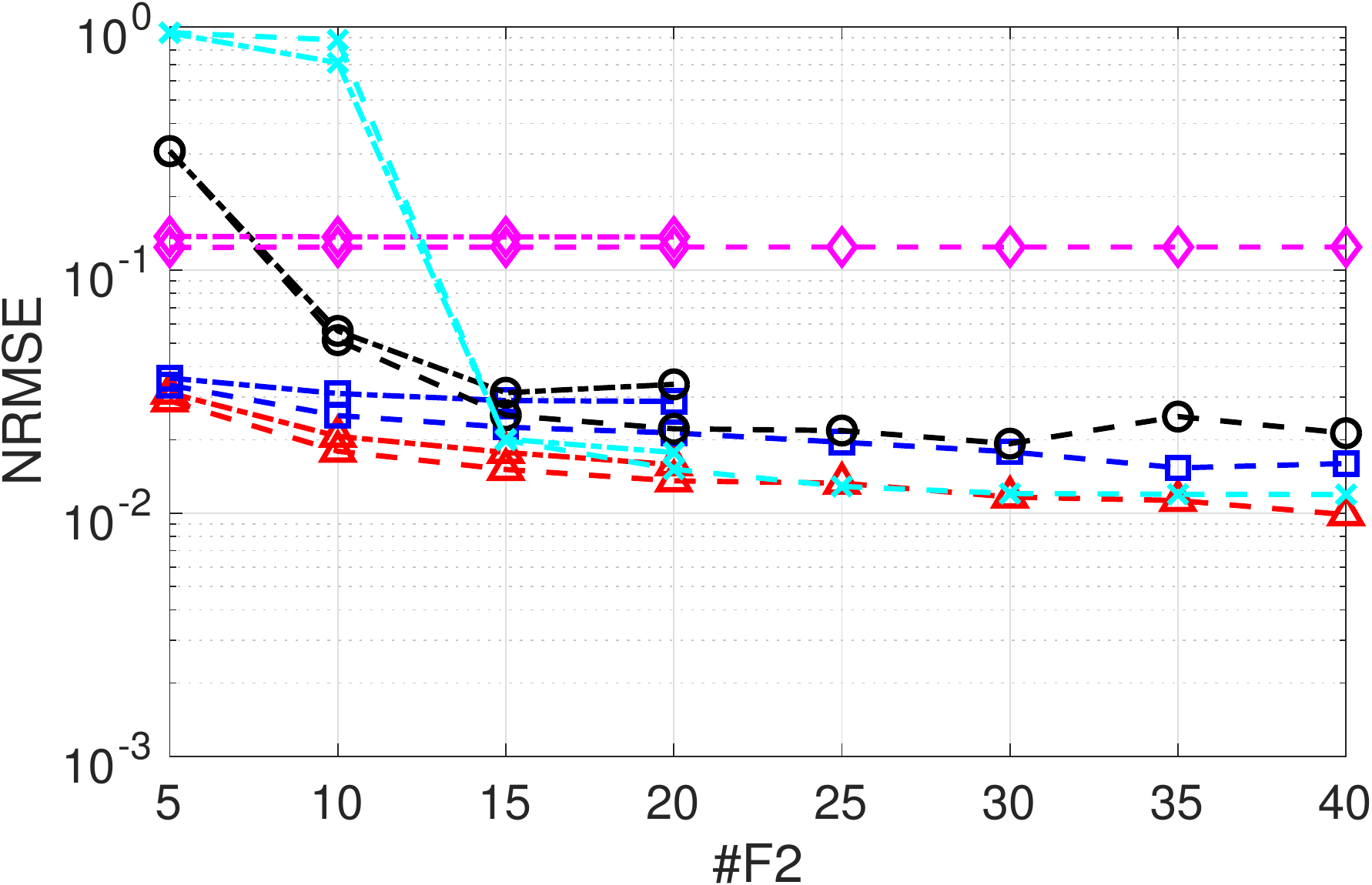}
		\caption{Velocity profile near the pipe junction.}
	\end{subfigure}
	\caption{Normalized root mean square errors (NRMSE) against 18 F2 test values on the two-fidelity turbulent mixing flow simulation for \ours with (red) and without (blue) active learning, NARGP, Greedy NAR and SC.}
	\label{mf1}
\end{figure}

We conducted tests with up to 40 F1 and 40 F2 training points, generated with randomly selected inputs.  For 20 and 40 F1 training points, the number of F2 training samples was gradually  increased to assess the performance of \ours (with and without active learning) and the other methods. We computed the NRMSE against 18 F2 test points (generated in the same way as the training inputs) for each quantity of interest, averaged from a 5-fold cross validation with random shuffling of training and test data. {The results are shown in Fig.~\ref{mf1} and Table \ref{tableE1} in Appendix C}, from which it is seen that \ours generally outperforms NARGP and SC, especially with active learning and with a small number of F2 training data points; this is a significant advantage when data is scarce. {For example, for 40 F1 and 10 F2 training points in the case of the pressure profile near the exit, \ours without active learning has a 9\%, 41\% and 48\% lower NRMSE than NARGP, SC and Greedy NAR, respectively. In the case of the velocity profile near the exit, the equivalent figures are 77\%, 41\% and 75\%, while for the  velocity profile near the pipe junction they are 51\%, 80\% and 97\%. }  NARGP and ResGP without active learning converge to a similar level of error for 40 F1 training points, provided NARGP does not fail, as is the case for the velocity profile near the exit. {In the latter case, the NRMSE for NARGP is around 1 for all F2 training point numbers, for both 20 and 40 F1 training points, whereas \ours with (without) active learning attains values of 0.1014 (0.1323) and 0.1267 (0.2039) for 20 F2 and 40 F2 training points, respectively.} ResGP is more accurate with active learning, with the exception of the pressure profile near the pipe junction when the number of F2 training samples is low. For this quantity of interest, NARGP and both ResGP methods exhibit similar levels of accuracy up to 25 F2 training points for 40 F1 training samples. 

Greedy NAR performs well in three cases, although, like NARGP, it fails for the velocity profile near the pipe exit.   For the pressure profile near the pipe junction, it exhibits the best performance, {especially for a low number of F2 training points (for 40 F1 and 15 F2 training points it is at least 40\% more accurate than the other methods).} Noticeable, however, is that it tends to perform poorly for small numbers of F2 training points on the other quantities of interest. Also evident from these figures is that \ours is the most robust of the methods. It is not surprising that both NARGP and Greedy NAR perform worse when F2 data is scarce, since the number of parameters in these models (Table 1) is high. Only when sufficient data is available can these parameters be learned with accuracy.

\subsection{Test problem 3: molecular dynamics simulation model} 
In this example we consider a molecular dynamics (MD) model based on the Lennard-Jones (LJ) potential $u(\cdot)$ ~\citep{lee2016computational} for the interatomic interactions
\begin{equation}
u\left(r_{ij}\right)=4\varepsilon \left [ \left(\frac{\sigma}{r_{ij}}\right)^{12}-\left(\frac{\sigma}{r_{ij}}\right)^6 \right ],
\label{eq:LJ1}
\end{equation}
where $r_{ij}$ is the pairwise distance between particles $i$ and $j$, $\epsilon$ is the potential well depth and $\sigma$ is the length scale for the interatomic interaction. In order to prevent  numerical instabilities, which can arise when the time step is too large,  the magnitude of the repulsive interactions for closely approaching atoms was capped when the ratio  $\sigma/r_{ij}$ exceeded 1.2. 
   
The system parameters were taken to be the  temperature and density, which define the phase space.  We used values of $\sigma=3$ $\angstrom$ and $\varepsilon=1$ kcal mol$^{-1}$ to fully define the LJ potential. The domain was a  cube with width $L=27.05$ $\angstrom$ and  periodic boundary conditions were assumed to hold on all sides. The molecular mass of each particle was set to $m=12.01$ g mol$^{-1}$. The temperature $T$ and density $\rho=Nm/V$ were used as the inputs. Here, $V=L^{3}$ is the  box volume and $N$ is Avogadro's number. The dimensionless density $\rho^{\ast}=N\sigma^3/V$ was therefore in the range $\rho^{\ast} \in [0.05,0.95]$, corresponding to a molecule number in the range 36 to 696. The simulations were performed using the Large-scale Atomic/Molecular Massively Parallel Simulator (LAMMPS) code.  Integration of the equations of  motion is based on the Verlet algorithm, and a microcanonical (NVE) ensemble was used. The quantities of interest were the radial distribution function (RDF), mean squared displacement (MSD), and self-diffusion coefficient (SDC). All multi-fidelity methods were applied to the quantities of interest separately. The RDF was recorded at $d=512$ points, the MSD at $d=2000$ points and the SDC is a scalar ($d=1$). Low-fidelity (F1) and high-fidelity (F2) simulations were defined by time steps of 10 and 1 fs, respectively. 
We tested all methods with up 40 F1 and 40 F2 training points, generated with randomly selected inputs in the ranges $T\in[500, 1000]\,\text{K}$ and $\rho\in[36.27,701.29]\,\text{kg m$^{-3}$}$. For 20 and 40 F1 training points, the number of F2 training points was increased gradually to assess the performance of each method by calculating the NRMSE against 34  F2 test points, generated in the same way as the training points. Again, the experiments were repeated 5 times with random shuffling of  training and test data and the average NRMSE was used.

\begin{figure}[!h]
	\centering
	\begin{subfigure}[b]{0.85\linewidth}
		\centering
		\includegraphics[width=1\textwidth]{legend_5col}
	\end{subfigure}\\
	\begin{subfigure}[b]{0.49\linewidth}
		\includegraphics[width=1\textwidth]{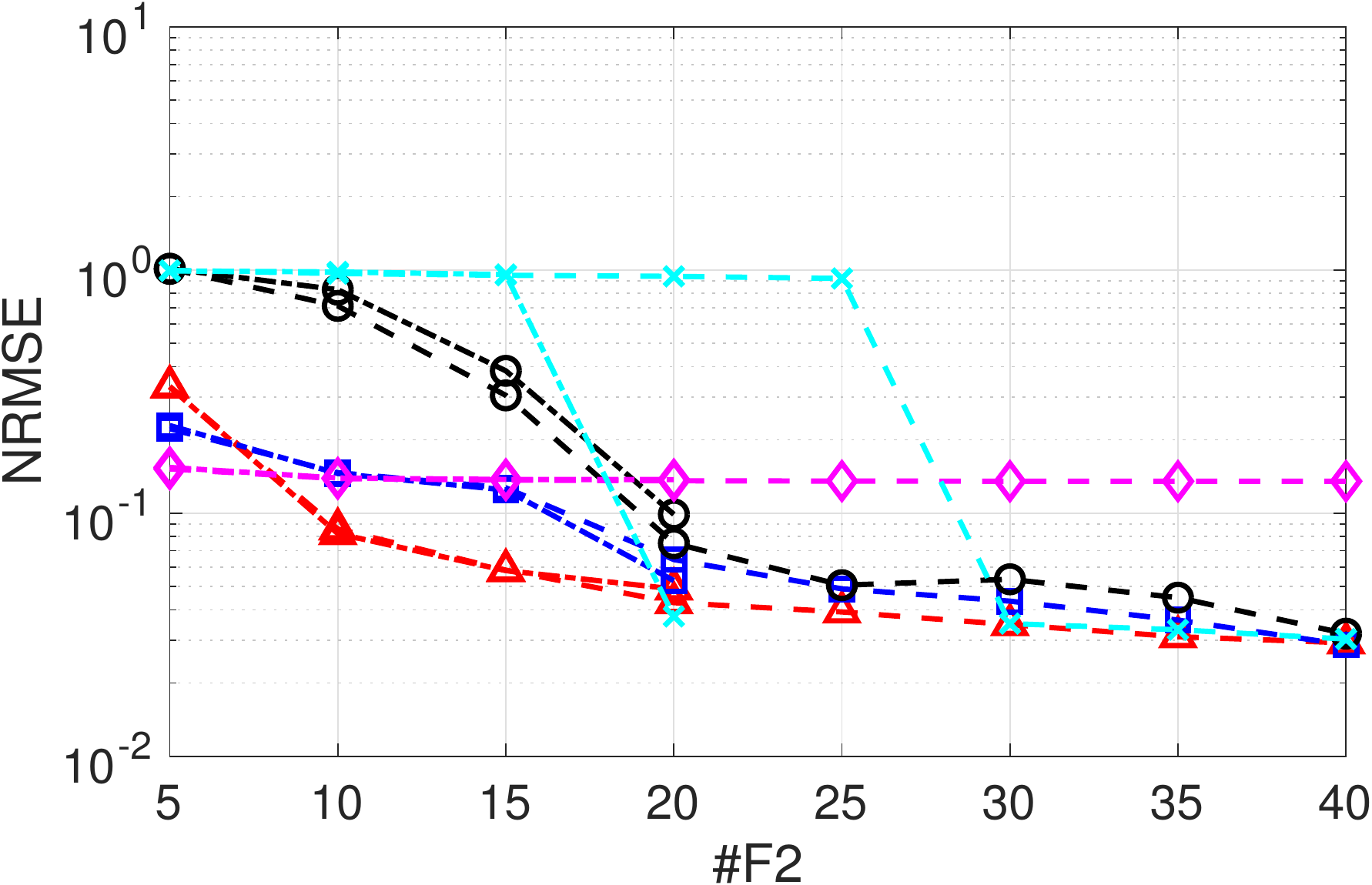}
		\caption{Radial distribution function}
	\end{subfigure}
	\begin{subfigure}[b]{0.49\linewidth}
		\includegraphics[width=1\textwidth]{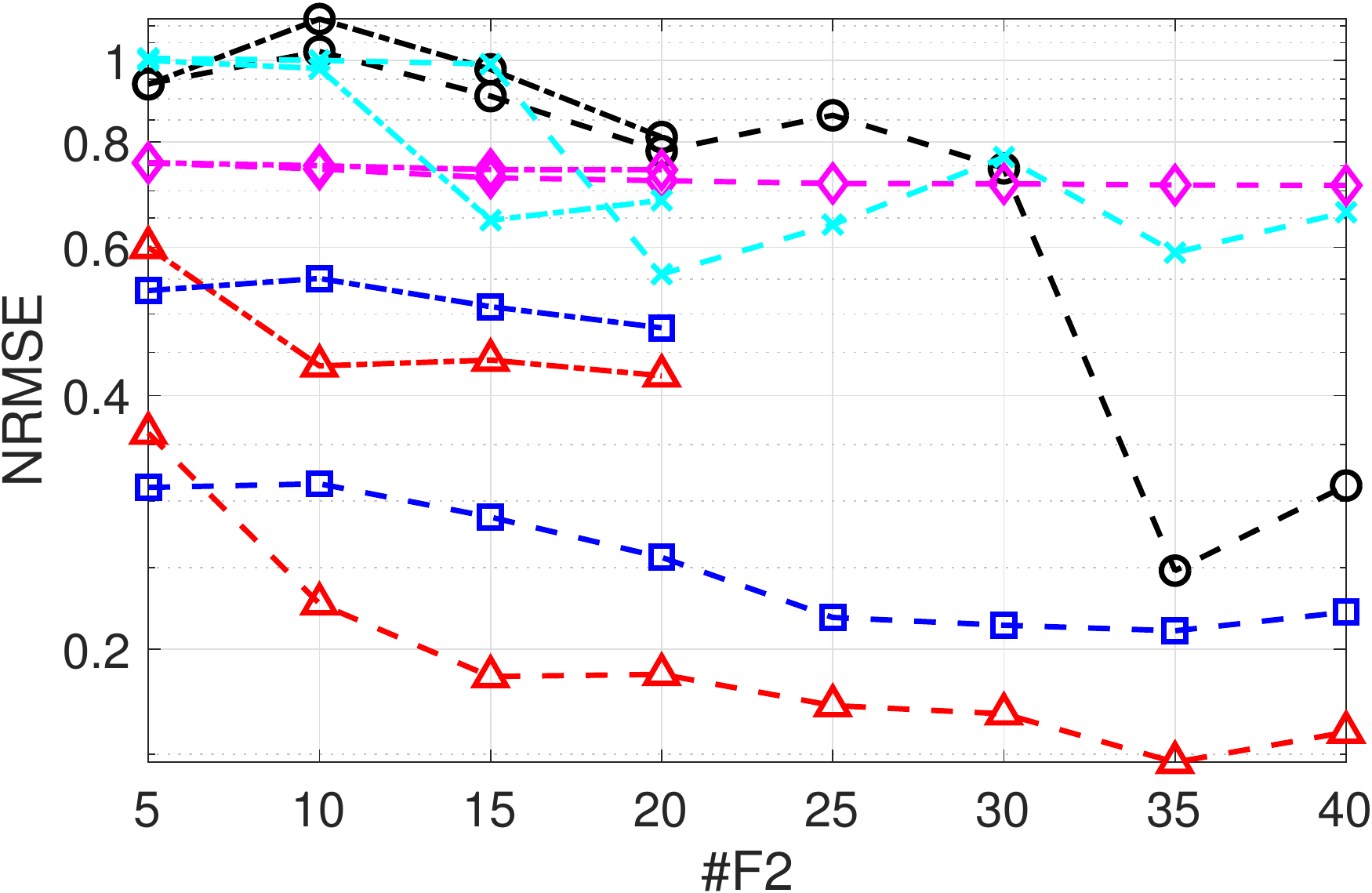}
		\caption{Mean Squared Distance}
	\end{subfigure}
	\begin{subfigure}[b]{0.49\linewidth}
		\centering
		\includegraphics[width=1\textwidth]{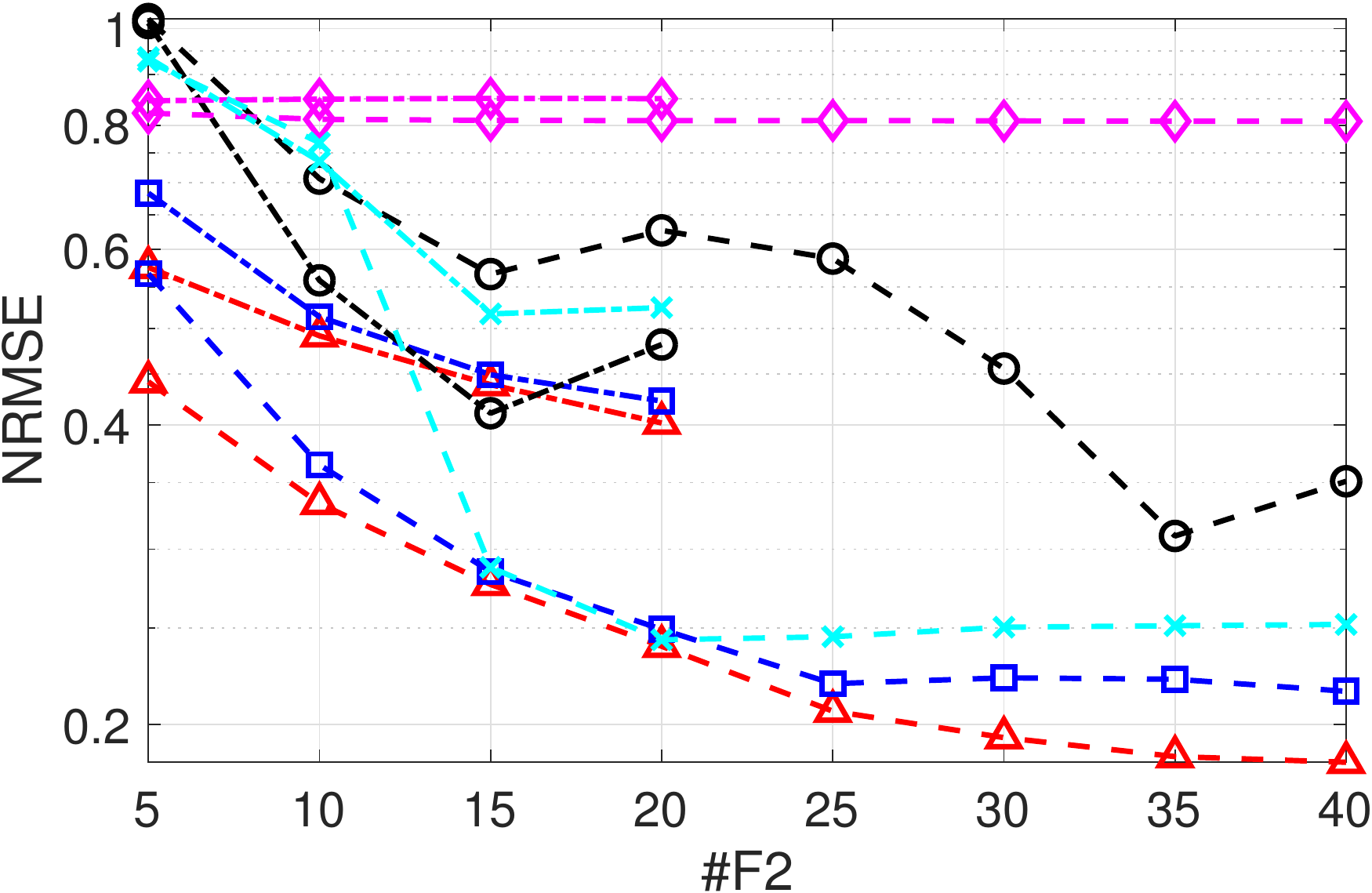}
		\caption{Self diffusion coefficient}
	\end{subfigure}
	\caption{Normalized root mean square errors (NRMSE) against 34 F2 test values on the two-fidelity MD simulation for  \ours with (red) and without (blue) active learning, NARGP, Greedy NAR and SC.}
	\label{md1}
\end{figure}
{{For each of the quantities of interest, the results for \ours (with and without active learning), SC,  NARGP and Greedy NAR are shown in  Fig. ~\ref{md1} and Table \ref{tableE2} in Appendix C}. In this example the superiority of \ours (both with and without active learning) is more obvious, as is the greater accuracy for low numbers of F2 training points. {For example, for 40 F1 and 10 F2 training points in the case of the RDF, \ours with active learning has an 88\%, 38\% and 91\% lower NRMSE than NARGP, SC and Greedy NAR, respectively. For the MSD, the equivalent figures are 78\%, 70\% and 77\%, while for the  SDC they are 53\%, 59\% and 57\%.} Active learning is seen again to enhance the performance of \ours. For the MSD, the failure of NARGP and Greedy NAR is due to the linear scaling of the number of parameters in both methods with $d$ (in this case $d=2000$). For a high number of F2 training data, the accuracy of both NARGP and  Greedy NAR (especially the latter) improves  for the RDF and SDC. Again, this is due to the greater demand for   training samples in order to learn the high number of parameters accurately.}

\subsection{Test Problem 4: solid oxide fuel cell model}\label{ssec:sofc}

\begin{figure}[h!]
	\centering
	\includegraphics[width=0.65\textwidth]{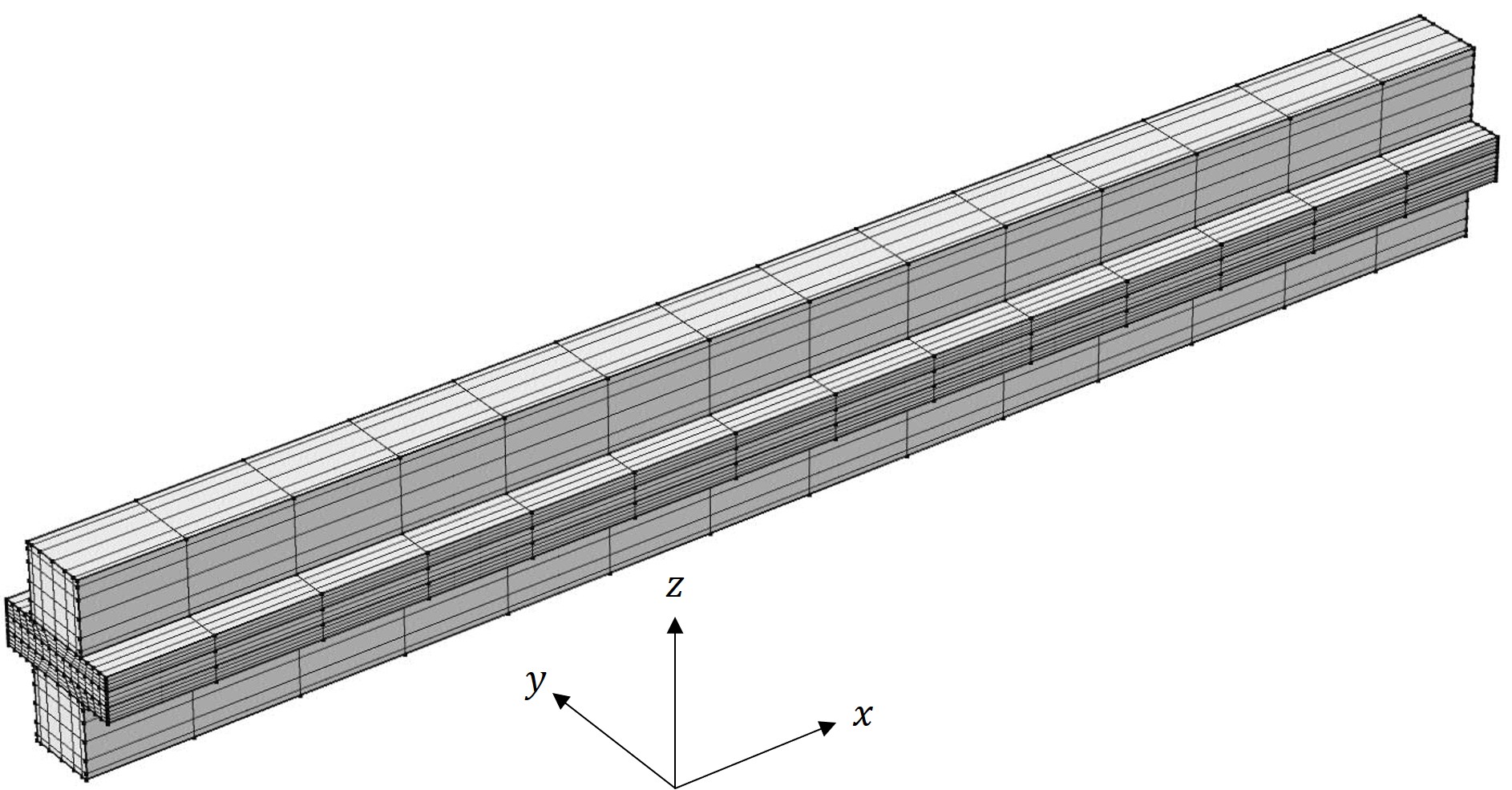}
	\caption{Computational domain for the SOFC example: gas channels, electrodes and electrolyte, with the cathode at the top. From top to bottom, the layers are a channel, electrode, electrolyte, electrode and channel. The channel dimensions are ($x\times y\times z$) 1 cm $\times$ 0.5 mm $\times$  0.5 mm, the electrode dimensions are 1 cm $\times$ 1 mm $\times$  0.1 mm, and the electrolyte dimensions are 1 cm $\times$ 1 mm $\times$  0.1 mm. The anode inlet is located at $x=0$ and the cathode inlet at $x=1$ cm.}
	\label{fig:sofc} 
\end{figure}
In the last example we consider a steady-state 3-d solid oxide fuel cell model. The geometry is depicted in Fig.  \ref{fig:sofc}. The model includes: electronic and ionic charge balances (Ohm’s law); the flow distribution in the gas channels (Navier-Stokes equations); flow in the porous electrodes (Brinkman equation); and gas-phase mass balances in both gas channels and the porous electrodes  (Maxwell-Stefan diffusion and convection). Butler-Volmer charge transfer kinetics are assumed for the reactions in the anode ($\mbox{H}_2+\mbox{O}^{2-}\rightarrow \mbox{H}_2\mbox{O}+2\mbox{e}^{-}$)
 and cathode ($\mbox{O}_2+4\mbox{e}^{-}\rightarrow 2\mbox{O}^{2-}$). The cell operates in potentiostatic mode (constant cell voltage). The model was solved in COMSOL Multiphysics\footnote{\url{https://www.comsol.com/model/current-density-distribution-in-a-solid-oxide-fuel-cell-514}} (Application ID: 514), which is based in the finite-element method. 
 
\begin{figure}[!hbp]
	\centering
	\begin{subfigure}[b]{0.85\linewidth}
		\centering
		\includegraphics[width=1\textwidth]{legend_5col}
	\end{subfigure}\\
	\begin{subfigure}[b]{0.49\linewidth}
		\includegraphics[width=1\textwidth]{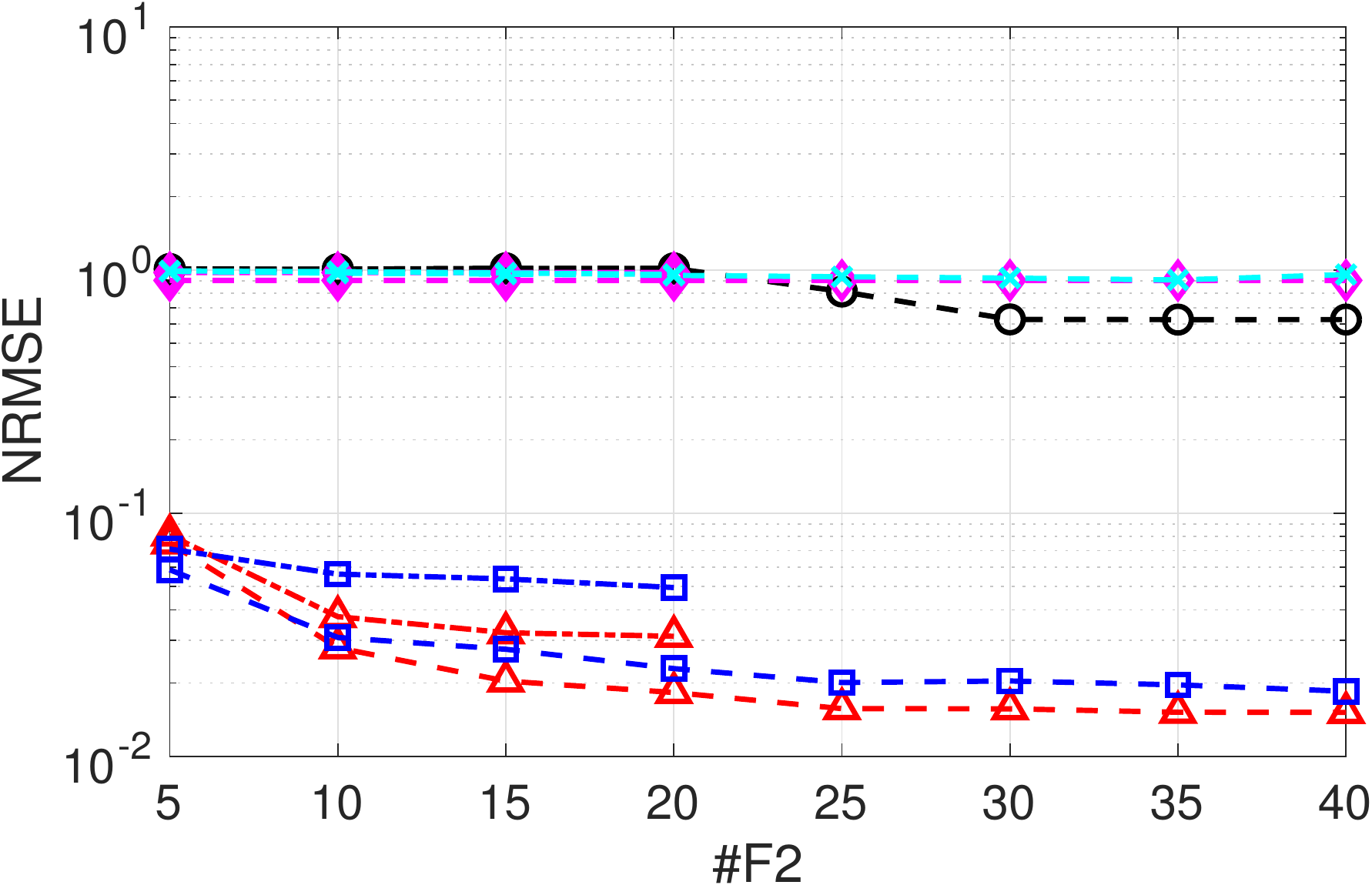}
		\caption{Electrolyte current density.}
	\end{subfigure}
	\begin{subfigure}[b]{0.49\linewidth}
		\includegraphics[width=1\textwidth]{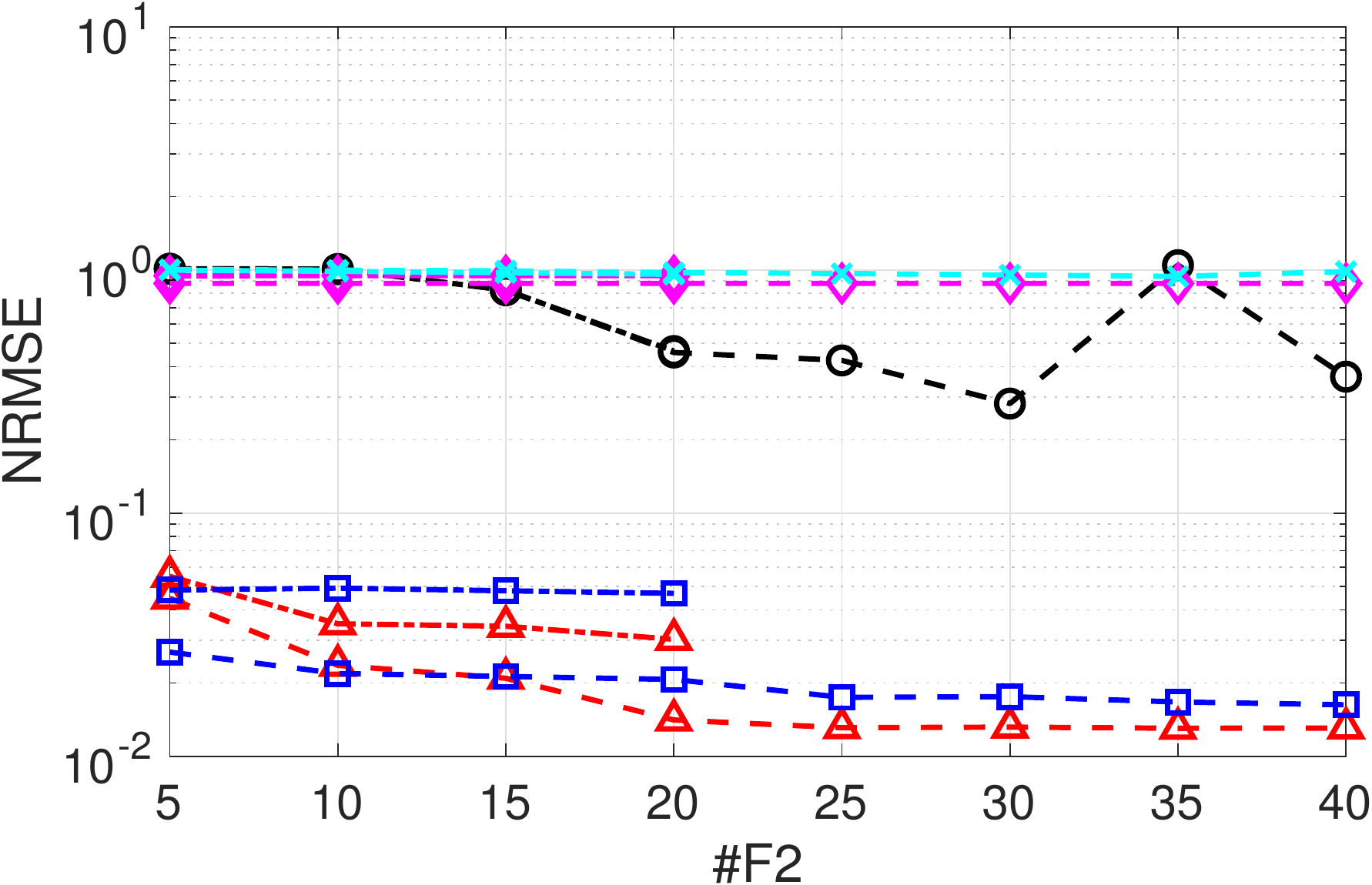}
		\caption{Ionic potential.}
	\end{subfigure}
	\caption{Normalized root mean square errors (NRMSE) against 40 F2 test values on  the two-fidelity SOFC simulation for \ours with (red) and without (blue) active learning, NARGP, Greedy NAR and SC.}
	\label{sofc1}
\end{figure}
 \begin{figure}[!h]
	\centering
	\includegraphics[width=0.32\textwidth]{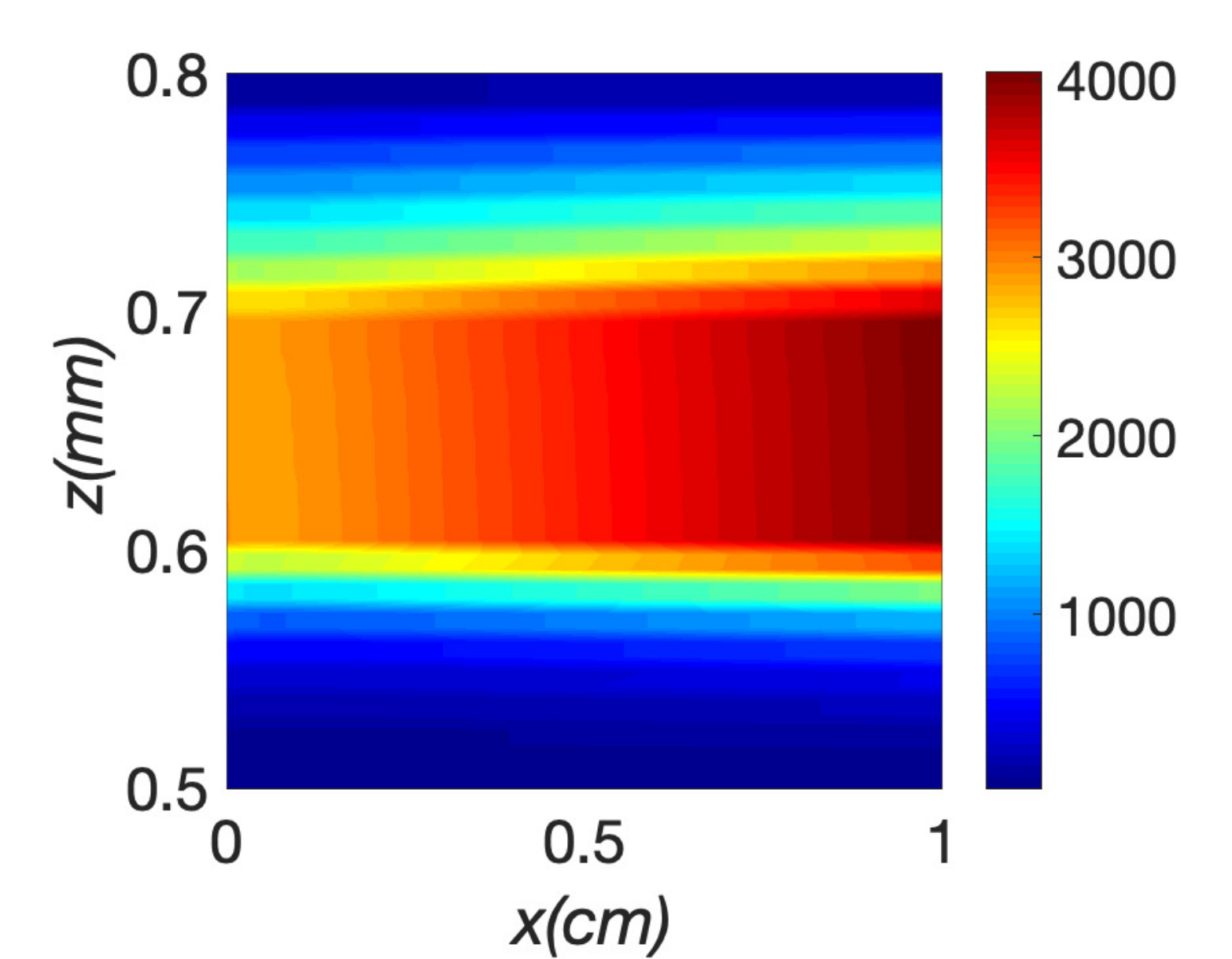}
	\includegraphics[width=0.32\textwidth]{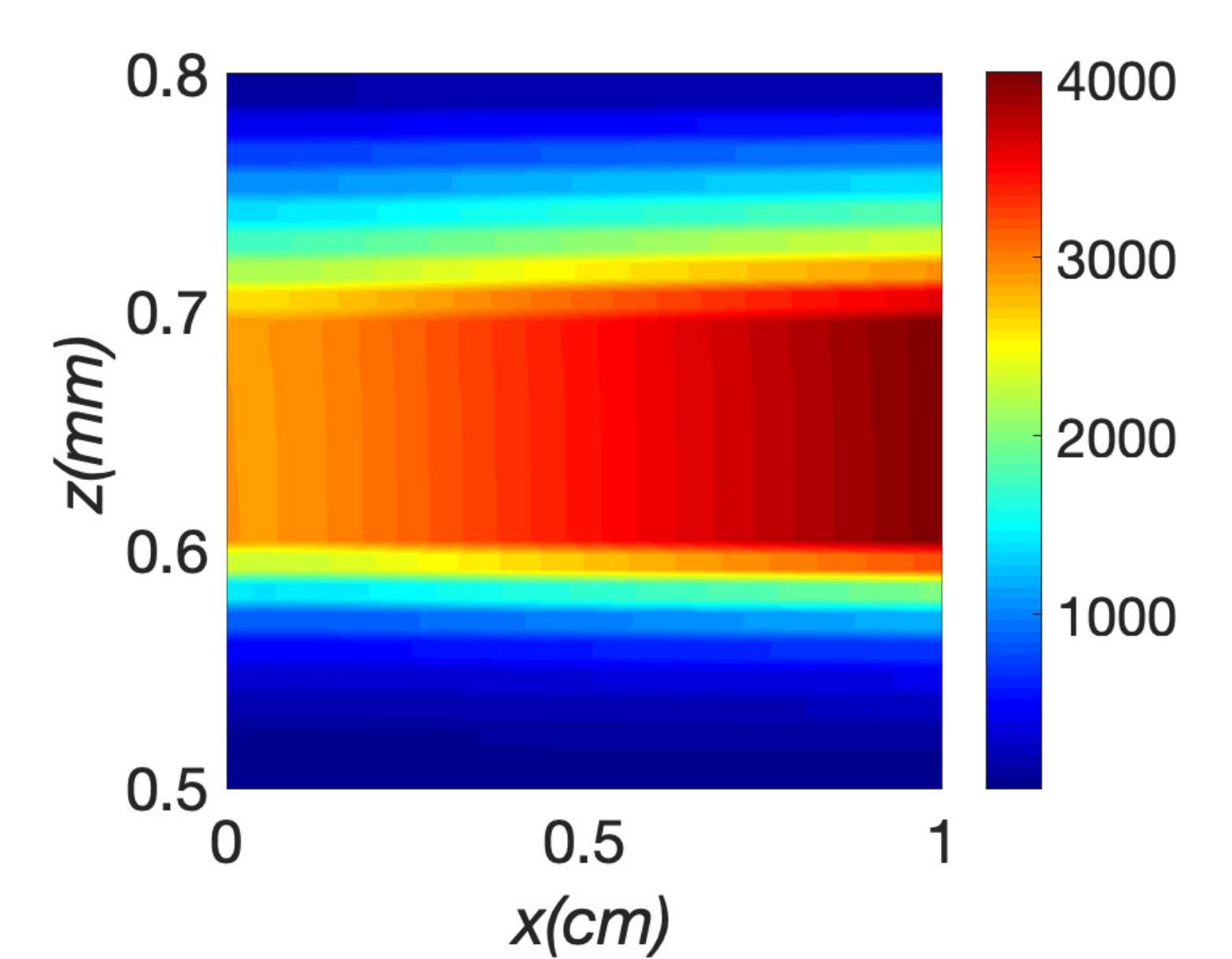}
	\includegraphics[width=0.32\textwidth]{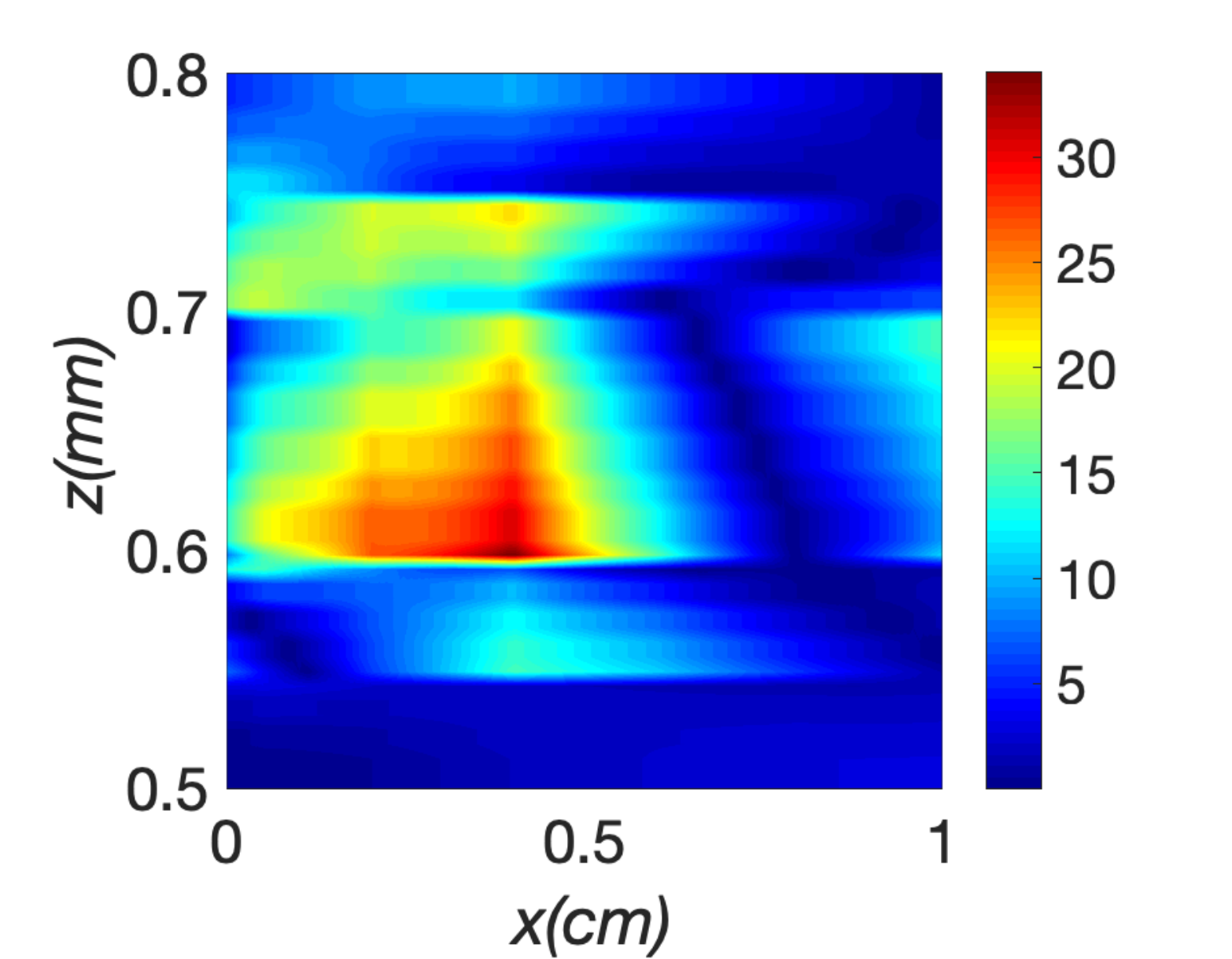}\\
		\includegraphics[width=0.32\textwidth]{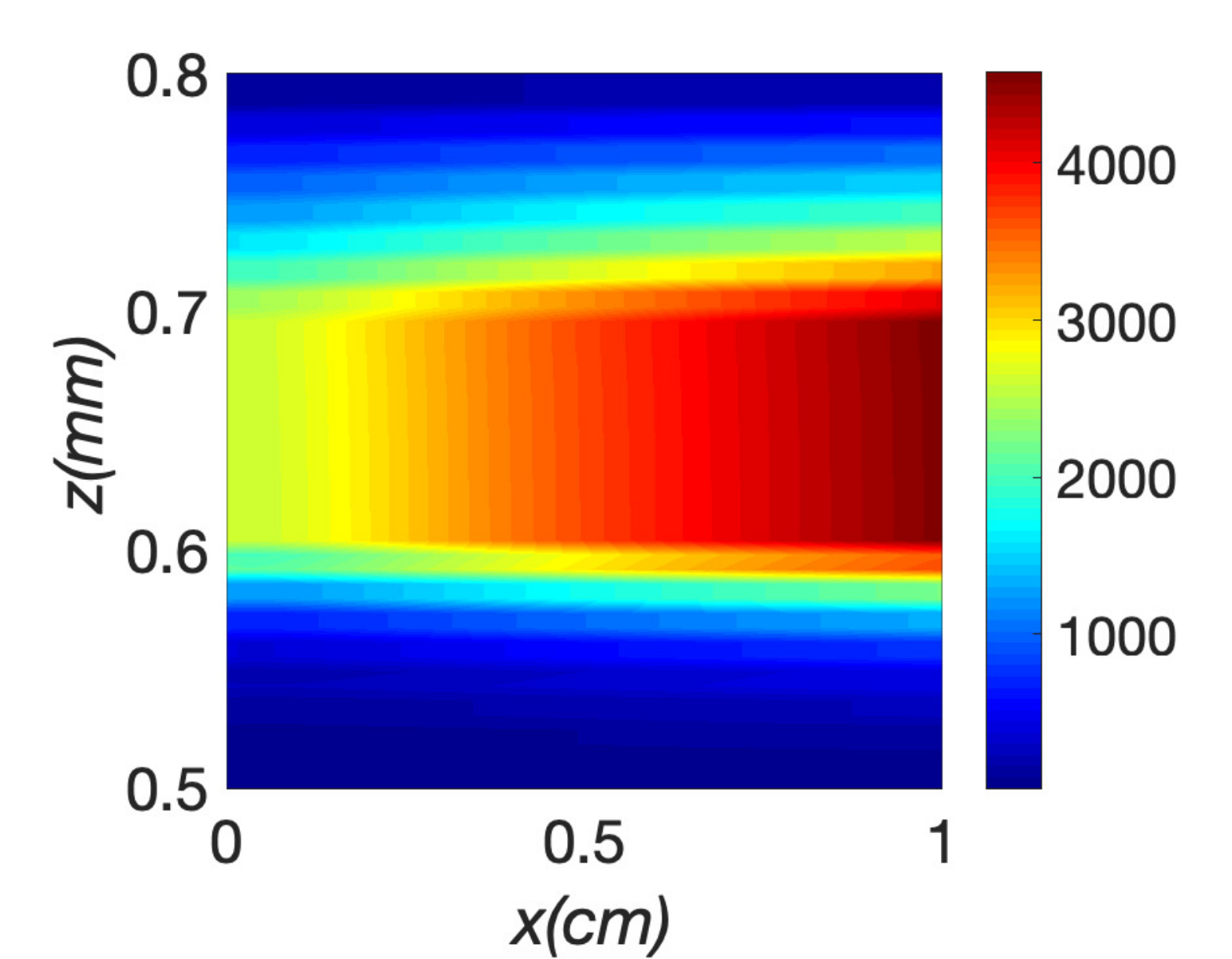}
	\includegraphics[width=0.32\textwidth]{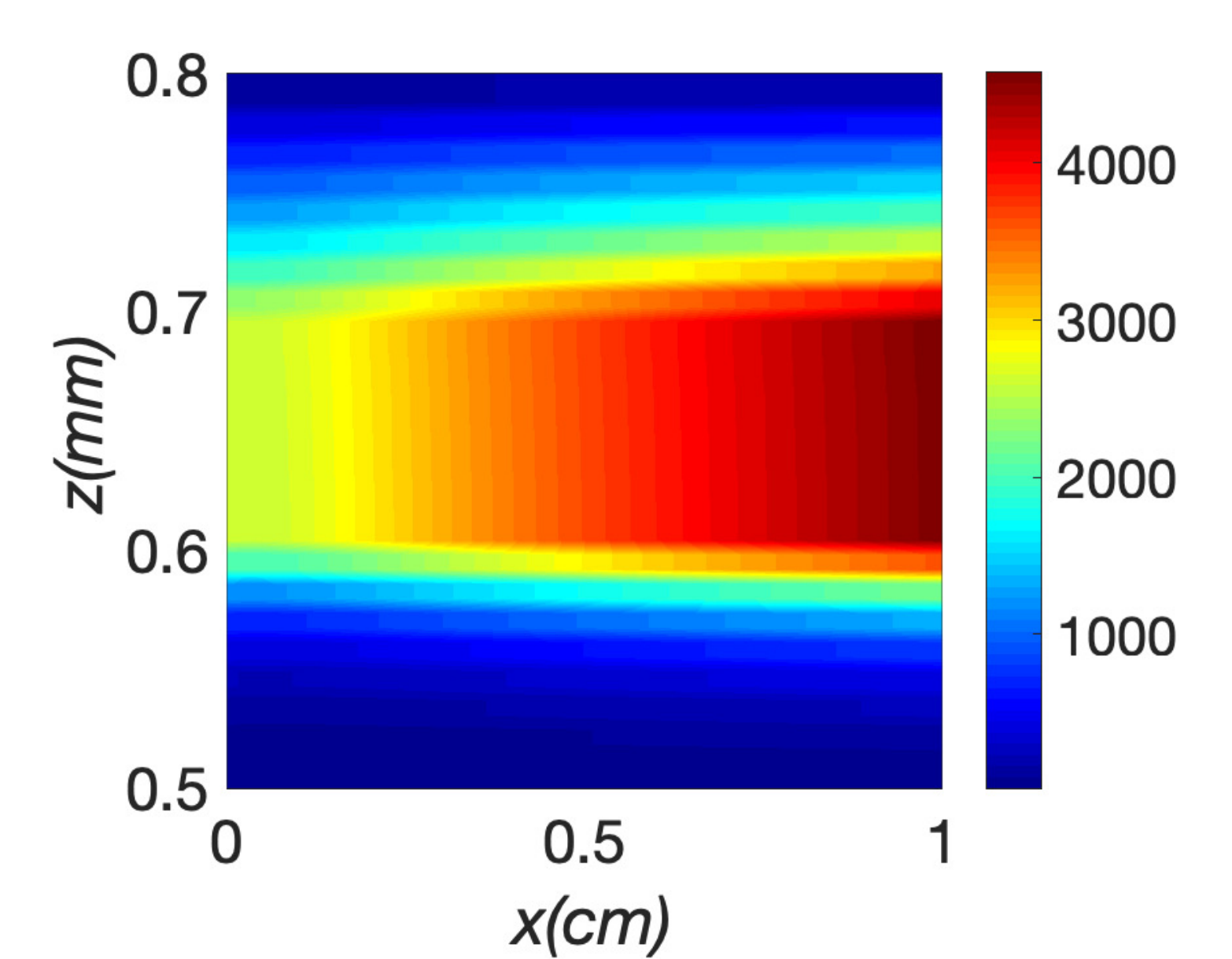}
	\includegraphics[width=0.32\textwidth]{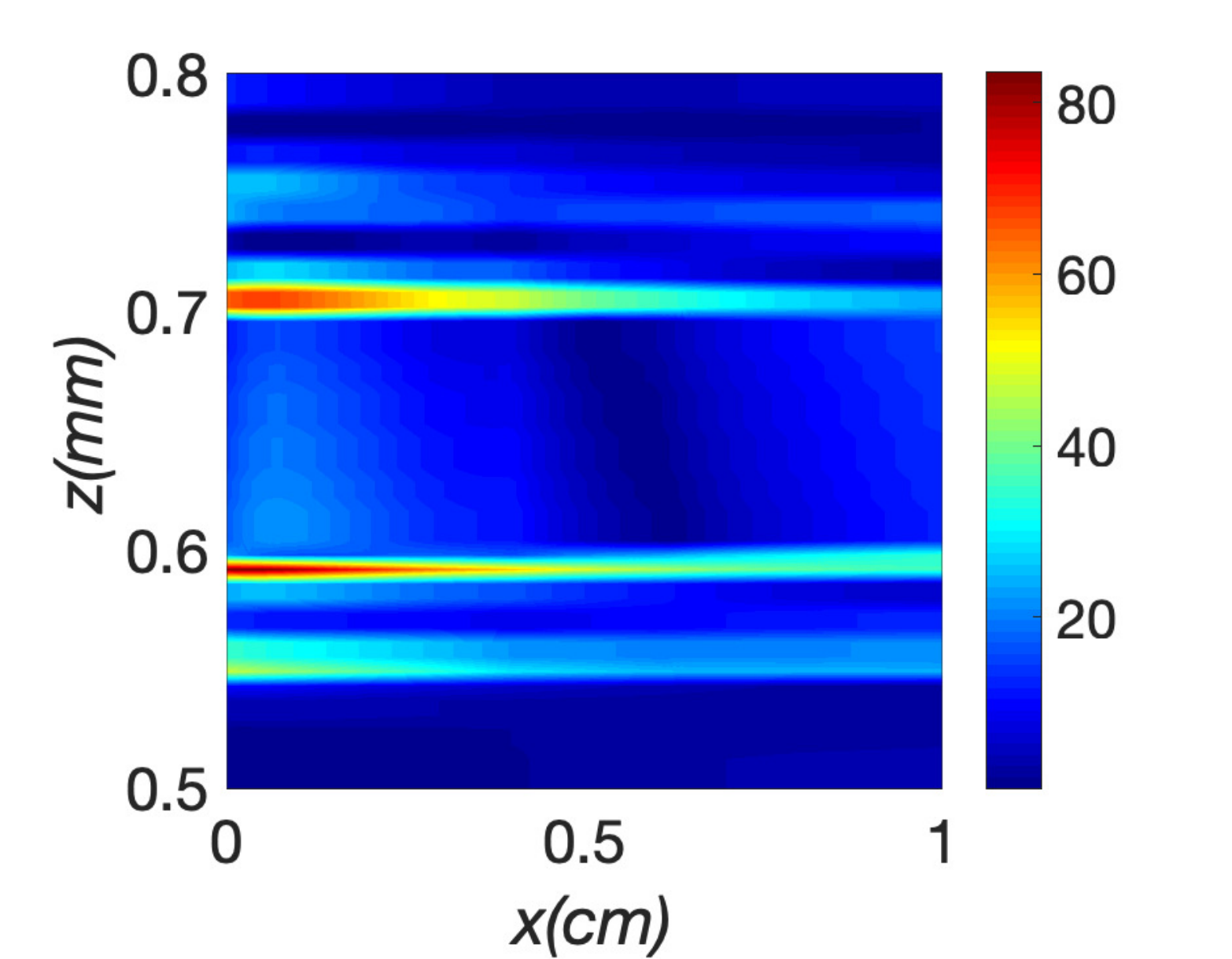}\\
		\includegraphics[width=0.32\textwidth]{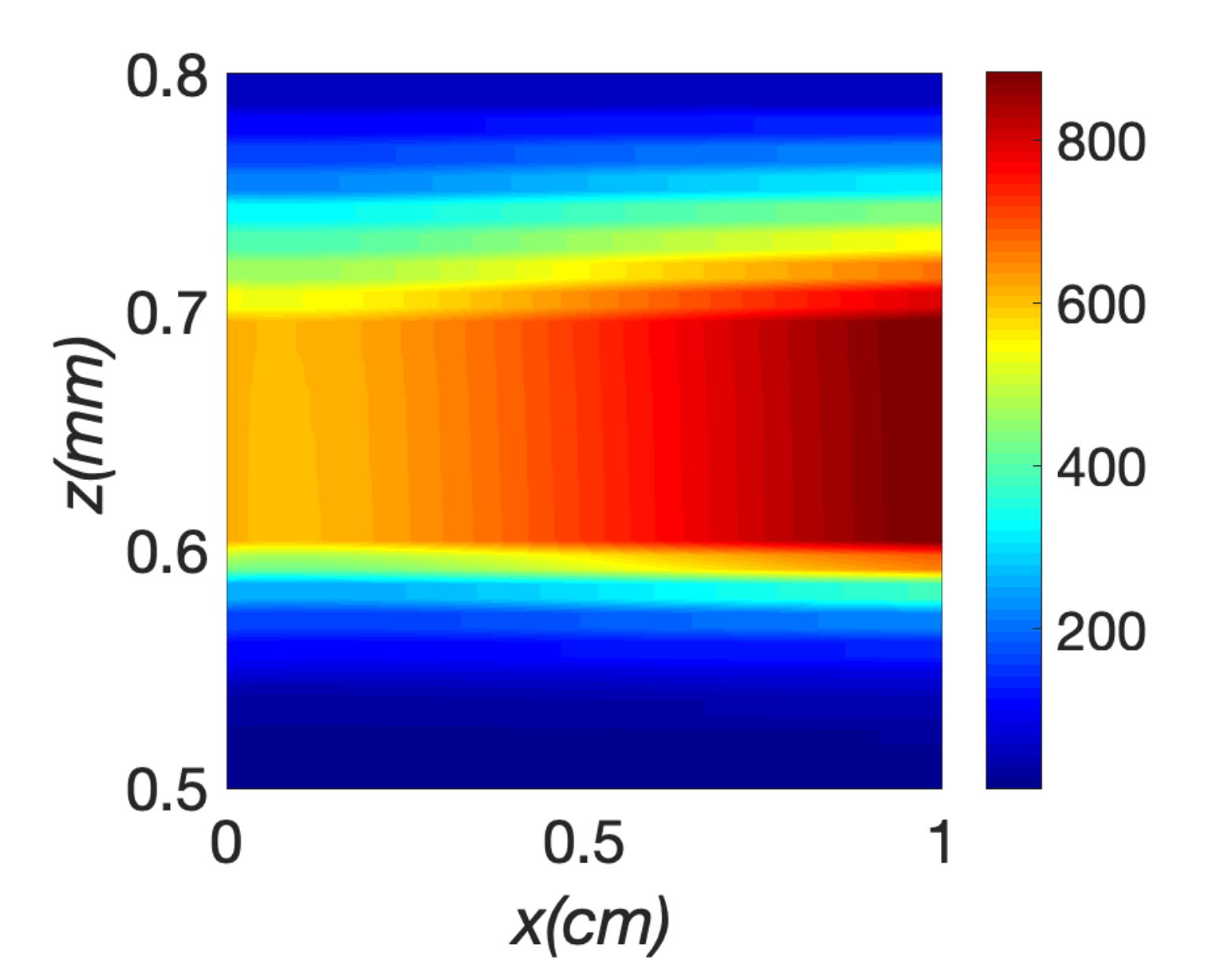}
	\includegraphics[width=0.32\textwidth]{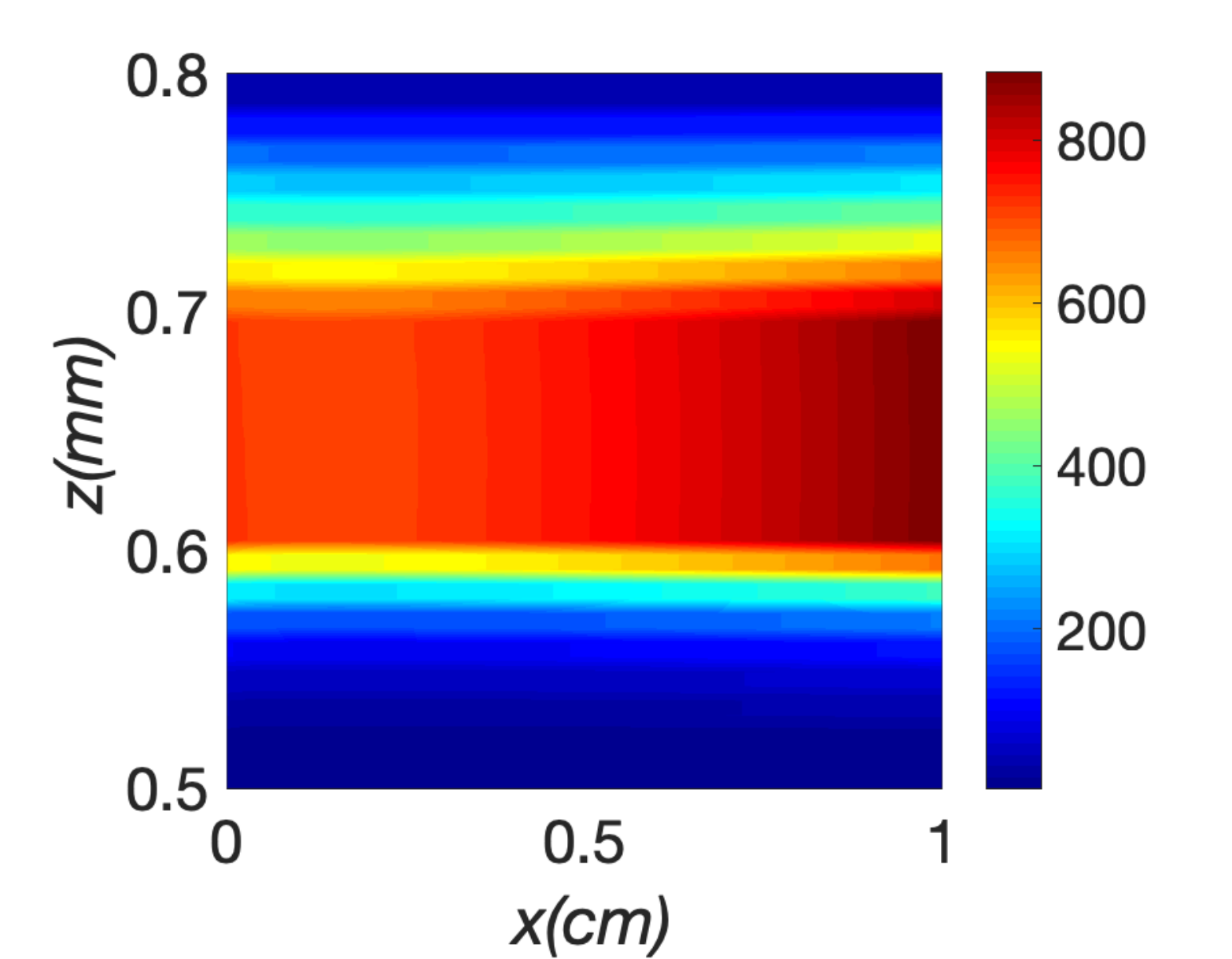}
	\includegraphics[width=0.32\textwidth]{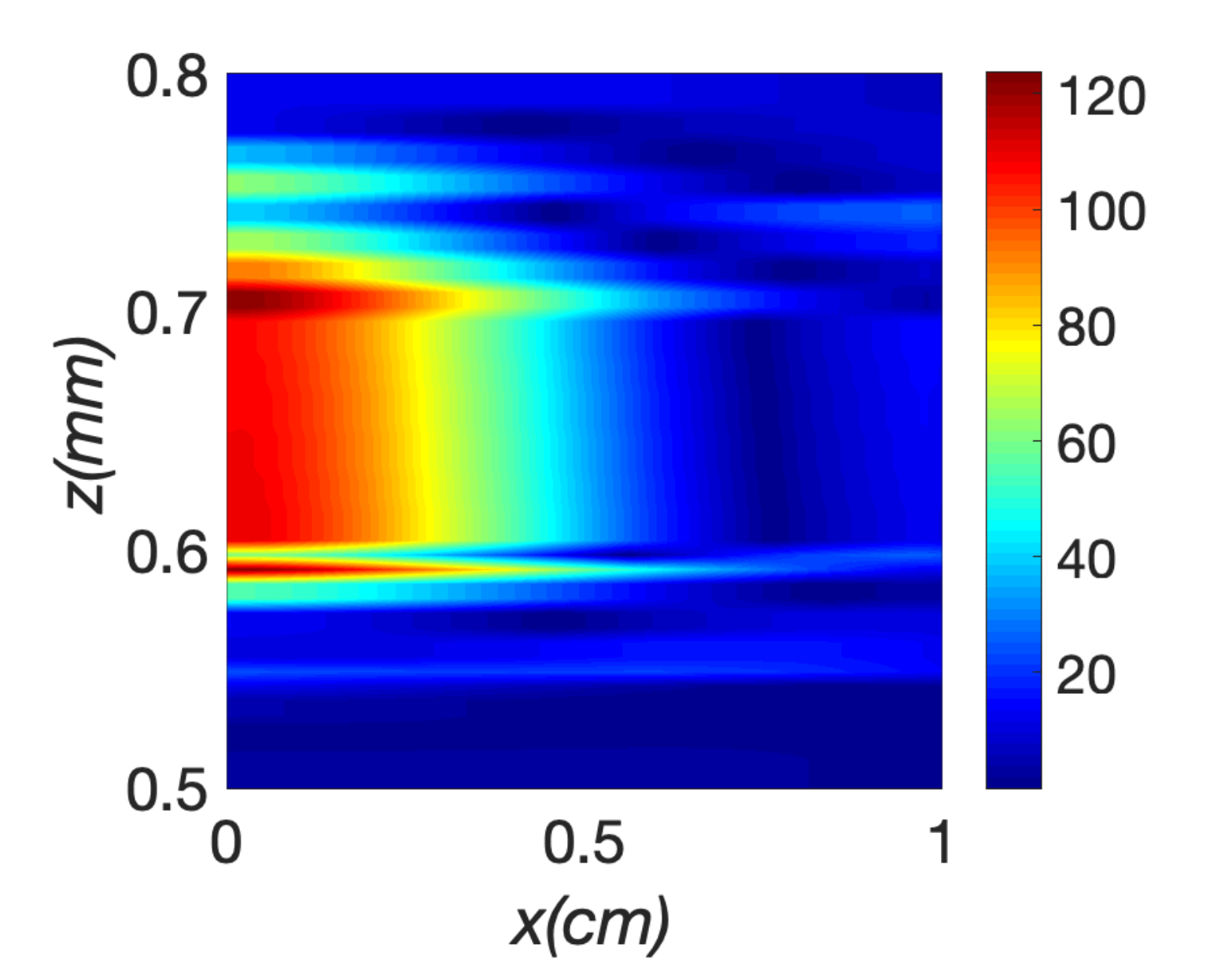}
	\caption{Predictions of the electrolyte current density (A m$^{-2}$) in the $x-z$ plane located at the centre of the channels in Fig.  \ref{fig:sofc} for 40 F1 and 20 F2 training points. These predictions correspond from the top to bottom row to the lowest error, the median error and the highest error  for the 5-fold cross validation. The columns from left to right are the prediction using \ours, the ground truth (test) and the pointwise absolute differences.}
	\label{fig:sofc2} 
\end{figure}

\begin{figure}[h!]
	\centering
	\includegraphics[width=0.32\textwidth]{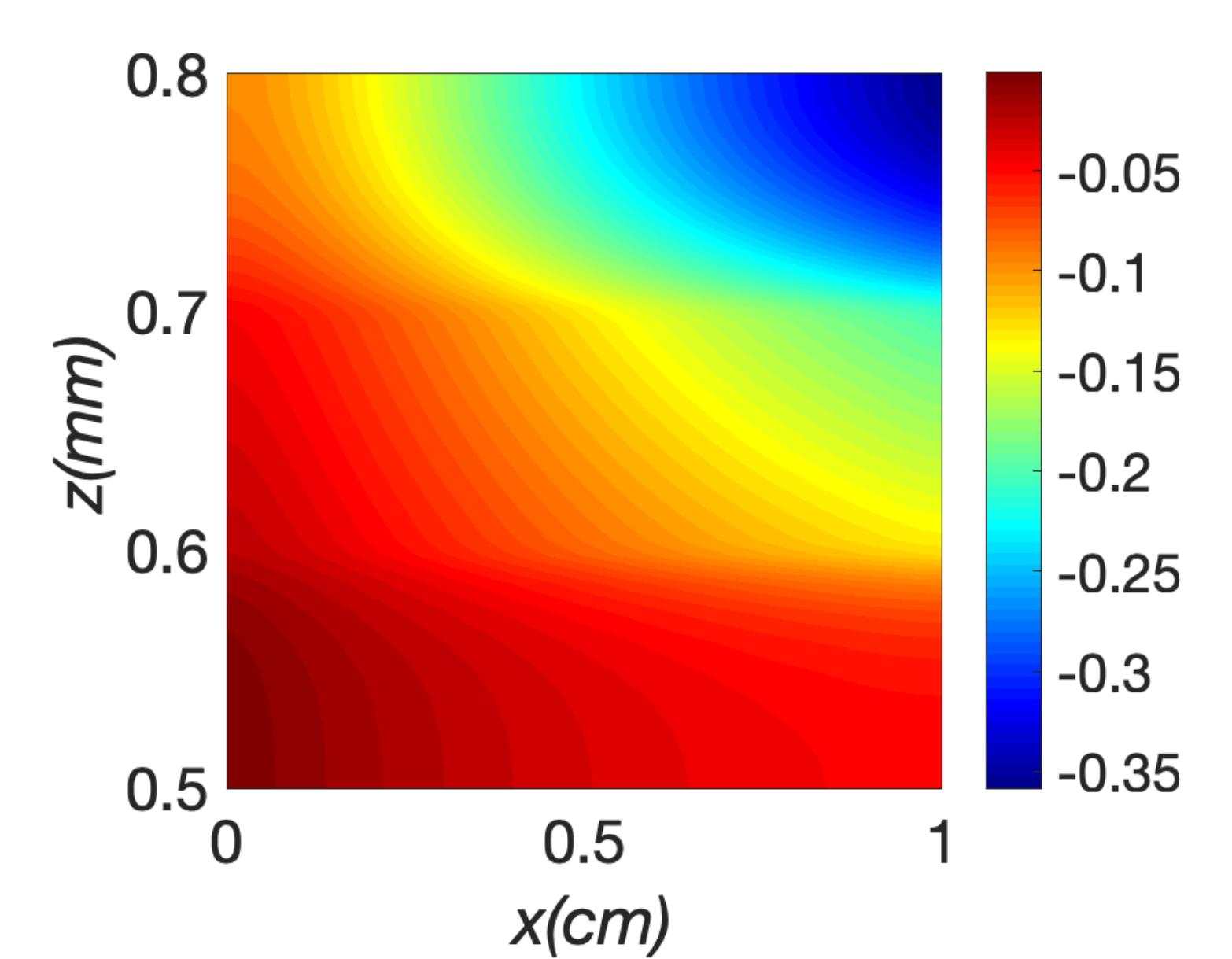}
	\includegraphics[width=0.32\textwidth]{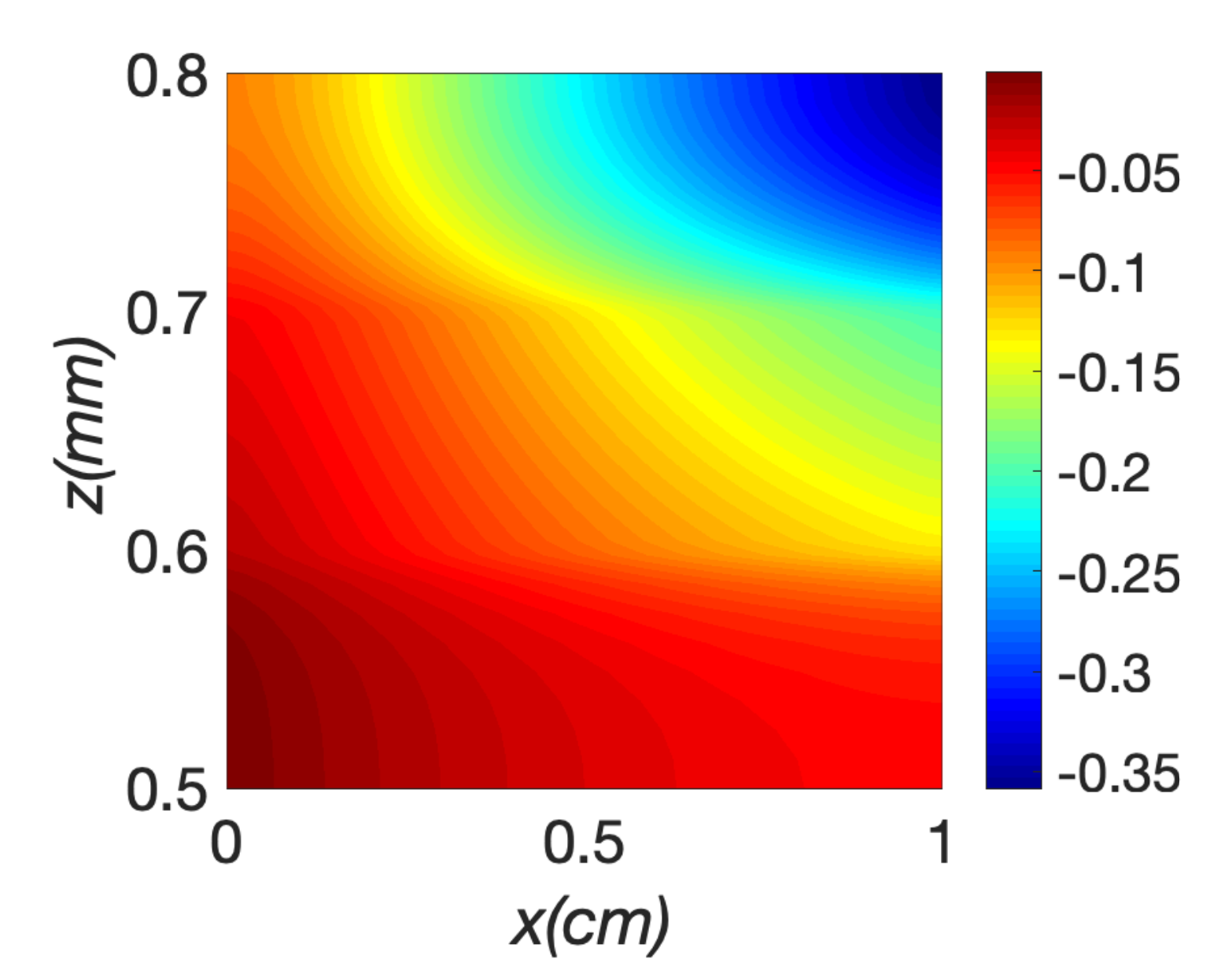}
	\includegraphics[width=0.32\textwidth]{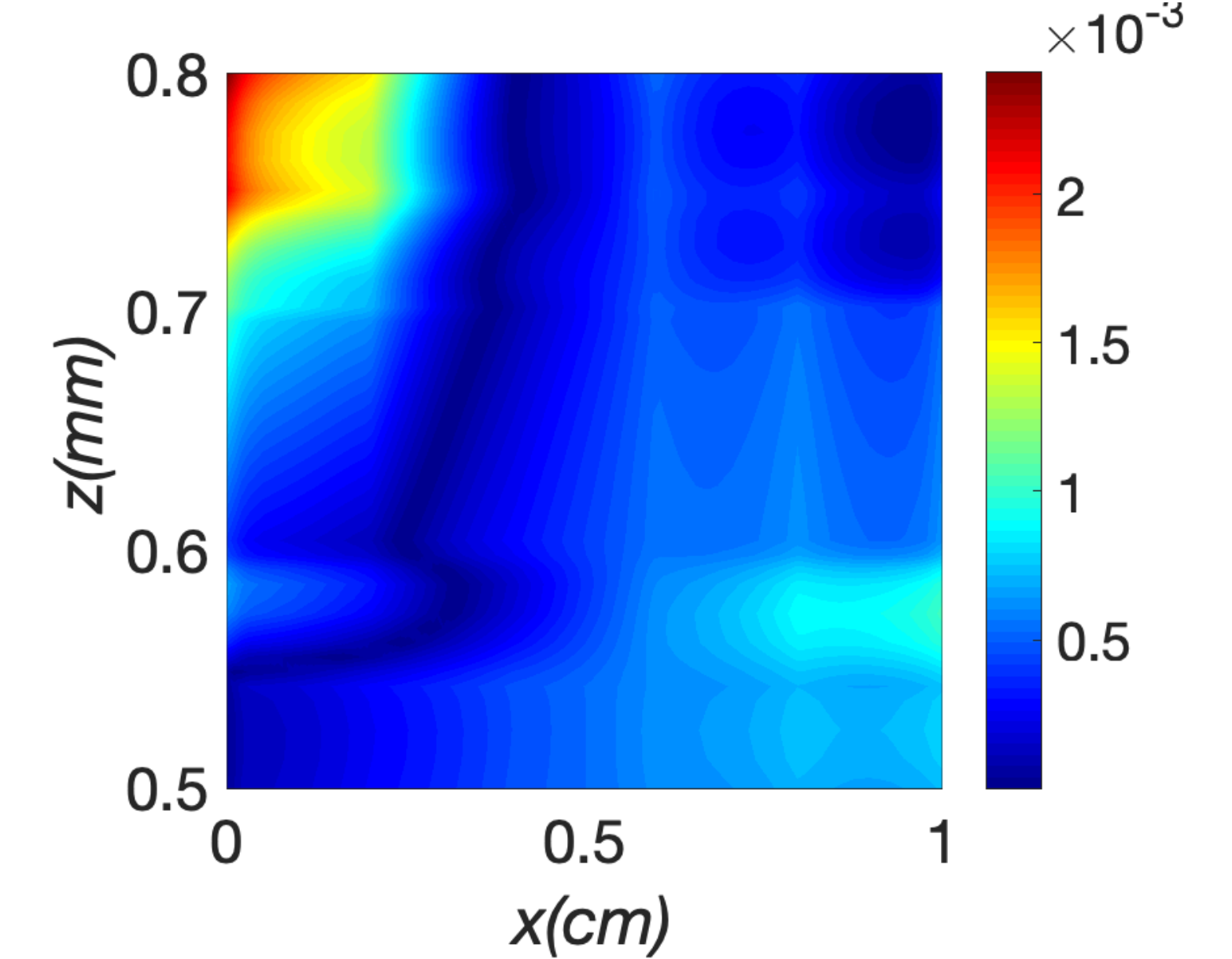}\\
		\includegraphics[width=0.32\textwidth]{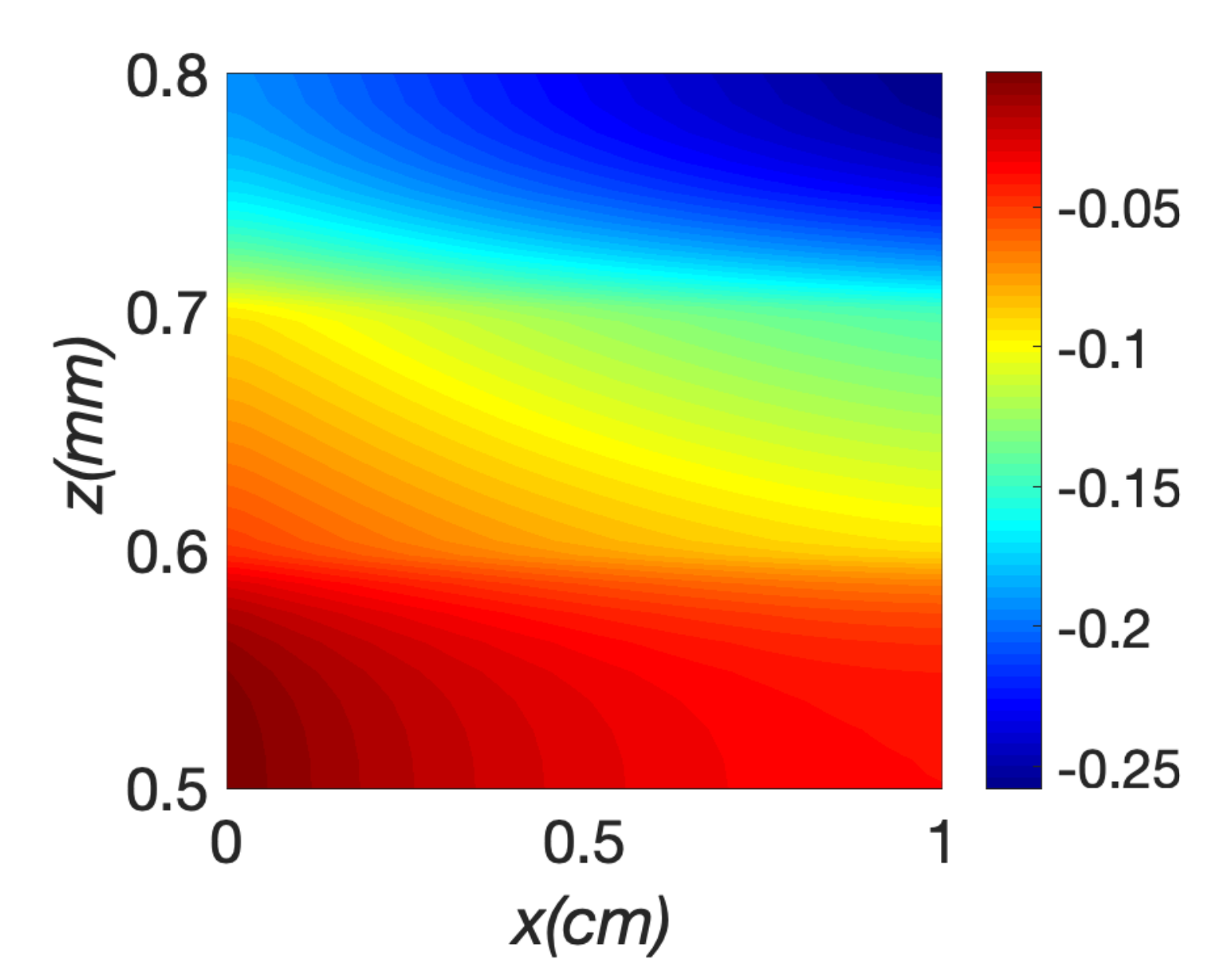}
	\includegraphics[width=0.32\textwidth]{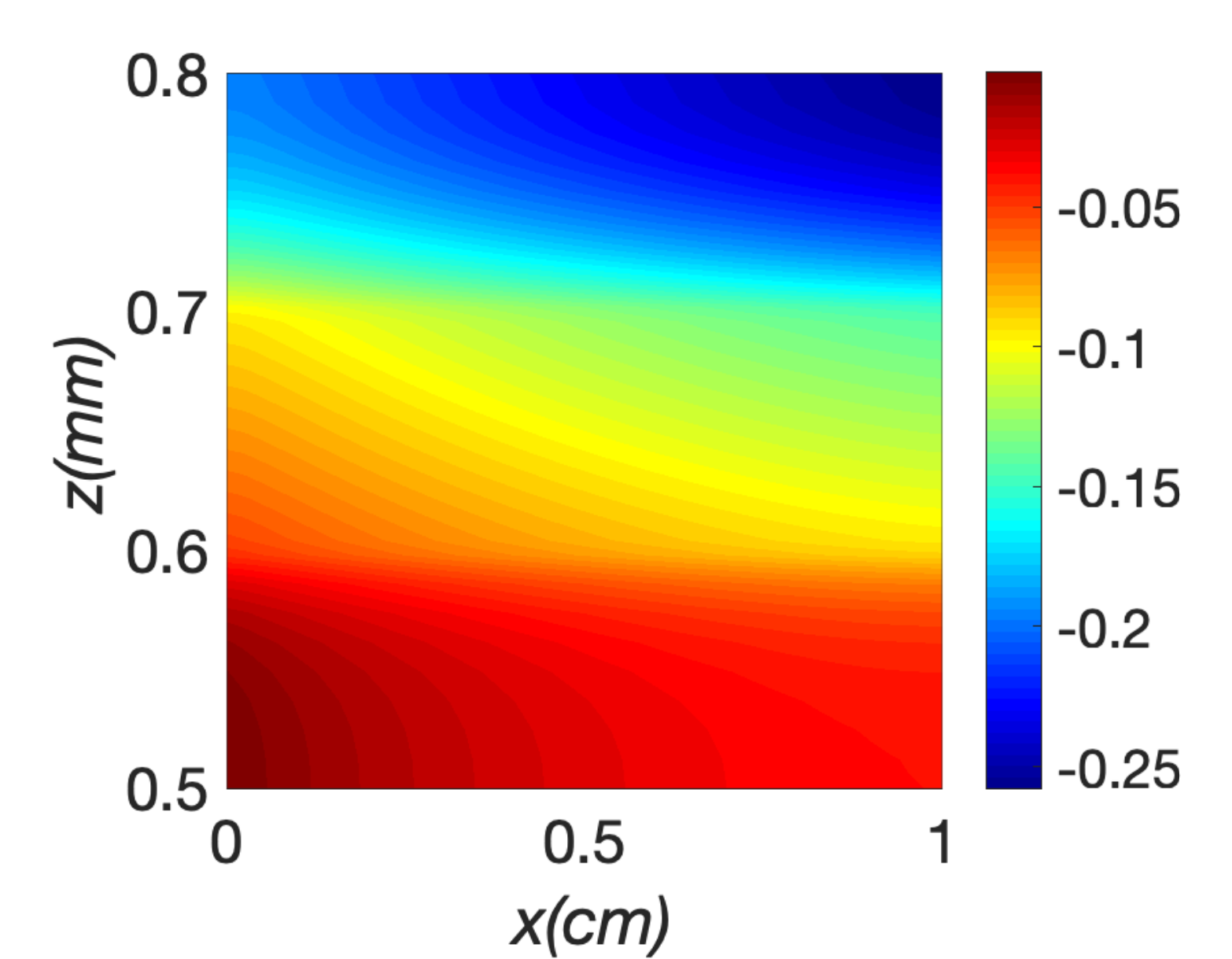}
	\includegraphics[width=0.32\textwidth]{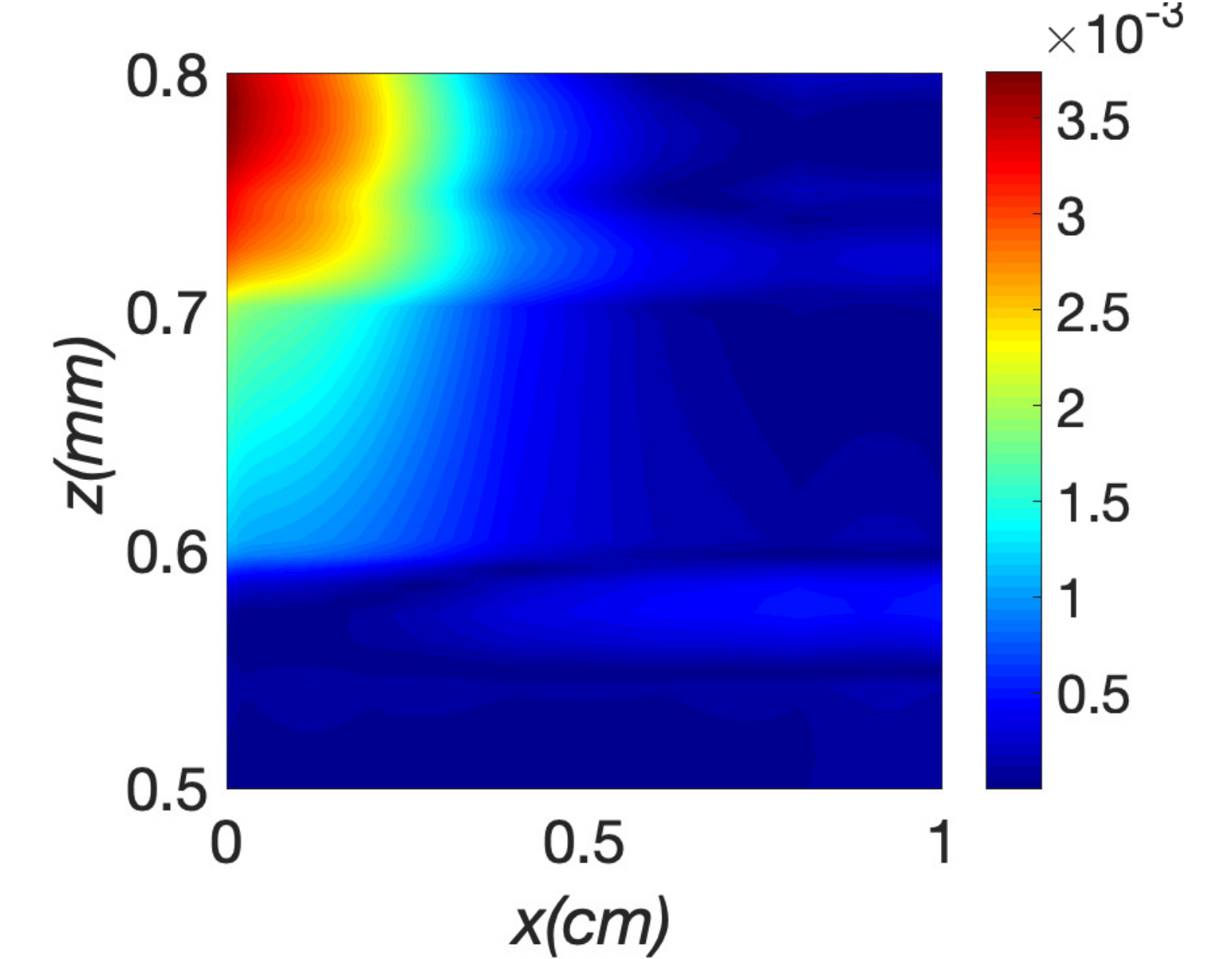}\\
		\includegraphics[width=0.32\textwidth]{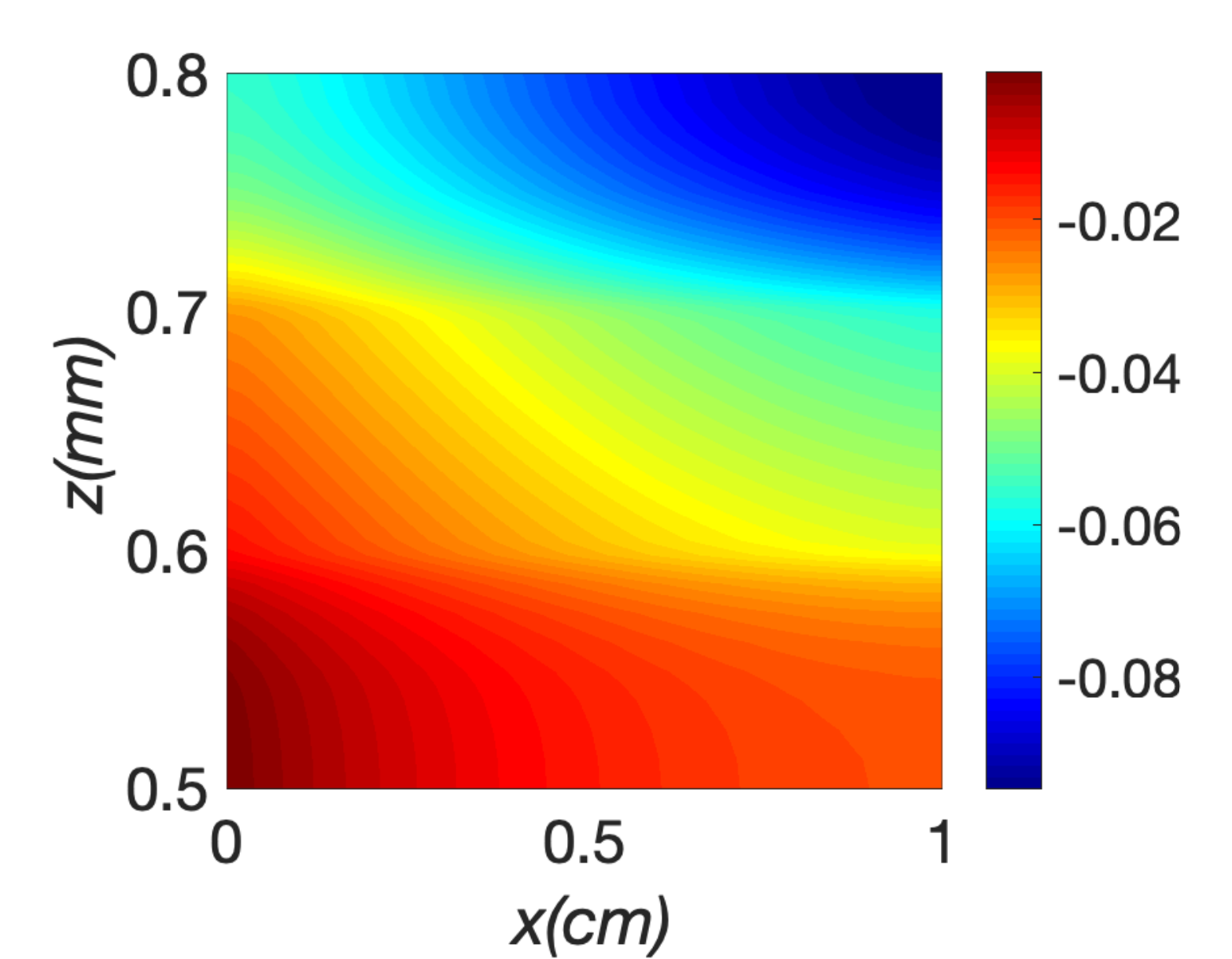}
	\includegraphics[width=0.32\textwidth]{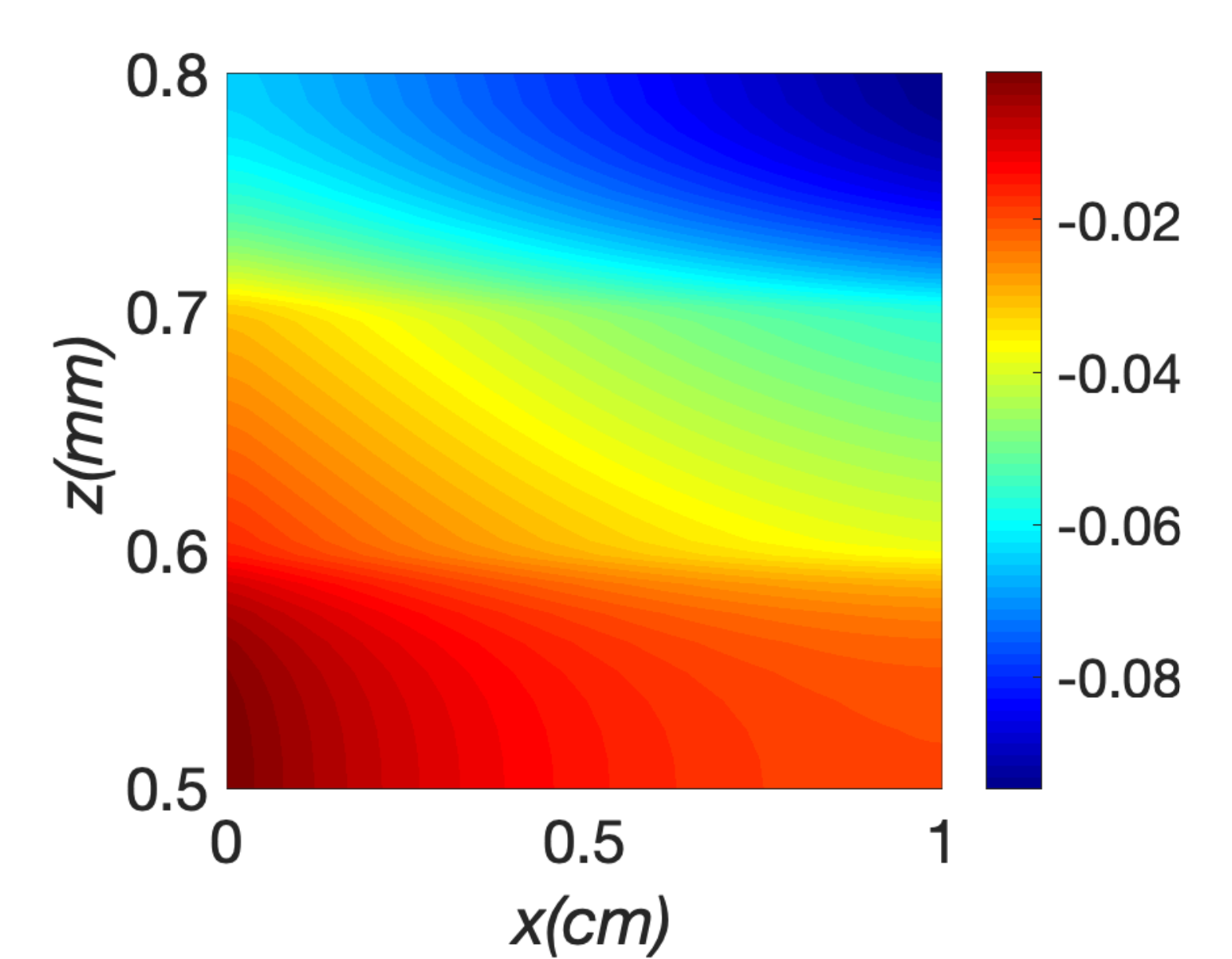}
	\includegraphics[width=0.32\textwidth]{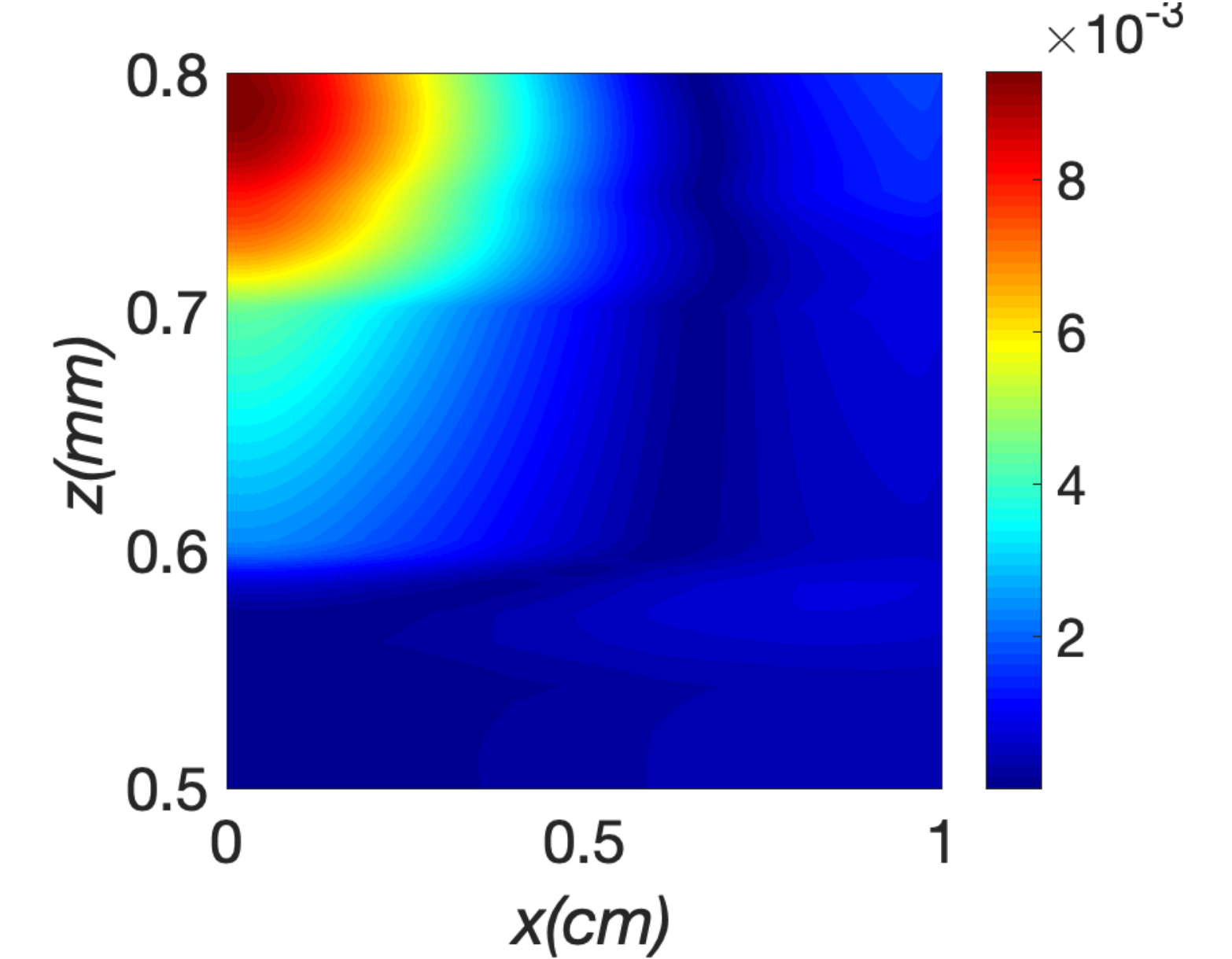}
	\caption{Predictions of the  ionic potential (V) in the $x-z$ plane located at the centre of the channels in Fig.  \ref{fig:sofc} for 40 F1 and 20 F2 training points.    {These predictions correspond from the top to bottom row to the lowest error, the median error  and the highest error for the 5-fold cross validation.} The columns from left to right are the prediction using \ours, the ground truth (test) and the pointwise absolute differences. }
	\label{fig:sofc3} 
\end{figure}

The inputs were taken to be the the electrode porosities $\epsilon\in[0.4,0.85]$, the 
cell voltage  $E_c\in[0.2,0.85]$ V, the temperature $T\in[973,1273]$ K, and the pressure in the channels  $P\in[0.5,2.5]$ atm. 60 inputs were selected using a Sobol sequence in the ranges indicated for the low- and high-fidelity simulations. A further 40 points were selected randomly (in the ranges above) for high-fidelity tests points. The low-fidelity F1 model used 3164 mapped elements (shown in Fig.  \ref{fig:sofc}) and  a relative tolerance of 0.1, while the high-fidelity model used  37064 elements and a relative tolerance of 0.001. The COMSOL model also uses a V cycle geometric multigrid. The quantities of interest were taken to be profiles of the electrolyte current density (A m$^{-2}$)  and ionic potential (V) in the $x-z$ plane located at the centre of the channels (Fig.  \ref{fig:sofc}).  In both cases, the number of points recorded was $d=100\times50=5000$  and both profiles were vectorised to form the training and test outputs. 

{{{The NRMSE (with five-fold cross validation) for 20 and 40 F1 training samples and an increasing number of F2 training samples is shown in Fig. \ref{sofc1}  and Table \ref{tableE3} in Appendix C  for \ours (with and without active learning), NARGP, Greedy NAR and SC.}  In this example, none of the other methods worked well, which is hardly surprising in the case of NARGP and Greedy NAR considering that $d=5000$. For SC, without out-of-sample F1 data, the performance is generally poor. \ours, on the other hand, shows a steady decline in the NRMSE as the number of F2 training samples is increased. {For 40 F1 and 10 F2 training points, \ours with active learning has a 97\% lower NRMSE than each of the other methods for the electrolyte current density prediction.  For the ionic potential, the equivalent values are almost identical, and these levels of improvement in the accuracy are maintained up to 40 F2 training points (over 90\% lower NRSME than all of the other methods).} } 

{Predictions of the quantities of interest for 40 F1 and 20 F2 training points using \ours (without active learning) are shown  in Figs. \ref{fig:sofc2} and \ref{fig:sofc3}, together with  the ground truths (tests) and pointwise absolute differences. The electrolyte current density predictions in Fig. \ref{fig:sofc2} correspond to  the lowest error ($\epsilon=0.67$, $E_c=0.32$ V, $T=1151.2$ K, $P=1.63$ atm), the median error ($\epsilon=0.48$, $E_c=0.36$ V, $T=981.7$ K, $P=1.97$ atm) and the highest error ($\epsilon=0.75$, $E_c=0.74$ V, $T=1025.7$ K, $P=1.24$ atm) for the 5-fold cross validation at 40 F1 and 20 F2 training points  in Fig. \ref{sofc1}(a). Likewise, the ionic potential predictions in Fig. \ref{fig:sofc3} correspond to  the lowest error ($\epsilon=0.79$, $E_c=0.28$ V, $T=1191.4$ K, $P=2.12$ atm), the median error ($\epsilon=0.59$, $E_c=0.30$ V, $T=1080.7$ K, $P=1.87$ atm) and the highest error ($\epsilon=0.81$, $E_c=0.69$ V, $T=926.6$ K, $P=1.04$ atm) for the 5-fold cross validation at 40 F1 and 20 F2 training points  in Fig. \ref{sofc1}(b). Even in the case of the largest error, the qualitative and quantitative accuracy of ResGP is high.}

\section{Summary and Conclusions}

In this paper we introduced an additive residual structure for multi-fidelity models in order to capture the connection between data at different fidelities. The result is a  non-parametric Bayesian model that is  equipped with  a closed-form solution for the predictive posterior. This permits tasks such as  uncertainty estimation and Bayesian optimization to be conducted efficiently and accurately.
Under a  noise-free assumption for the multi-fidelity data, the model scales efficiently  accurately  to high-dimensional problems. {We derive error bounds for the univariate case, which may be of use in applications such as control.}

{Four  benchmark problems in a variety of settings demonstrated that \ours can not only provide accurate posterior predictions but also faithful estimates of model uncertainty. The great drawback of SC is the requirement of out-of-sample simulations for making predictions, which the other methods, including \ours, avoid. When comparing with NARGP and Greedy NAR, the advantages of \ours are clear, not least in terms of stable and accurate predictions for high-dimensional problems. Both of these methods require large numbers of high-fidelity  training points to yield accurate predictions, since the number of parameters is high. This is a major drawback and a significant advantage of \ours, which is particularly good for sparse high-fidelity data. Moreover,  the accuracy is further improved for small data sets by appealing to active learning {\em via} variance reduction. Lastly, the performance of \ours is found to be markedly superior for high-dimensional problems since the number of parameters does not scale with the output space dimensionality. This makes \ours applicable to a broader range of problems than other state-of-the-art methods.}

\section*{Acknowledgment}
P. Wang were partially supported by the National Key Research and Development Program of China (Grant No. 2017YFB0701700) \& (Grant No. 2018YFB0703902). W.~Xing and S.~Zhe were supported by DARPA TRADES Award HR0011-17-2-0016. R. M. Kirby was sponsored by ARL under Cooperative Agreement Number W911NF-12-2-0023. The views and conclusions contained in this document are those of the authors and should not be interpreted as representing the official policies, either expressed or implied, of ARL or the U.S. Government. The U.S. Government is authorized to reproduce and distribute reprints for Government purposes not withstanding any copyright notation herein.

\clearpage

\renewcommand\thesubsection{\Alph{subsection}}
 \renewcommand{\theequation}{A-\arabic{equation}}
  \setcounter{equation}{0}  \setcounter{section}{0}
\section*{Appendices}
\subsection{Proofs}\label{proofs}

{	\begin{proof}[Proof Lemma \ref{th:errbound_with} ]
Using the definition of $\mu^{F}(\bxi)$ in (\ref{eq:rmgp noise free}) for the univariate case, we have
		\begin{equation}
		\begin{array}{ll}
 \displaystyle \left|\mu^{F}(\bxi)-\mu^{F}(\bxi')\right|& \displaystyle=
\left|\sum_{f=1}^F\left(\mu^{f}_r(\bxi)-\mu^{f}_r(\bxi')\right)\right| \vspace{2mm}\\
 & \displaystyle\le \sum_{f=1}^F\left|\mu^{f}_r(\bxi)-\mu^{f}_r(\bxi')\right| 
 \vspace{2mm}\\
 & \displaystyle =  \sum_{f=1}^F\left|\left[\k^{f}(\bxi) -\k^{f}(\bxi')\right]^T(\K^{f})^{-1}\R^{f}\right|
  \vspace{2mm}\\
 & \displaystyle \le  \sum_{f=1}^FL^f_k \sqrt{N_f}\lVert \bxi -\bxi'\rVert \lVert (\K^{f})^{-1}\R^{f}\rVert 
   \vspace{2mm}\\
 & \displaystyle = \left( \sum_{f=1}^FL^f_k \sqrt{N_f}\lVert (\K^{f})^{-1}\R^{f}\rVert \right)\lVert \bxi -\bxi'\rVert, \quad \forall \bxi,\bxi'\in\mathcal{X},
	\end{array}
	\end{equation}
	from which the bound on the Lipschitz constant $L_{\mu^F}$ is derived.
	\\
\indent For the standard deviation bound, we use the definitions of $v^f_r(\bxi)$, the univariate equivalent of $\V_r^f(\bxi)$, and $v^F(\bxi)$, the equivalent of $\V^F(\bxi)$ in (\ref{varwithout}), to obtain 
		\begin{equation}
		\begin{array}{ll}
 \displaystyle \left|v^{F}(\bxi)-v^{F}(\bxi')\right|& \displaystyle\le \sum_{f=1}^F\left|v^{f}_r(\bxi)-v^{f}_r(\bxi')\right| 
 \vspace{2mm}\\
 & \displaystyle=\sum_{f=1}^F\left|k^{f}(\bxi, \bxi|\btheta^{f}) -k^{f}(\bxi, \bxi'|\btheta^{f})+k^{f}(\bxi, \bxi'|\btheta^{f}) -k^{f}(\bxi', \bxi'|\btheta^{f})\right. 
  \vspace{2mm}\\
  & \displaystyle \le  \sum_{f=1}^F2L^f_k\lVert \bxi-\bxi'\rVert
 +\lVert\k^{f}(\bxi')-\k^{f}(\bxi)\rVert \lVert(\K^{f} )^{-1}\rVert_2 \lVert\k^{f}(\bxi)+\k^{f}(\bxi')\rVert, \quad \forall \bxi,\bxi'\in\mathcal{X}.
	\end{array}
	\end{equation}
	Since 	
			\begin{equation}
		\begin{array}{ll}&\lVert\k^{f}(\bxi')-\k^{f}(\bxi)\rVert \le \sqrt{N_f}L^f_k\lVert \bxi-\bxi'\rVert \vspace{2mm}\\ \mbox{and} \quad &\lVert\k^{f}(\bxi)+\k^{f}(\bxi')\rVert\le 2 \sqrt{N_f}\max\limits_{\bxi,\bxi'\in\mathcal{X}}k^{f}(\bxi, \bxi'|\btheta^{f}),
	\end{array}
	\end{equation}
we obtain
			\begin{equation}
		\begin{array}{ll}
 \displaystyle \left|v^{F}(\bxi)-v^{F}(\bxi')\right|& \displaystyle\le  \sum_{f=1}^F2L^f_k\lVert \bxi-\bxi'\rVert
 +\lVert\k^{f}(\bxi')-\k^{f}(\bxi)\rVert \lVert(\K^{f} )^{-1}\rVert_2 \lVert\k^{f}(\bxi)+\k^{f}(\bxi')\rVert,
	\end{array}
	\end{equation}
To obtain the modulus of continuity for the standard deviation $\sigma^F(\bxi)=\sqrt{v^F(\bxi)}=\sqrt{\sum_{f=1}^Fv^f_r(\bxi)}$ we first write 
\begin{align}
\left|v^F(\bxi)-v^F(\bxi')\right|=
\left|\sigma^F(\bxi)-\sigma^F(\bxi')\right|\left|\sigma^F(\bxi)+\sigma^F(\bxi')\right|\ge \left|\sigma^F(\bxi)-\sigma^F(\bxi')\right|^2,
	\end{align}
	since $\left|\sigma^F(\bxi)+\sigma^F(\bxi')\right|\ge \left|\sigma^F(\bxi)-\sigma^F(\bxi')\right|$ by the positive semidefiniteness of the standard deviation.
This yields
				\begin{equation}
		\begin{array}{ll}
 \displaystyle \left|\sigma^F(\bxi)-\sigma^F(\bxi')\right|& \displaystyle\le  \left[\sum_{f=1}^F2 L^f_k\left(1
 +N_f\max\limits_{\bxi,\bxi'\in\mathcal{X}}k^{f}(\bxi, \bxi'|\btheta^{f})\lVert(\K^{f} )^{-1}\rVert_2\right)\lVert \bxi-\bxi'\rVert\right]^{1/2},\quad \forall \bxi,\bxi'\in\mathcal{X},
	\end{array}
	\end{equation}
from which we obtain $\omega_{\sigma}^F(\cdot)$. 
 \end{proof}}
{	\begin{proof}[Proof Theorem \ref{th:errbound} ]
Pick $\delta\in(0,1)$. For every 
	design $\mathcal{X}_{\tau}^F\subset \mathcal{X}$ with $|\mathcal{X}_{\tau}^F|$ grid points and for a fill distance for this design satisfying
	\begin{align}
	\sup\limits_{\bxi\in\mathcal{X}} \min\limits_{\bxi'\in\mathcal{X}_{\tau}^F}
	\|\bxi-\bxi'\|\leq \tau,
	\label{eq:gridconstant}
	\end{align}
	it holds with probability of at least
	$1-|\mathcal{X}_{\tau}^F|\mathrm{e}^{-\beta(\tau)/2}$ that \citep{6138914} [Lemma 5.1]
	\begin{align}
	|y^F(\bxi)-\mu^{F}(\bxi)|\leq \sqrt{\beta(\tau)}\sigma^F(\bxi), 
	\quad \forall\bxi\in \mathcal{X}_{\tau}^F.
	\end{align}
	Choosing \mbox{$\beta(\tau)=2\log\left(\frac{|\mathcal{X}_{\tau}^F|}{\delta}\right)$}, the inequality 
	\begin{align}
	|y^F(\bxi)-\mu^{F}(\bxi)|\leq \sqrt{\beta(\tau)}\sigma^F(\bxi), 
	\quad \forall\bxi\in \mathcal{X}_{\tau}^F
	\end{align}
	holds with probability of at least $1-\delta$. Using Lemma \ref{th:errbound_with}   and the continuity of $y^{F}(\bxi)$, as well as the monotonicity of $\omega_{\sigma}^F(\cdot)$, for every $\bxi'\in\mathcal{X}_{\tau}^F$ and $\bxi\in\mathcal{X}$ we obtain
				\begin{equation}
		\begin{array}{ll}
 |y^F(\bxi)-\mu^{F}(\bxi)|&\displaystyle\le|y^F(\bxi)-y^F(\bxi')|+|\mu^{F}(\bxi')-\mu^{F}(\bxi)|+|y^F(\bxi')-\mu^{F}(\bxi')|\vspace{2mm}\\
  & \displaystyle \le
L_y\lVert \bxi-\bxi'\rVert+ L_{\mu^F}\lVert \bxi-\bxi'\rVert+\sqrt{\beta(\tau)}\sigma^F(\bxi') \vspace{2mm}\\
    & \displaystyle \le
 \tau L_y+\tau L_{\mu^F}+\sqrt{\beta(\tau)}\sigma^F(\bxi') \vspace{2mm}\\
   & \displaystyle =
 \tau L_y+\tau L_{\mu^F}+\sqrt{\beta(\tau)}|\sigma^F(\bxi) +\sigma^F(\bxi') -\sigma^F(\bxi)|\vspace{2mm}\\
  & \displaystyle \le
 \tau L_y+\tau L_{\mu^F}+\sqrt{\beta(\tau)}\omega_{\sigma}^F(\tau)+\sqrt{\beta(\tau)}\sigma^F(\bxi). 
	\end{array}
	\end{equation}
The final result follows from the fact that the minimum number of grid points satisfying condition (\ref{eq:gridconstant}) is given by $M(\tau,\mathcal{X})$.
	\end{proof}}
	
	 \renewcommand{\theequation}{B-\arabic{equation}}
  \setcounter{equation}{0}  	
\subsection{Synthetic examples}\label{synth}

{	\noindent 1. The Currin function is a two-dimensional problem with inputs $\bxi\in[0,1]^2$. The high- and low-fidelity  functions are given respectively by
	\begin{align}
		y^H\left(\bxi\right) = \left[1 - \exp\left(-\frac{1}{2\xi_2}\right)\right]\frac{2300\xi_1^3 + 1900\xi_1^2 + 2092\xi_1 + 60}{100\xi_1^3 + 500\xi_1^2 + 4\xi_1 + 20}, 
	\end{align}
	\begin{align}
		y^L\left(\bxi\right) = & \frac{1}{4}\left[y^H\left(\xi_1 + 0.05, \xi_2 + 0.05\right) + y^H\left(\xi_1 + 0.05, \max\left(0, \xi_2 - 0.05\right)\right)\right] +\\
		&\frac{1}{4}\left[y^H\left(\xi_1 - 0.05, \xi_2 + 0.05\right) + y^H\left(\xi_1 - 0.05, \max\left(0, \xi_2 - 0.05\right)\right)\right].
	\end{align}
	}
	
{	\noindent 2. The Park function is a four-dimensional problem with inputs $\bxi\in[0,1]^4$. 
The high- and low-fidelity  functions are given respectively by
	\begin{align}
		y^H\left(\bxi\right) = \frac{\xi_1}{2}\left[\sqrt{1 + \left(\xi_2 + \xi_3^2\right)\frac{\xi_4}{\xi_1^2}} - 1\right] + \left(\xi_1 + 3\xi_4\right)\exp\left[1 + \sin\left(\xi_3\right)\right],
	\end{align}
	\begin{align}
		y^L\left(\bxi\right) = \left[1 + \frac{\sin\left(\xi_1\right)}{10}\right]y^H\left(\bxi\right) - 2\xi_1 + \xi_2^2 + \xi_3^2 + 0.5.
	\end{align}
}

{	\noindent 3. The Borehole example is an eight-dimensional problem with inputs $\xi_1\in \left[0.05,0.15\right]$, $\xi_2\in\left[100,50000\right]$, $\xi_3\in\left[63070,115600\right]$, $\xi_4\in\left[990,1110\right]$, $\xi_5\in\left[63.1,115\right]$, $\xi_6\in\left[700,820\right]$, $\xi_7\in\left[1120,1680\right]$, $\xi_8\in\left[9855,12045\right]$.
The high- and low-fidelity  functions are given respectively by
	\begin{align}
		y^H\left(\bxi\right) = \frac{2\pi \xi_3\left(\xi_4-\xi_6\right)}{\log\left(\xi_2/\xi_1\right)\left(1 + \frac{2\xi_7\xi_3}{\log\left(\xi_2/\xi_1\right)\xi_1^2\xi_8}\right) + \frac{\xi_3}{\xi_5}},
	\end{align}
	\begin{align}
		y^L\left(\bxi\right) = \frac{5\xi_3\left(\xi_4 - \xi_6\right)}{\log\left(\xi_2/\xi_1\right)\left(1.5 + \frac{2\xi_7\xi_3}{\log\left(\xi_2/\xi_1\right)\xi_1^2\xi_8}\right) + \frac{\xi_3}{\xi_5}}.
	\end{align}
	}

{	\noindent 4. The three-level Branin function is a two-dimensional problem with inputs $\bxi\in\left[-5,10\right]\times\left[0,15\right]$. Three fidelities are considered, defined by
	\begin{align}
		y^1\left(\bxi\right) = \left(\frac{-1.275\xi_1^2}{\pi^2} + \frac{5\xi_1}{\pi} + \xi_2 - 6\right)^2 + \left(10 - \frac{5}{4\pi}\right)\cos\left(\xi_1\right) + 10,
	\end{align}
	\begin{align}
		y^2\left(\bxi\right) = 10\sqrt{y^H\left(\bxi - 2\right)} + 2\left(\xi_1 - 0.5\right) - 3\left(3\xi_2 - 1\right) -1,
	\end{align}
	\begin{align}
		y^3\left(\bxi\right) = y^2\left(1.2\left(\bxi + 2\right)\right) - 3\xi_2 + 1. 
	\end{align}}

{	\noindent 5. 	The Hartmann-3d example has inputs $\bxi\in\left[0,1\right]^3$.
	The  fidelity $f=1,2,3$ observations are given by
	\begin{align}
		y^f\left(\bxi\right) = \sum_{i=1}^4\alpha_i\exp\left(-\sum_{j=1}^3A_{ij}\left(\xi_j - P_{ij}\right)^2\right) \text{,}
	\end{align}
where 
		$$
			A =
			\begin{bmatrix}
			    3	   & 10 & 30 \\
			    0.1   & 10 & 35 \\
			    3      & 10 & 30 \\
			    0.1	  & 10 & 35
			\end{bmatrix} \quad\quad \text{and}\quad\quad 
			P = \begin{bmatrix}
				    0.3689	& 0.1170 & 0.2673 \\
				    0.4699  & 0.4387 & 0.7470 \\
				    0.1091  & 0.8732 & 0.5547 \\
				    0.0381  & 0.5743 & 0.8828
				\end{bmatrix}  \text{.}
		$$ 
\noindent $\pmb{\alpha}$ is set to $(1.0,1.2,3.0,3.2)^T$ and is updated to $\pmb{\alpha}_f = \pmb{\alpha} + (3 - f)\pmb{\delta}$ for lower fidelities, in which $\pmb{\delta} = (0.01,-0.01,-0.1,0.1)^T$.}

\subsection{{Normalized root mean square errors on the test sets for the multivariate examples 4.1-4.3}}
\renewcommand{\thetable}{C-\arabic{table}}
  \setcounter{table}{0} 

\begin{table}[H]
	\centering
	\caption{\label{tableE1}{Normalized root mean square errors (NRMSE) against 18 F2 test values on the two-fidelity turbulent mixing flow simulation for \ours with (ResGP) and without (ResGP-NA) active learning, NARGP, Greedy NAR and SC  (see Figure 2).}}
\scriptsize
\begin{tabular}{|l|c|c|c|c|c|c|c|c|c|}
\hline
\multicolumn{10}{|c|}{\bf Pressure profile near the pipe exit}\\
\hline
&\textbf{$N_1$}&\textbf{$N_2=5$}&\textbf{$N_2=10$}&\textbf{$N_2=15$}&\textbf{$N_2=20$}&\textbf{$N_2=25$}&\textbf{$N_2=30$}&\textbf{$N_2=35$}&\textbf{$N_2=40$}\\\hline
\textbf{ResGP}&20&0.3080&0.2369&0.1409&0.1267&N/A&N/A&N/A&N/A\\\hline
\textbf{ResGP}&40&0.2750&0.2428&0.1374&0.1267&0.1171&0.1063&0.0999&0.1014\\\hline
\textbf{ResGP-NA}&20&0.3344&0.2250&0.2098&0.2039&N/A&N/A&N/A&N/A\\\hline
\textbf{ResGP-NA}&40&0.3074&0.2051&0.1788&0.1706&0.1531&0.1387&0.1412&0.1323\\\hline
\textbf{NARGP}&20&0.7096&0.2902&0.2524&0.2481&N/A&N/A&N/A&N/A\\\hline
\textbf{NARGP}&40&0.6716&0.2253&0.1939&0.1645&0.1380&0.1395&0.1337&0.1347\\\hline
\textbf{SC}&20&0.3694&0.3694&0.3706&0.3706&N/A&N/A&N/A&N/A\\\hline
\textbf{SC}&40&0.3474&0.3486&0.3484&0.3507&0.3502&0.3502&0.3504&0.3504\\\hline
\textbf{GreedyNAR}&20&0.8927&0.2006&0.1461&0.1335&N/A&N/A&N/A&N/A\\\hline
\textbf{GreedyNAR}&40&0.9117&0.3801&0.1222&0.1088&0.1032&0.1014&0.0989&0.0985\\\hline
\multicolumn{10}{|c|}{\bf Velocity profile near the pipe exit}\\
\hline
\textbf{ResGP}&20&0.2159&0.2111&0.1803&0.1679&N/A&N/A&N/A&N/A\\\hline
\textbf{ResGP}&40&0.2002&0.1973&0.1643&0.1491&0.1406&0.1327&0.1250&0.1197\\\hline
\textbf{ResGP-NA}&20&0.2826&0.2495&0.2453&0.2399&N/A&N/A&N/A&N/A\\\hline
\textbf{ResGP-NA}&40&0.2515&0.2215&0.2122&0.2051&0.1917&0.1871&0.1823&0.1794\\\hline
\textbf{NARGP}&20&1.0145&1.0171&1.0188&1.0221&N/A&N/A&N/A&N/A\\\hline
\textbf{NARGP}&40&1.0145&1.0171&1.0188&1.0221&1.0275&1.0320&1.0279&1.0287\\\hline
\textbf{SC}&20&0.4606&0.4514&0.4521&0.4522&N/A&N/A&N/A&N/A\\\hline
\textbf{SC}&40&0.4297&0.4248&0.4246&0.4237&0.4240&0.4241&0.4239&0.4239\\\hline
\textbf{GreedyNAR}&20&1.0165&1.0148&1.0231&1.0232&N/A&N/A&N/A&N/A\\\hline
\textbf{GreedyNAR}&40&1.0102&1.0085&1.0178&1.0181&1.0175&1.0181&1.0178&1.0198\\\hline
\multicolumn{10}{|c|}{\bf Pressure profile near the pipe junction}\\
\hline
\textbf{ResGP}&20&0.1363&0.1593&0.0898&0.0653&N/A&N/A&N/A&N/A\\\hline
\textbf{ResGP}&40&0.1326&0.1665&0.0955&0.0697&0.0475&0.0398&0.0360&0.0351\\\hline
\textbf{ResGP-NA}&20&0.1556&0.0923&0.0872&0.0475&N/A&N/A&N/A&N/A\\\hline
\textbf{ResGP-NA}&40&0.1523&0.1111&0.1067&0.0690&0.0538&0.0500&0.0441&0.0483\\\hline
\textbf{NARGP}&20&0.3360&0.1055&0.1156&0.0812&N/A&N/A&N/A&N/A\\\hline
\textbf{NARGP}&40&0.3153&0.1061&0.1197&0.0851&0.0440&0.0711&0.0591&0.0478\\\hline
\textbf{SC}&20&0.3088&0.3048&0.3045&0.3041&N/A&N/A&N/A&N/A\\\hline
\textbf{SC}&40&0.2904&0.2884&0.2873&0.2879&0.2874&0.2870&0.2876&0.2874\\\hline
\textbf{GreedyNAR}&20&0.5140&0.0602&0.0434&0.0371&N/A&N/A&N/A&N/A\\\hline
\textbf{GreedyNAR}&40&0.6438&0.0610&0.0463&0.0321&0.0308&0.0293&0.0346&0.0345\\\hline
\multicolumn{10}{|c|}{\bf Velocity profile near the pipe junction}\\
\hline
\textbf{ResGP}&20&0.0312&0.0207&0.0177&0.0158&N/A&N/A&N/A&N/A\\\hline
\textbf{ResGP}&40&0.0289&0.0180&0.0151&0.0136&0.0133&0.0116&0.0113&0.0099\\\hline
\textbf{ResGP-NA}&20&0.0359&0.0310&0.0290&0.0288&N/A&N/A&N/A&N/A\\\hline
\textbf{ResGP-NA}&40&0.0336&0.0252&0.0225&0.0214&0.0196&0.0179&0.0153&0.0160\\\hline
\textbf{NARGP}&20&0.3080&0.0561&0.0312&0.0338&N/A&N/A&N/A&N/A\\\hline
\textbf{NARGP}&40&0.3076&0.0514&0.0252&0.0222&0.0218&0.0193&0.0249&0.0213\\\hline
\textbf{SC}&20&0.1373&0.1367&0.1367&0.1367&N/A&N/A&N/A&N/A\\\hline
\textbf{SC}&40&0.1238&0.1241&0.1241&0.1242&0.1242&0.1242&0.1242&0.1242\\\hline
\textbf{GreedyNAR}&20&0.9421&0.7107&0.0201&0.0177&N/A&N/A&N/A&N/A\\\hline
\textbf{GreedyNAR}&40&0.9456&0.8824&0.0199&0.0151&0.0129&0.0120&0.0119&0.0119\\\hline
\end{tabular}
\end{table}

\begin{table}[H]
	\centering
	\caption{\label{tableE2}{Normalized root mean square errors (NRMSE) against 34 F2 test values on the two-fidelity MD simulation for   ResGP with (ResGP) and without (ResGP-NA) active learning, NARGP, Greedy NAR and SC (see Figure 3).}}
\scriptsize
\begin{tabular}{|l|c|c|c|c|c|c|c|c|c|}
\hline
\multicolumn{10}{|c|}{\bf Radial distribution function}\\
\hline
&\textbf{$N_1$}&\textbf{$N_2=5$}&\textbf{$N_2=10$}&\textbf{$N_2=15$}&\textbf{$N_2=20$}&\textbf{$N_2=25$}&\textbf{$N_2=30$}&\textbf{$N_2=35$}&\textbf{$N_2=40$}\\\hline
\textbf{ResGP}&20&0.3296&0.0823&0.0580&0.0487&N/A&N/A&N/A&N/A\\\hline
\textbf{ResGP}&40&0.3318&0.0860&0.0583&0.0428&0.0392&0.0348&0.0310&0.0293\\\hline
\textbf{ResGP-NA}&20&0.2290&0.1454&0.1244&0.0527&N/A&N/A&N/A&N/A\\\hline
\textbf{ResGP-NA}&40&0.2222&0.1466&0.1269&0.0644&0.0488&0.0434&0.0363&0.0287\\\hline
\textbf{NARGP}&20&1.0103&0.8290&0.3844&0.0988&N/A&N/A&N/A&N/A\\\hline
\textbf{NARGP}&40&1.0103&0.7109&0.3046&0.0751&0.0505&0.0534&0.0449&0.0319\\\hline
\textbf{SC}&20&0.1540&0.1393&0.1374&0.1372&N/A&N/A&N/A&N/A\\\hline
\textbf{SC}&40&0.1518&0.1385&0.1367&0.1357&0.1354&0.1352&0.1351&0.1351\\\hline
\textbf{GreedyNAR}&20&0.9900&0.9806&0.9565&0.0376&N/A&N/A&N/A&N/A\\\hline
\textbf{GreedyNAR}&40&0.9952&0.9658&0.9504&0.9406&0.9218&0.0351&0.0331&0.0303\\\hline
\multicolumn{10}{|c|}{\bf Mean Squared Distance}\\
\hline
\textbf{ResGP}&20&0.5999&0.4338&0.4408&0.4223&N/A&N/A&N/A&N/A\\\hline
\textbf{ResGP}&40&0.3610&0.2264&0.1855&0.1866&0.1713&0.1675&0.1467&0.1593\\\hline
\textbf{ResGP-NA}&20&0.5331&0.5510&0.5101&0.4814&N/A&N/A&N/A&N/A\\\hline
\textbf{ResGP-NA}&40&0.3111&0.3143&0.2869&0.2570&0.2180&0.2135&0.2102&0.2213\\\hline
\textbf{NARGP}&20&0.9370&1.1201&0.9763&0.8112&N/A&N/A&N/A&N/A\\\hline
\textbf{NARGP}&40&0.9383&1.0250&0.9072&0.7806&0.8609&0.7446&0.2478&0.3128\\\hline
\textbf{SC}&20&0.7563&0.7500&0.7422&0.7417&N/A&N/A&N/A&N/A\\\hline
\textbf{SC}&40&0.7554&0.7434&0.7258&0.7200&0.7145&0.7139&0.7114&0.7107\\\hline
\textbf{GreedyNAR}&20&1.0001&0.9783&0.6462&0.6822&N/A&N/A&N/A&N/A\\\hline
\textbf{GreedyNAR}&40&1.0062&1.0007&0.9903&0.5576&0.6384&0.7679&0.5913&0.6605\\\hline
\multicolumn{10}{|c|}{\bf Self diffusion coefficient}\\
\hline
\textbf{ResGP}&20&0.5757&0.4915&0.4389&0.4017&N/A&N/A&N/A&N/A\\\hline
\textbf{ResGP}&40&0.4418&0.3335&0.2765&0.2397&0.2062&0.1940&0.1857&0.1833\\\hline
\textbf{ResGP-NA}&20&0.6836&0.5135&0.4491&0.4226&N/A&N/A&N/A&N/A\\\hline
\textbf{ResGP-NA}&40&0.5674&0.3646&0.2848&0.2492&0.2198&0.2228&0.2221&0.2160\\\hline
\textbf{NARGP}&20&1.0133&0.5587&0.4109&0.4816&N/A&N/A&N/A&N/A\\\hline
\textbf{NARGP}&40&1.0226&0.7069&0.5668&0.6271&0.5875&0.4556&0.3093&0.3510\\\hline
\textbf{SC}&20&0.8465&0.8497&0.8512&0.8506&N/A&N/A&N/A&N/A\\\hline
\textbf{SC}&40&0.8227&0.8109&0.8089&0.8083&0.8086&0.8080&0.8074&0.8072\\\hline
\textbf{GreedyNAR}&20&0.9353&0.7357&0.5170&0.5246&N/A&N/A&N/A&N/A\\\hline
\textbf{GreedyNAR}&40&0.9277&0.7682&0.2876&0.2433&0.2450&0.2504&0.2513&0.2520\\\hline
\end{tabular}
\end{table}

\begin{table}[H]
	\centering
	\caption{\label{tableE3}{Normalized root mean square errors (NRMSE) against 40 F2 test values on  the two-fidelity SOFC simulation for ResGP with (ResGP) and without (ResGP-NA) active learning, NARGP, Greedy NAR and SC (see Figure 5).}}
\scriptsize
\begin{tabular}{|l|c|c|c|c|c|c|c|c|c|}
\hline
\multicolumn{10}{|c|}{\bf Electrolyte current density}\\
\hline
&\textbf{$N_1$}&\textbf{$N_2=5$}&\textbf{$N_2=10$}&\textbf{$N_2=15$}&\textbf{$N_2=20$}&\textbf{$N_2=25$}&\textbf{$N_2=30$}&\textbf{$N_2=35$}&\textbf{$N_2=40$}\\\hline
\textbf{ResGP}&20&0.0811&0.0375&0.0322&0.0311&N/A&N/A&N/A&N/A\\\hline
\textbf{ResGP}&40&0.0752&0.0279&0.0204&0.0183&0.0157&0.0157&0.0152&0.0151\\\hline
\textbf{ResGP-NA}&20&0.0711&0.0561&0.0536&0.0494&N/A&N/A&N/A&N/A\\\hline
\textbf{ResGP-NA}&40&0.0585&0.0308&0.0275&0.0229&0.0201&0.0204&0.0196&0.0185\\\hline
\textbf{NARGP}&20&1.0079&1.0063&1.0142&1.0138&N/A&N/A&N/A&N/A\\\hline
\textbf{NARGP}&40&1.0079&1.0063&1.0142&1.0138&0.8145&0.6257&0.6245&0.6247\\\hline
\textbf{SC}&20&0.9745&0.9752&0.9752&0.9752&N/A&N/A&N/A&N/A\\\hline
\textbf{SC}&40&0.9048&0.9053&0.9054&0.9054&0.9054&0.9054&0.9054&0.9054\\\hline
\textbf{GreedyNAR}&20&0.9895&0.9758&0.9641&0.9524&N/A&N/A&N/A&N/A\\\hline
\textbf{GreedyNAR}&40&0.9890&0.9778&0.9664&0.9527&0.9409&0.9274&0.9127&0.9359\\\hline
\multicolumn{10}{|c|}{\bf Ionic potential}\\
\hline
\textbf{ResGP}&20&0.0549&0.0350&0.0342&0.0302&N/A&N/A&N/A&N/A\\\hline
\textbf{ResGP}&40&0.0446&0.0236&0.0209&0.0141&0.0131&0.0132&0.0130&0.0131\\\hline
\textbf{ResGP-NA}&20&0.0482&0.0491&0.0479&0.0467&N/A&N/A&N/A&N/A\\\hline
\textbf{ResGP-NA}&40&0.0268&0.0219&0.0213&0.0207&0.0175&0.0176&0.0167&0.0163\\\hline
\textbf{NARGP}&20&1.0106&1.0078&0.8257&0.4638&N/A&N/A&N/A&N/A\\\hline
\textbf{NARGP}&40&1.0106&1.0078&0.8252&0.4573&0.4243&0.2819&1.0458&0.3637\\\hline
\textbf{SC}&20&0.9460&0.9464&0.9465&0.9465&N/A&N/A&N/A&N/A\\\hline
\textbf{SC}&40&0.8798&0.8800&0.8800&0.8800&0.8800&0.8800&0.8800&0.8800\\\hline
\textbf{GreedyNAR}&20&0.9974&0.9861&0.9730&0.9586&N/A&N/A&N/A&N/A\\\hline
\textbf{GreedyNAR}&40&1.0038&0.9970&0.9905&0.9745&0.9661&0.9532&0.9399&0.9843\\\hline
\end{tabular}
\end{table} 

\end{document}